\documentclass[pdflatex,sn-chicago,iicol]{sn-jnl}

\usepackage{gensymb}
\usepackage{hyperref}
\usepackage{algpseudocode}
\usepackage{algorithm}

\newcommand{\RR}{\mathbb{R}}
\newcommand{\s}{\sigma}
\newcommand{\D}{{\boldsymbol{\Delta}}}
\newcommand{\de}{{\boldsymbol{\delta}}}
\newcommand{\X}{{\bf X}}
\newcommand{\CC}{{\bf C}}
\newcommand{\E}{{\bf E}}
\newcommand{\K}{{\bf K}}
\newcommand{\N}{{\bf N}}
\newcommand{\U}{{\bf U}}
\newcommand{\V}{{\bf V}}

\newcommand{\Rcell}{{}^{cam} \! R_{ell}}
\newcommand{\Rccone}{{}^{cam} \! R_{cone}}
\newcommand{\Rellcone}{{}^{ell} \! R_{cone}}

\newcommand{\ab}{$\{A,B'\}$}
\newcommand{\abb}{$\{A,B'\}${ }}

\newcommand{\rev}[1]{\textcolor{red}{#1}}

\newif\ifhighlightrev
\highlightrevfalse

\newif\ifboxes
\boxesfalse

\newif\ifboxeseq
\boxeseqfalse

\newtheorem{res}{Result}

\jyear{2022}%

\theoremstyle{plain}
\newtheorem{theo}{Theorem}

\begin{document}

\title[Perspective-$1$-Ellipsoid]{Perspective-$1$-Ellipsoid: Formulation, Analysis and Solutions of the Camera Pose Estimation Problem from One Ellipse-Ellipsoid Correspondence}


\author*[1,2]{\fnm{Vincent} \sur{Gaudilli\`ere}}\email{vincent.gaudilliere@uni.lu}

\author[1]{\fnm{Gilles} \sur{Simon}}\email{gilles.simon@loria.fr}

\author[1]{\fnm{Marie-Odile} \sur{Berger}}\email{marie-odile.berger@inria.fr}

\affil[1]{\orgname{Université de Lorraine, CNRS, Inria, LORIA}, \orgaddress{\postcode{F-54000}, \city{Nancy}, \country{France}}}

\affil[2]{\orgdiv{SnT - Interdisciplinary Centre for Security, Reliability and Trust}, \orgname{University of Luxembourg}, \orgaddress{\street{29 Avenue John F. Kennedy}, \postcode{L-1855}, \city{Luxembourg}, \country{Luxembourg}}}


\abstract{
In computer vision, camera pose estimation from correspondences between 3D geometric entities and their projections into the image has been a widely investigated problem. Although most state-of-the-art methods exploit low-level primitives such as points or lines, the emergence of very effective CNN-based object detectors in the recent years has paved the way to the use of higher-level features carrying semantically meaningful information. Pioneering works in that direction have shown that modelling 3D objects by ellipsoids and 2D detections by ellipses offers a convenient manner to link 2D and 3D data. However, the mathematical formalism most often used in the related litterature does not enable to easily distinguish ellipsoids and ellipses from other quadrics and conics, leading to a loss of specificity potentially detrimental in some developments. Moreover, the linearization process of the projection equation creates an over-representation of the camera parameters, also possibly causing an efficiency loss. In this paper, we therefore introduce an ellipsoid-specific theoretical framework and demonstrate its beneficial properties in the context of pose estimation. More precisely, we first show that the proposed formalism enables to reduce the pose estimation problem to a position or orientation-only estimation problem in which the remaining unknowns can be derived in closed-form. Then, we demonstrate that it can be further reduced to a 1 Degree-of-Freedom (1DoF) problem and provide the analytical derivations of the pose as a function of that unique scalar unknown. We illustrate our theoretical considerations by visual examples
\ifhighlightrev
    \rev{
\fi
and include a discussion on the practical aspects. Finally, we release this paper along with the corresponding source code in order to contribute towards more efficient resolutions of ellipsoid-related pose estimation problems. The source code is available here: \url{https://gitlab.inria.fr/vgaudill/p1e}.
\ifhighlightrev
    }
\fi
}

\keywords{Pose estimation, Object modeling, Ellipsoid, Ellipse}

\maketitle

\section{Introduction}
\label{sec:intro}

\begin{figure}
    \centering
    \includegraphics[width=\linewidth]{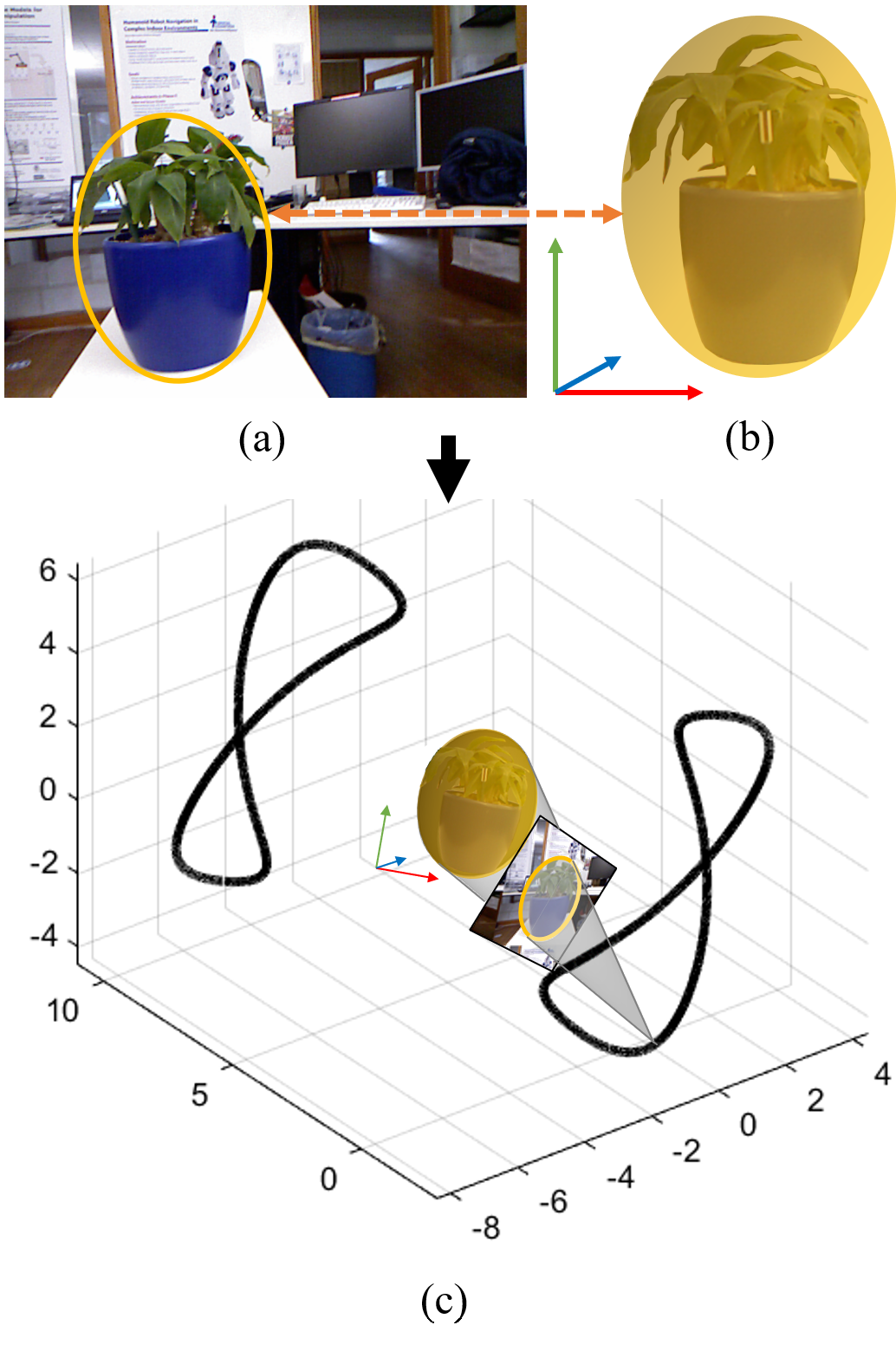}
    \caption{The \textit{Perspective-1-Ellipsoid} (P1E) problem aims at estimating the locus of solutions (c) to the camera pose estimation problem from one ellipse-ellipsoid correspondence (a,b).}
    \label{fig:teaser}
\end{figure}

Estimating the relative pose between a camera and a scene has been representing a core aspect of computer vision for many years. Indeed, this task is at the root of many applications, from robot navigation \citep{Bonin-FontOO08} to Augmented Reality \citep{MarchandUS16}. 

Historically, pose estimation has been addressed by leveraging 2D-3D correspondences between low-level geometric features such as points (P$n$P: Perspective-$n$-Point)\citep{LepetitMF09,Hartley2004} or lines (P$n$L: Perspective-$n$-Line)\citep{PnL}. More recently, the field has been significantly impacted by the raise of deep learning and pose estimation is now widely addressed by end-to-end trainable methods \citep{HoqueAXMW21}. However, while deep learning has proven to be indispensable in solving the problem of perception, it is still not the best choice in terms of accuracy throughout all steps of a pose estimation pipeline, as figured out in recent object pose estimation challenges \citep{KisantalSPIMD20,PARK2023}. In short, most accurate methods were \textit{hybrid} approaches in which keypoints are located by deep regression models then used as inputs to a P$n$P solver.

Following the same proven hybrid approach, the appearance of very effective object detectors in the recent years \citep{Redmon_2016_CVPR,Redmon_2017_CVPR,YOLOv3,LiuAESRFB16,Girshick_2014_CVPR,Girshick_2015_ICCV} has been enabling the substitution of low-level primitives (\textit{e.g.} points, lines), often extracted in droves and carrying limited semantic information, by object-level features providing a deeper scene understanding at a lower computational matching cost. Therefore, the choice of object representation has become crucial.

While modeling 3D objects by cuboids along with their 2D projections by bounding boxes has been investigated in the context of pose computation in unknown scenes \citep{context_relevance_iros_2017,wide_baseline_2018,LiMD19}, it appears that the ellipse-ellipsoid modeling paradigm has the unparalleled advantage of analytically linking 2D and 3D models \citep{Hartley2004,Eberly-backproj}. In other words, ellipsoids always project onto planes in the form of ellipses, and the underlying closed-form projection equation can be leveraged to efficiently solve the pose estimation problem \citep{Crocco_2016_CVPR,7919240,IROS,ISMAR,IROS2,RAL}. As an indicator of that increasingly attractive research direction, more and more object detectors have been modeling object projections by ellipses instead of traditional bounding boxes aligned with image axes \citep{Li19,PanFWZR21,ZhaoJFLY21,ZinsSB20,DongRPI21,abs-2101-05212}.

However, performing pose estimation at the level of objects through ellipse/ellipsoid-modeling has mainly been formulated under the standard projective geometry formalism \citep{Hartley2004} \citep{Crocco_2016_CVPR,Gay_2017_ICCV,7919240,QSLAM,ROB-059,abs-2004-05303,abs-2110-08977}, and mostly through Least Squares estimations where the unknowns are general quadric surfaces.
This framework may present limitations since ellipses (resp. ellipsoids) are specific categories of conics (quadrics) and since the linearization of the projection equation increases the numbers of apparent unknowns (see Section \ref{sec:rel-work}). In addition, these papers do not address the case of a small number of ellipse-ellipsoid correspondences, which is of high practical importance when a few objects are observed or when computing the solutions with minimal sampling size in RANSAC-like algorithms \citep{FischlerB81}.


\ifhighlightrev
    \rev{
\fi
In this paper, we address the fundamental problem of camera pose estimation from one ellipse-ellipsoid correspondence (see Fig.~\ref{fig:teaser}), referred to as \textit{Perspective-$1$-Ellipsoid} (P$1$E) in what follows. Conversely, it consists in estimating the pose of an ellipsoid of known size from its projected ellipse. We assume that the camera intrinsic parameters are known, and rely on an ellipsoid-specific Euclidean formalism.
\ifhighlightrev
    }
\fi

There are several interests in solving the P$1$E problem.
\ifhighlightrev
    \rev{
\fi
First, on the theoretical side, and except in the case of a spherical object, we demonstrate that the solutions are a variety of dimension 1 and provide an effective way to compute the camera locus (\textit{i.e.} set of solutions). 
\ifhighlightrev
    }
\fi
This problem has been addressed in \cite{WokesP10} in the particular case of a spheroid (specific ellipsoid having an axis of revolution) but was never considered for general ellipsoids. To the best of our knowledge, we are the first to propose a constructive solution to the P$1$E problem without resorting to any additional approximation nor prior knowledge.

In this study, we also consider two particular cases of important practical interest: (i) computing the camera position when the orientation is known (ii) computing the orientation when the position is known.

Besides the theoretical aspects, solving the P$1$E problem opens the way towards automatic positioning solutions in texture-less or low-textured environments, for instance leveraging several ellipse-ellipsoid correspondences or one with several point pairs. Industrial or other indoor scenes, in which objects would be approximated by ellipsoids, represent concrete places that could take advantage of these results.



The paper is organized as follows:  In Section \ref{sec:rel-work}, we discuss the State-of-the-Art ellipsoid-based pose estimation methods and the limits of the homogeneous representation of ellipses and ellipsoids, justifying investigating another representation. In Section \ref{sec:Eberly-eq-res}, we present the Euclidean formulation of the relative ellipse-ellipsoid pose estimation problem, previously introduced in \cite{Eberly-backproj}. 
\ifhighlightrev
    \rev{
\fi
Sections \ref{sec:properties} and \ref{sec:decoupl} combine some results already introduced in our previous papers with novel ones, while Section \ref{sec:closedform} contains the core contribution of this paper. More precisely: 
\begin{itemize}
    \item In Section 4, we exhibit several mathematical properties of the P$1$E solutions needed in the demonstrations of Sections \ref{sec:decoupl} and \ref{sec:closedform}. Results \ref{res:mu}, \ref{res:Delta2mu} and \ref{res:Link-sD} have never been introduced before and represent contributions of this work. 
    \item Section \ref{sec:decoupl} is dedicated to the specific case of P$1$E where either the relative position or orientation is known. We recall that the problem formalism induces an inherent decoupling between orientation and position, \textit{i.e.} one of which can be inferred in closed-form from the other one. This decoupling property and the two sub-cases derivations have already been introduced and leveraged in practical settings in our previous work (position from orientation: \cite{IROS,ISMAR,IROS2,RAL}, orientation from position: \cite{Rathinam_2022_IAC}). 
    \item The general P$1$E problem is solved in Section \ref{sec:closedform}. We demonstrate that the 6DoF pose estimation problem can be reduced to a 1DoF problem, and present a complete characterization of the solutions according to the type of ellipsoid under consideration. These results are all unpublished to date, and represent the core contribution of this paper.
    \item Section \ref{sec:discussion} discusses how these theoretical results can be leveraged in practice.
\end{itemize}
\ifhighlightrev
    }
\fi

\section{Related Work}
\label{sec:rel-work}


Most methods dealing with pose estimation at the level of objects and using ellipsoidal models are based on the projective geometry formalism. Under this framework, any quadric $Q\in\RR^{4\times4}$ is linked to its projected conic $C\in\RR^{3\times3}$ by the Equation
\begin{equation*}
    C^{*}=PQ^{*}P^{\top}
\end{equation*}
where  $P\in\RR^{3\times4}$ is the camera projection matrix  
\ifhighlightrev
    \rev{
\fi
and $Q^{*}~=~Q^{-1},C^{*}~=~C^{-1}$ (up to scale factors) are the adjugate matrices of $Q$ and $C$ \citep{Hartley2004} (Result 8.9, page 201).
\ifhighlightrev
    }
\fi
It is important noting that, under that formalism, ellipses and ellipsoids can be distinguished from other conics and quadrics only by certain algebraic conditions on $C$ and $Q$ entries, these conditions being difficult to leverage in practice.

Quadric-modeling of objects has been implemented in the context of Semantic SLAM \citep{ROB-059} for improving the process accuracy through multi-objective optimization \citep{QSLAM,abs-2110-08977,abs-2004-05303}. On a theoretical level, \cite{Crocco_2016_CVPR} addresses the object-based Structure-from-Motion (\textit{SfM}) problem and introduces an analytical solution to reconstruct both quadrics and affine camera poses. The problem is solved in a
\ifhighlightrev
    \rev{
\fi
Least Squares
\ifhighlightrev
    }
\fi
sense with a matrix $P$ over-represented by a $6\times 10$  Kronecker product $P \otimes P$. This work is extended with CAD model priors in \cite{Gay_2017_ICCV}, while \cite{7919240} present a closed-form solution to the problem of reconstructing a quadric from three calibrated pinhole camera views in which the object projections are detected. However, in this method, nothing ensures that the reconstructed quadric is an ellipsoid, forcing the authors to add a costly post-processing non-linear optimization of the results.

Another limitation while using homogeneous quadrics and over-representation of $P$ appears in \cite{P12Q}. In this method, the so-called \textit{gold-standard} algorithm \citep{Hartley2004} initially used to retrieve the camera projection matrix from 2D-3D point correspondences is adapted to conic-quadric correspondences. To compute $P$, 12 conic-quadric pairs are required whereas only 6 point pairs are sufficient under the same conditions.

We argue that these limitations may be due in part to the fact that ellipse and ellipsoid homogeneous formulations are not clearly enough distinguished from other members of their geometric families, and also to a non-minimal representation of the projection matrix. To overcome these difficulties, we therefore propose an ellipsoid-specific theoretical framework and highlight its advantages.




\ifhighlightrev
    \rev{
\fi
\section{Formulation of the P$1$E Problem}
\label{sec:Eberly-eq-res}
\ifhighlightrev
    }
\fi

\ifhighlightrev
    \rev{
\fi
This section aims at providing the fundamental equation of the P1E problem, i.e. the relation between an ellipsoid and its projected ellipse, given the projection parameters. The way followed consists in formulating, on the one hand, the equation of the projection cone associated with the ellipsoid and, on the other hand, the equation of the backprojection cone associated with the ellipse. The P1E equation is then obtained by stating that the two cones are aligned.
\ifhighlightrev
    }
\fi

\subsection{Problem Statement}
\label{ssec:Eberly-eq}

\begin{figure}[thpb]
      \centering
      \includegraphics[trim = 32mm 0mm 75mm 5mm, clip, width=\linewidth]{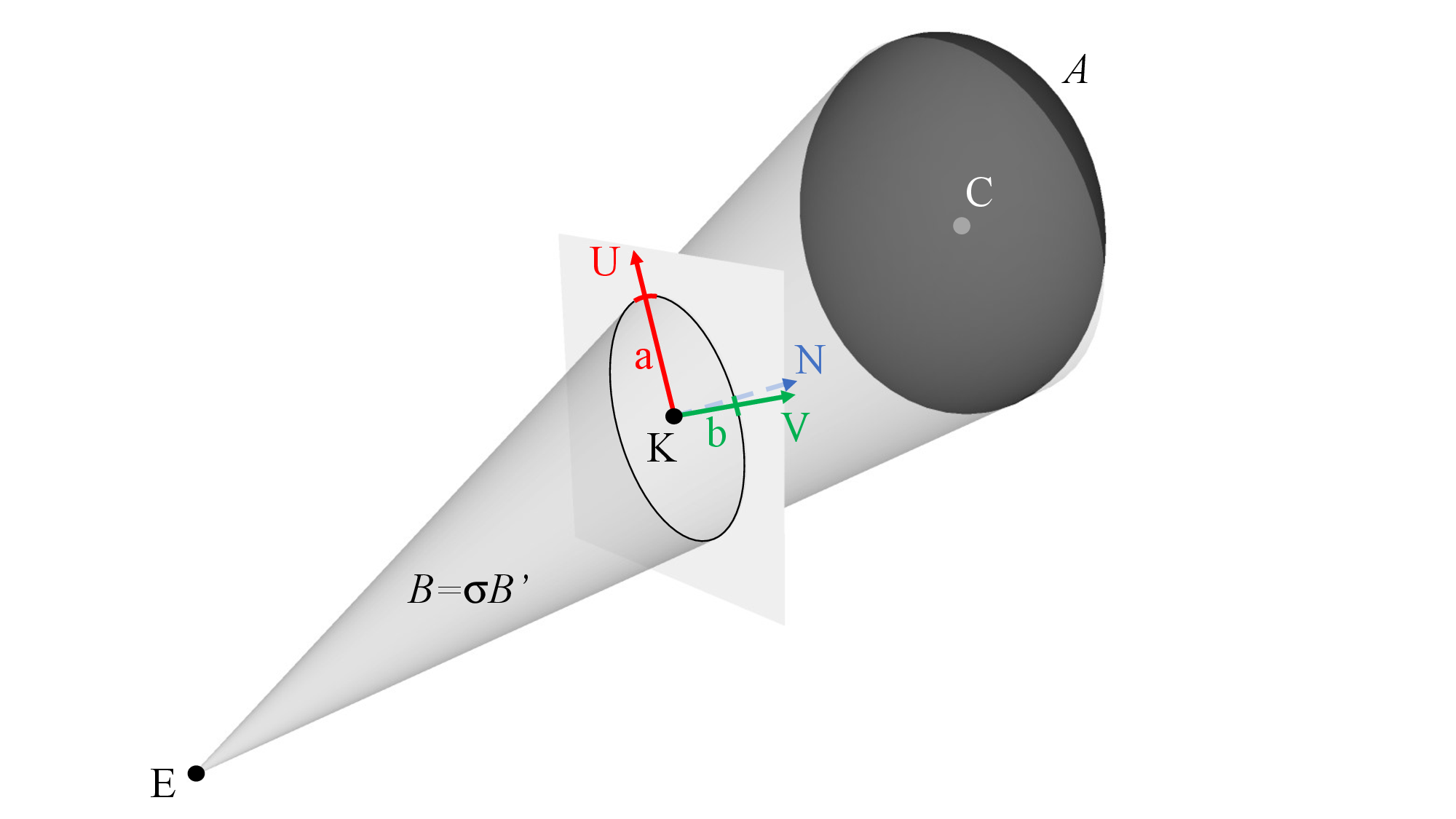}
      \caption{Illustrating the image plane, camera center, ellipsoid and projected ellipse.}
      \label{fig:schema}
\end{figure}

Following the notations introduced in \cite{Eberly-backproj} and presented in Fig. \ref{fig:schema}, we consider an ellipsoid $(\CC,A)$ defined by Equation
\begin{equation*}
    (\X-\CC)^\top A(\X-\CC)=1\mbox{,}
\end{equation*}
\ifhighlightrev
    \rev{
\fi
where $\CC\in\RR^3$ is the center of the ellipsoid, $A$ is a real positive definite $3\times3$ matrix characterizing its orientation and size, and $\X\in\RR^3$ is any point on it.
\ifhighlightrev
    }
\fi

Given a center of projection $\E\in\RR^3$ and a projection plane with normal $\N\in\RR^3$  which does not contain $\E$, the projection of the ellipsoid is an ellipse with center $\K\in\RR^3$ and semi-diameters $a,b\in\RR^{+}$. Its principal directions are represented by unit vectors $\U,\V\in\RR^3$ such that $\{\U,\V,\N\}$ is an orthonormal set.

\subsubsection{Projection Cone}

The {\it projection cone} $(\E,B)$ refers to the cone of vertex $\E$ tangent to the ellipsoid. According to \cite{Eberly-backproj}, it is characterized by matrix
\begin{equation*}
    B\overset{def}{=}A\D\D^\top A-(\D^\top A\D-1)A\mbox{,}
\end{equation*}
where $\D=\E-\CC\in\RR^3$ is the vector going from the ellipsoid center $\CC$ to the camera center $\E$. The points $\X\in\RR^3$ belonging to the projection cone are those who satisfy the Equation $(\X-\E)^\top B(\X-\E)=0$. Note that $B$ is a real, symmetric and invertible  $3\times3$ matrix which has two eigenvalues of the same sign and the third one of the opposite sign,
\ifhighlightrev
    \rev{
\fi
since a reduced-form cone equation is by definition
\begin{equation*}
    \frac{x^2}{x_0^2}+\frac{y^2}{y_0^2}-\frac{z^2}{z_0^2}=0\mbox{,}
\end{equation*}
where $x,y,z\in\RR$ are the 3D coordinates of any 3D point and $x_0,y_0,z_0\in\RR$ are characteristic dimensions of the cone.
\ifhighlightrev
    }
\fi

\subsubsection{Backprojection Cone}

The {\it backprojection cone} $(\E,B')$ refers to the cone generated by the lines passing through $\E$ and any point on the ellipse. Eberly shows that it is characterized by matrix
\begin{equation*}
    B'\overset{def}{=}\rev{S}^\top M\rev{S}-Q\mbox{,}
\end{equation*}
where
\begin{align*}
    M &\overset{def}{=} \U\U^\top/a^2+\V\V^\top/b^2\mbox{,}\\
    {\bf W} &\overset{def}{=} \N/(\N\cdot(\K-\E))\mbox{,}\\
    \rev{S} &\overset{def}{=} I-(\K-\E){\bf W}^\top\mbox{,}\\
    Q &\overset{def}{=} {\bf W}{\bf W}^\top\mbox{.}
\end{align*}
Here again, the points $\X$ on the backprojection cone are those who meet $(\X-\E)^\top B'(\X-\E)=0$, and $B'$ has the same properties as $B$ (\textit{i.e.} real, symmetric, invertible $3\times3$ matrix with signature (2,1) or (1,2)).

\subsubsection{The Cone Alignment Equation}

Given an ellipsoid, a central projection (center and plane) and an ellipse in the projection plane, the ellipse is the projection of the ellipsoid if and only if the projection and backprojection cones are aligned \citep{Eberly-backproj}, \textit{i.e.} if and only if there is a non-zero scalar $\s\in\RR\setminus\{0\}$ such that $B=\s B'$:

\noindent
\ifboxeseq
    \framebox{\parbox{\linewidth}{
\fi
\begin{equation} \label{eq:main_pb}
    A\D\D^\top A+\mu A=\s B'\mbox{,}
\end{equation}
where $\mu=1-\D^\top A \D\in\RR$.
\ifboxeseq
    }}
\fi

Note that $\mu$ encapsulates the relative configuration between the camera center $\E$ and the ellipsoid $(\CC,A)$. More precisely, it is negative if $\E$ is outside the ellipsoid, equal to zero if $\E$ belongs to the ellipsoid and positive otherwise. 

\ifhighlightrev
    \rev{
\fi
Equation \eqref{eq:main_pb} therefore encodes the P$1$E problem. In what follows, we refer to it as the {\it Cone Alignment Equation}.
\ifhighlightrev
    }
\fi
An equivalent formulation is given by Equation \eqref{eq:eq_pb} (see proof of equivalence in Appendix 
\ref{secA1}):

\noindent
\ifboxeseq
    \framebox{\parbox{\linewidth}{
\fi
\begin{equation} \label{eq:eq_pb} \tag{\ref{eq:main_pb}'}
    \D\D^\top = A^{-1}-\frac{\mu}{\s}B'^{-1}\mbox{.}
\end{equation}
\ifboxeseq
    }}
\fi

\subsection{P$1$E Problem Analysis}

\ifhighlightrev
    \rev{
\fi
\paragraph{Reference frames}
\ifhighlightrev
    }
\fi
While Equation \eqref{eq:main_pb} has been established in the camera coordinate frame, it is also valid in the ellipsoid coordinate frame, where matrix $A$ is diagonal, and in the cone coordinate frame where $B'$ is diagonal. Indeed, since $A$ and $B'$ are real and symmetric, both can be diagonalized using an orthogonal matrix. For that reason, Equation \eqref{eq:main_pb} remains the same whatever the choice of the coordinate frame (camera, ellipsoid or cone) in which matrices and vectors are expressed. In the following, if there is no restriction on the coordinate frame, we will adopt notations without subscript. Otherwise, subscripts ${cam}$, ${cone}$ or ${ell}$ will be used to specify the reference frame in which elements are expressed.

\ifhighlightrev
    \rev{
\fi
\paragraph{Camera or ellipsoid pose}
\ifhighlightrev
    }
\fi
One can theoretically distinguish between camera pose estimation, which consists in estimating the pose of the camera with respect to its environment, and its reference frame counterpart, \textit{i.e.} object pose estimation, which consists in estimating the pose of an object with respect to the camera. Fundamentally, the sought transformations are the inverse of each other thus, in this paper, we may focus on estimating the pose of the ellipsoid or the pose of the camera, according to the most convenient setup for mathematical developments, without loss of generality.

\ifhighlightrev
    \rev{
\fi
\paragraph{Knowns and unknowns}
\ifhighlightrev
    }
\fi
\ifhighlightrev
    \rev{
\fi
In the considered problem, the ellipse detected in the image, the camera intrinsic parameters and the ellipsoid size (\textit{i.e.} lengths of its three radii) are known.
\ifhighlightrev
    }
\fi
Therefore, $B_{cone}'$ (diagonal), $B_{cam}'$ and $A_{ell}$ (diagonal) are known, as well as $\E_{cone}=\E_{cam}=\CC_{ell}=\mathbf{0}$ (\textit{i.e.} the null vector, since these points are the origins of their corresponding reference frames). Since expressions of $A$ and $B'$ in specific coordinate frames are known, their eigenvalues are also known. In addition, matrix properties such as trace and determinant, that are fully constrained by the eigenvalues, are also known.

\ifhighlightrev
    \rev{
\fi
\paragraph{Overview of solutions computation}
\ifhighlightrev
    }
\fi
In the paper, we sometimes solve the P$1$E problem in the camera frame. We thus aim at computing vector $\D_{cam}=\E_{cam}-\CC_{cam}$ and matrix
\begin{align*}
    A_{cam} &= \Rcell A_{ell}\Rcell^\top\\
    &=\Rcell\begin{pmatrix}
    1/a^2 & 0 & 0\\
    0 & 1/b^2 & 0\\
    0 & 0 & 1/c^2
    \end{pmatrix}\Rcell^\top\mbox{,}
\end{align*}
from which we then retrieve the ellipsoid position $\CC_{cam}$ and orientation $\Rcell$.

In the other case, we aim at computing $\D_{ell}$ and $B_{ell}'$ to then derive the camera position $\E_{ell}$ and orientation $\Rcell^\top$.

It is important noting that vector $\D$ encodes the relative position between the ellipsoid and the camera, while couple \abb characterizes their relative orientation. In details, its expression in the camera frame is
\begin{equation*}
    \{\Rcell A_{ell}\Rcell^\top,B_{cam}'\}\mbox{,}
\end{equation*}
while its expression in the ellipsoid frame is
\begin{equation*}
    \{A_{ell},\Rcell^\top B_{cam}'\Rcell\}\mbox{,}
\end{equation*}
recalling that $A_{ell}$ and $B_{cam}'$ are known.

Solving the P$1$E problem therefore consists in solving Equation \eqref{eq:main_pb}, \textit{i.e.} determining the possible value(s) of $\s$ and expressions of $A,B',\D$ in a common coordinate frame (camera or ellipsoid).

\section{Properties of P$1$E Solutions}
\label{sec:properties}


In this section, we demonstrate some properties of the solutions of Equation \eqref{eq:main_pb}. 
Result \ref{res:GEVP} was presented in \cite{Eberly-backproj} and exhibits the link between \eqref{eq:main_pb} and a generalized eigenvalue problem. Result \ref{res:ABeig} was introduced for the first time in \cite{IROS}. Other results are novel. We especially show  how $\sigma$ and $\mu$ (scalar parameters of Equation \eqref{eq:main_pb}) can be derived from $A$ and $B'$ once their expressions in a common coordinate frame are known. These results are used in the following to solve for the camera solutions. It must be noted that the positive definite nature of matrix $A$, \textit{i.e.} what differentiates an ellipsoid from any other quadric, plays a fundamental role in the demonstrations of these properties.

\subsection{Link with a Generalized Eigenvalue Problem}
\label{ssec:prop-GEVP}

Let $(A,B',\D,\s)$ be a set of solutions of Equation \eqref{eq:main_pb}. Result \ref{res:GEVP} shows that they are also solutions of a generalized eigenvalue problem \citep{GoluVanl96}.

\noindent
\ifboxes
    \framebox{\parbox{\linewidth}{
\fi
\begin{res} \label{res:GEVP}
If $A,B',\D$ and $\s$ satisfy
\begin{equation} \tag{\ref{eq:main_pb}}
    A\D\D^\top A+(1-\D^\top A\D) A=\s B'\mbox{,}
\end{equation}
they also satisfy
\begin{equation} \label{eq:gevp}
    A\D=\s B'\D\mbox{.}
\end{equation}
\end{res}
\ifboxes
    }}
\fi

\begin{proof}
Let's right-multiply \eqref{eq:main_pb} by $\D$.
Since $\D^\top A\D$ is a scalar, the right hand term can be simplified:
\begin{align*}
    \s B'\D &= (A\D\D^\top A+(1-\D^\top A\D)A)\D\\
        &= (\D^\top A\D)A\D+A\D-\D^\top A\D A\D\\
        &= A\D\mbox{.}
\end{align*}
\end{proof}

Since $B'$ is invertible, the generalized eigenvectors and eigenvalues of pair \abb are the eigenvectors and eigenvalues of matrix $B'^{-1}A$.

\subsection{Generalized Eigenvalues of $\abb$ and characterisation of $\s$}
\label{ssec:prop-AB}

In this section, we go one step further and provide an explicit mean to identify $\s$ among the eigenvalues  of \ab:
\noindent
\ifboxes
    \framebox{\parbox{\linewidth}{
\fi
\begin{res} \label{res:ABeig}
The couple \abb has exactly two distinct generalized eigenvalues, $\s_1$  with multiplicity 1 and $\s_2$ with multiplicity 2, that are non-zero and of opposite signs.\\
$\s$ is the generalized eigenvalue with multiplicity 1:
\begin{equation} \label{eq:A2sigma}
  \s=\s_1
\end{equation}
\end{res}
\ifboxes
    }}
\fi

The proof is given in appendix \ref{appendix:ABeig}.

\ifhighlightrev
    \rev{
\fi
A direct application of Result \ref{res:ABeig} is considered in Section \ref{ssec:pos} in the particular case where the relative camera-ellipsoid orientation is known. In that case, $\s$ and $\D$ can be uniquely inferred from \abb. In the general P$1$E problem, we  additionally  use the characterization of $\mu$ as a function of \abb eigenvalues, presented in the next section,  to infer the complete set  of solutions.
\ifhighlightrev
    }
\fi


\subsection{Characterizations of $\mu$}
\label{ssec:prop-mu}

\ifhighlightrev
    \rev{
\fi
We provide in this section two expressions of the secondary scalar variable $\mu=1-\D^\top A\D$. The first one (Result \ref{res:mu}) shows that $\mu$ can be expressed as the ratio of the two \abb generalized eigenvalues, while Result \ref{res:Delta2mu} shows that  $\mu$ can be expressed uniquely as a function of $\s$. Indeed, even if $A$ and $B'$ expressions are not known in a common coordinate frame, their determinants can be computed, thus allowing to define directly $\mu$ as a function of $\s$. The demonstration of these results are provided in Appendix \ref{appendix:mu}.
\ifhighlightrev
    }
\fi
 
\noindent
\ifboxes
    \framebox{\parbox{\linewidth}{
\fi
\begin{res} \label{res:mu}
$\mu$ is equal to the ratio between the two generalized eigenvalues of \ab:
    \begin{equation}\label{eq:A2mu}
        \mu=\frac{\s_1}{\s_2}\mbox{.}
    \end{equation}
\end{res}
\ifboxes
    }}
\fi

Since $\s_1$ and $\s_2$ are of opposite signs, Equation \eqref{eq:A2mu} shows that $\mu<0$.

Result \ref{res:Delta2mu} now highlights the connection between $\mu$ and $\s$.

\noindent
\ifboxes
    \framebox{\parbox{\linewidth}{
\fi
\begin{res} \label{res:Delta2mu}
$\mu$ and $\s$ are linked through Equation \eqref{eq:Delta2mu}:
\begin{equation}\label{eq:Delta2mu}
    \mu=-\sqrt{\frac{det(B')}{det(A)}\s^3}\mbox{.}
\end{equation}
\end{res}
\ifboxes
    }}
\fi

\subsection{Link between $\s$ and $\|\D\|$}
\label{ssec:prop-sD}

\noindent
\ifboxes
    \framebox{\parbox{\linewidth}{
\fi
\begin{res} \label{res:Link-sD}
    The scalar $\s$ and the camera-ellipsoid distance $\|\D\|$ are linked through Equation \eqref{eq:Delta2sigma}:
    \begin{equation}\label{eq:Delta2sigma}
   \left( tr(B'^{-1})\right)^2\s=\frac{det(A)}{det(B')}\left(tr(A^{-1})-\|\D\|^2\right)^2\mbox{.}
    \end{equation}
\end{res}
\ifboxes
    }}
\fi

\begin{proof}
Injecting Equations \eqref{eq:A2sigma} and \eqref{eq:A2mu} into \eqref{eq:eq_pb} leads to
\begin{equation*}
    \D\D^\top=A^{-1}-\frac{1}{\s_2}B'^{-1}\mbox{.}
\end{equation*}
Applying $tr()$ then squaring:
\begin{equation}\label{eq:res5-1}
    \frac{1}{\s_2^2}\left(tr(B'^{-1})\right)^2=\left(tr(A^{-1})-\|\D\|^2\right)^2\mbox{.}
\end{equation}

Furthermore, injecting \eqref{eq:A2sigma} into \eqref{eq:detABs1s2} leads to the following expression for $\frac{1}{\s_2^2}$:
\begin{equation}
    \frac{1}{\s_2^2}=\frac{det(B')}{det(A)}\s\mbox{.}
    \label{eq:sigma2}
\end{equation}

Equation \eqref{eq:Delta2sigma} is then obtained by injecting \eqref{eq:sigma2} into \eqref{eq:res5-1}.
\end{proof}

\section{Decoupling between Orientation and Position}
\label{sec:decoupl}

In this section, we consider two sub-problems of significant practical interest: (i) computing the relative camera-ellipsoid position when the orientation is known and (ii) computing the orientation when the position is known.
We demonstrate that the position can be inferred in closed-form from the orientation (Section \ref{ssec:pos}, Result \ref{res:A2Delta}), while the latter can be analytically derived from the former up to the ellipsoid symmetries (Section \ref{ssec:ori}, Result \ref{res:Delta2A}). We assimilate these properties to a {\it decoupling} phenomenon between orientation and position.

\subsection{Position from Orientation}
\label{ssec:pos}
In this case, $A_{cam}$ or $B_{ell}'$ is known. Since $A_{ell}$ and $B_{cam}'$ are always known, eigenvalues of \ab, $\sigma$ (using Result \ref{res:ABeig}) and $\mu$ (Result \ref{res:mu}) can be retrieved.
Result $\ref{res:A2Delta}$ then provides that $\D$ is unique and fully determined.

\noindent
\ifboxes
    \framebox{\parbox{\linewidth}{
\fi

\begin{res} \label{res:A2Delta}
Assuming that the relative camera-ellipsoid orientation is known, their relative position is given by
\begin{equation}\label{eq:A2Delta}
    \D=k\de_1\mbox{,}
\end{equation}
where
\begin{equation*}
    \left\{
        \begin{array}{l}
            \mbox{$\de_1$ is a unit generalized eigenvector of \ab}\\\hspace{0.5cm}\mbox{corresponding to $\s_1$,}\\
            \de_1\cdot\N<0,\\
            k=\sqrt{tr(A^{-1})-\frac{1}{\s_2}tr(B'^{-1})}\mbox{.}
        \end{array}
    \right.
\end{equation*}
\end{res}
\ifboxes
    }}
\fi

\begin{proof}

Injecting Equations \eqref{eq:A2sigma}, \eqref{eq:A2mu} and $\D=k\de_1$ into Equation \eqref{eq:eq_pb} leads to
\begin{equation*}
    k^2\de_1\de_1^\top=A^{-1}-\frac{1}{\s_2}B'^{-1}\mbox{.}
\end{equation*}
Whence, by applying $tr()$,
\begin{equation*}
    k^2\|\de_1\|^2=k^2=tr(A^{-1})-\frac{1}{\s_2}tr(B'^{-1})\mbox{.}
\end{equation*}

The two vectors $\pm\sqrt{k^2}\de_1$ define ellipsoid centers $\CC$ that are symetric with respect to the camera center $\E$. The only one that satisfy the chirality constraint (ellipsoid located in front of the camera) is the one whose dot product with vector $\N$ is negative (see Fig. \ref{fig:schema}).
\end{proof}

Result \ref{res:A2Delta} highlights the fact that solving Equation \eqref{eq:main_pb} may consist only in determining $A_{cam}$ or $B_{ell}'$. Indeed, $\s$ and $\D$ can then be derived uniquely. In other words, the relative position is fully constrained by the orientation.

\ifhighlightrev
    \rev{
\fi
The main steps allowing to retrieve the camera position in the ellipsoid frame or equivalently the ellipsoid centre in the camera frame are summarized in Algorithms \ref{alg:ori2pos-ecf} and \ref{alg:ori2pos-ccf}.
\ifhighlightrev
    }
\fi

\begin{algorithm}
\caption{Position from orientation (camera centre expressed in the ellipsoid frame)}
\label{alg:ori2pos-ecf}
\begin{algorithmic}
\State $B_{ell}' \gets \Rcell^\top B_{cam}'\Rcell$
\State $\s_1 \gets$ generalized eigenvalue of $\{A_{ell},B_{ell}'\}$ with multiplicity 1
\State $\de_1 \gets$ generalized eigenvector of $\{A_{ell},B_{ell}'\}$ associated to $\s_1$
\State $k \gets \sqrt{tr(A_{ell}^{-1})-\frac{1}{\s_2}tr(B_{ell}'^{-1})}$
\State $\D_{ell} \gets k\de_1$
\State $\E_{ell} \gets \D_{ell}+\CC_{ell}$
\end{algorithmic}
\end{algorithm}
\begin{algorithm}
\caption{Position from orientation (ellipsoid centre expressed in the camera frame)}
\label{alg:ori2pos-ccf}
\begin{algorithmic}
\State $A_{cam} \gets \Rcell A_{ell}\Rcell^\top$
\State $\s_1 \gets$ generalized eigenvalue of $\{A_{cam},B_{cam}'\}$ with multiplicity 1
\State $\de_1 \gets$ generalized eigenvector of $\{A_{cam},B_{cam}'\}$ associated to $\s_1$
\State $k \gets \sqrt{tr(A_{cam}^{-1})-\frac{1}{\s_2}tr(B_{cam}'^{-1})}$
\State $\D_{cam} \gets k\de_1$
\State $\CC_{cam} \gets \E_{cam}-\D_{cam}$
\end{algorithmic}
\end{algorithm}

As mentioned above, this result is of high practical interest. Indeed, getting the camera orientation, e.g. from physical sensors or image analysis, is usually easier than getting the camera position, and especially indoors where the GPS is unusable. In a multi-object scene, the fact that only one ellipse-ellipsoid association is needed to compute the pose allows using a RANSAC-like strategy with low combinatorial cost to match object detections with their corresponding 3D primitives.

\begin{figure*}[h!]
    \centering
    \begin{tabular}{@{\hspace{0mm}}c@{\hspace{0mm}}c@{\hspace{0mm}}c@{\hspace{0mm}}c@{\hspace{0mm}}c@{\hspace{0mm}}}
    \includegraphics[width=0.19\linewidth,height=0.14\linewidth,trim=22mm 23mm 37mm 11mm,clip]{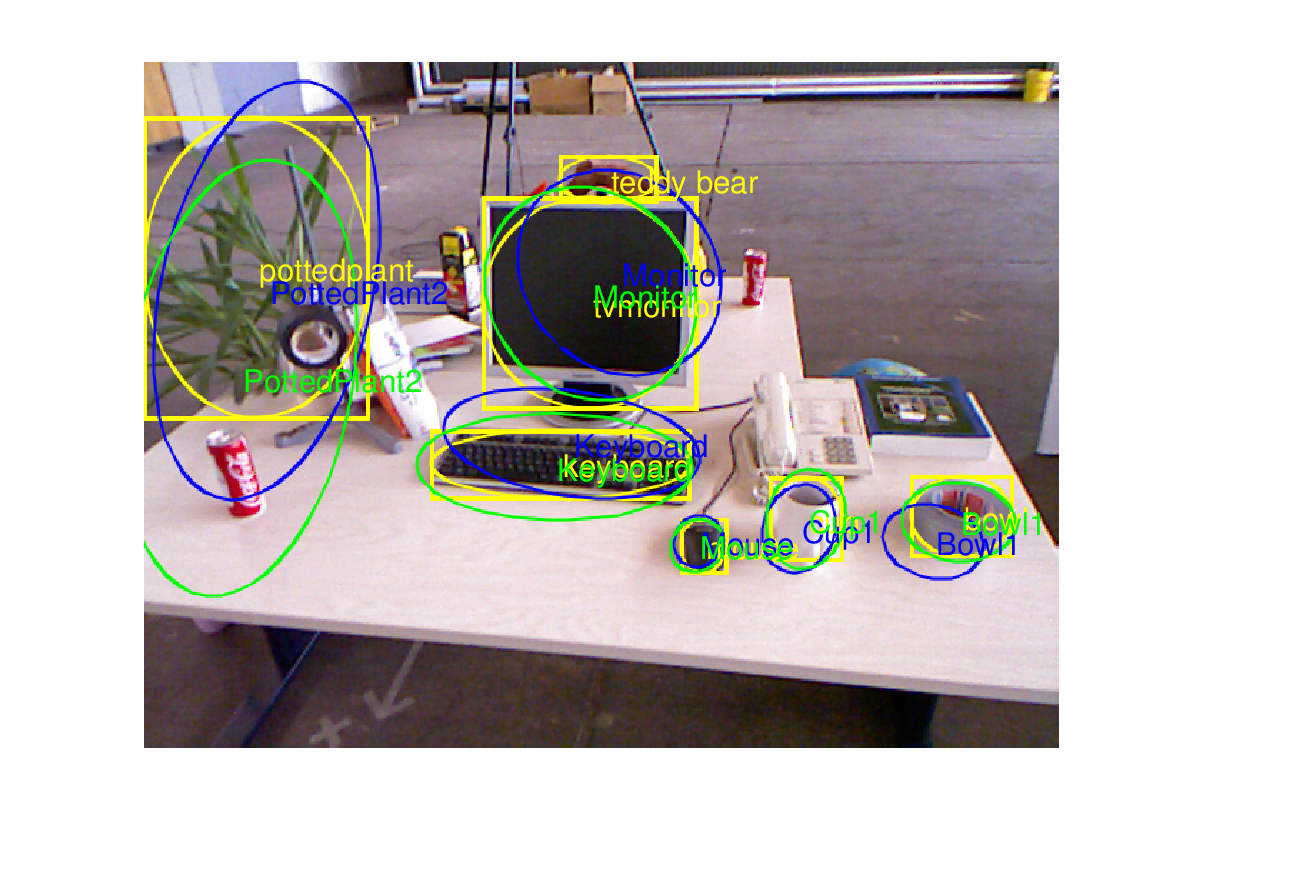}&
    \includegraphics[width=0.19\linewidth,height=0.14\linewidth,trim=22mm 23mm 37mm 11mm,clip]{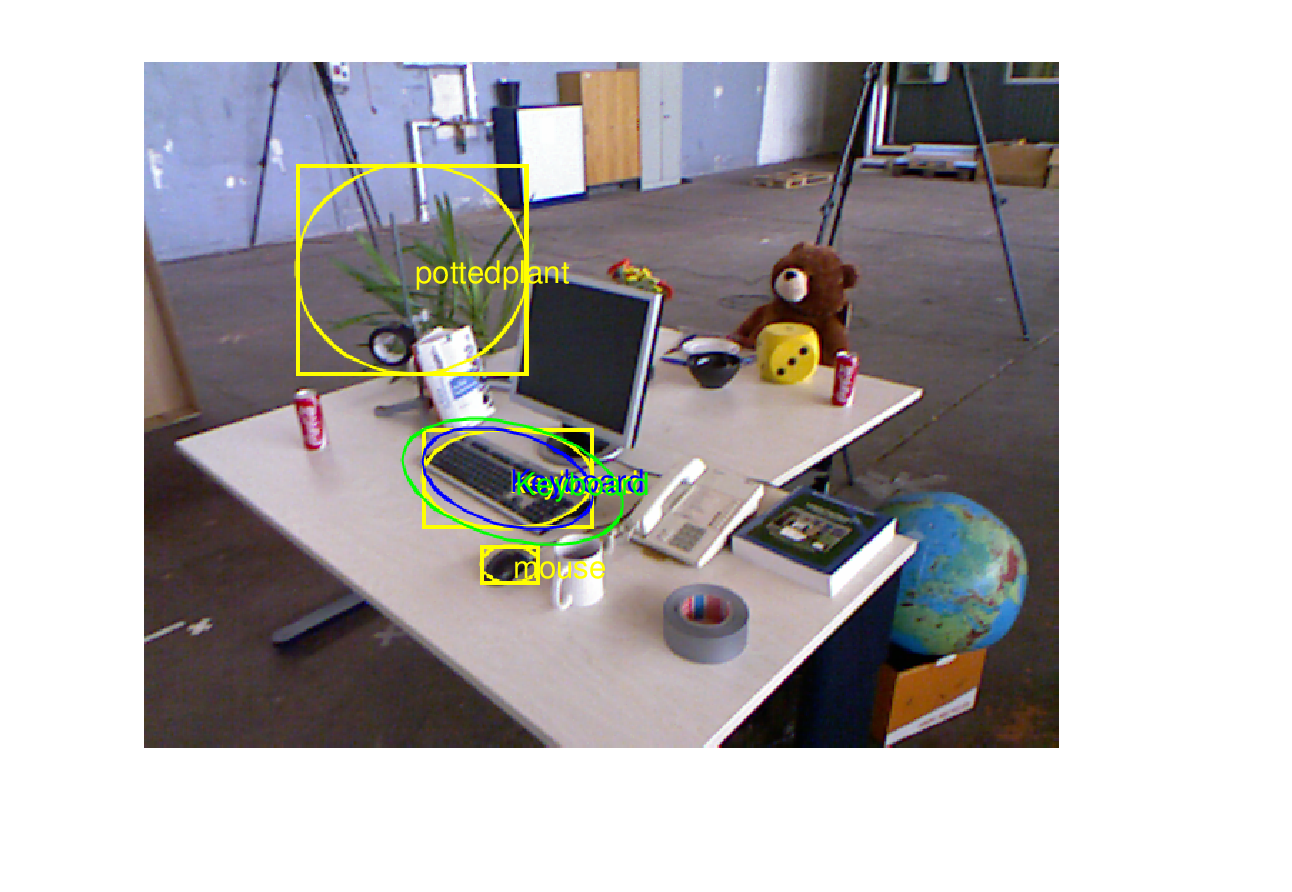}&
    \includegraphics[width=0.19\linewidth,height=0.14\linewidth,trim=22mm 23mm 37mm 11mm,clip]{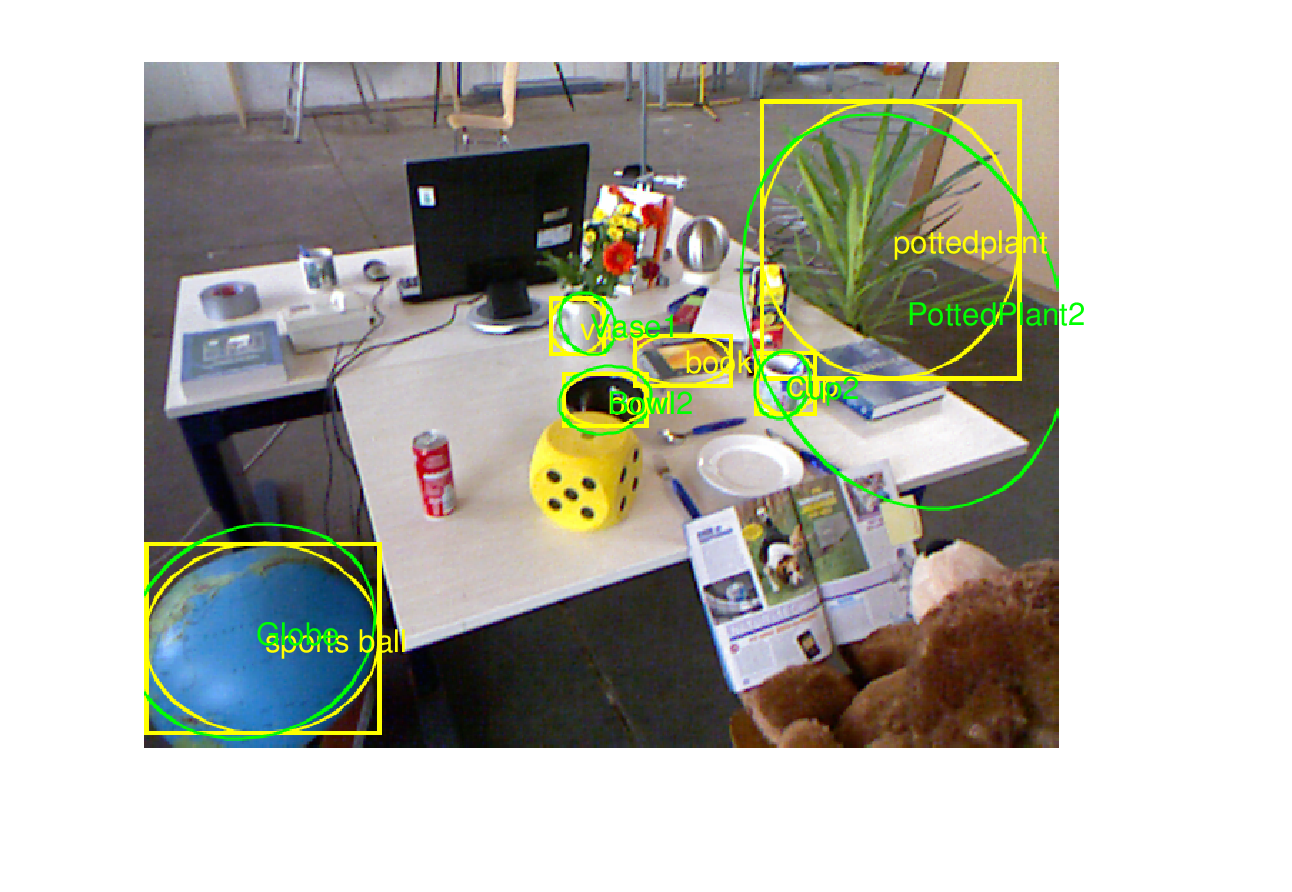}&
    \includegraphics[width=0.19\linewidth,height=0.14\linewidth,trim=22mm 23mm 37mm 11mm,clip]{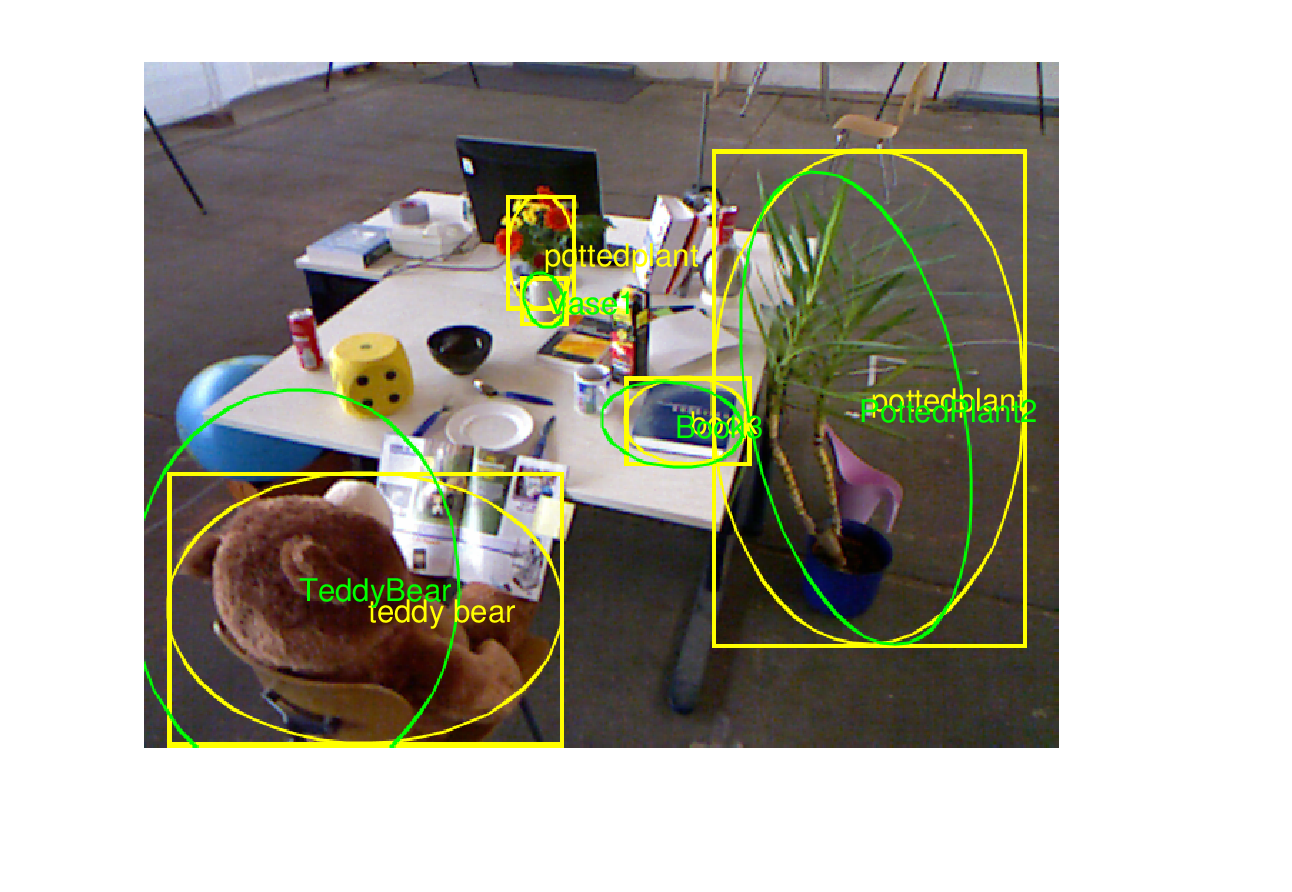}&
    \includegraphics[width=0.19\linewidth,height=0.14\linewidth,trim=22mm 23mm 37mm 11mm,clip]{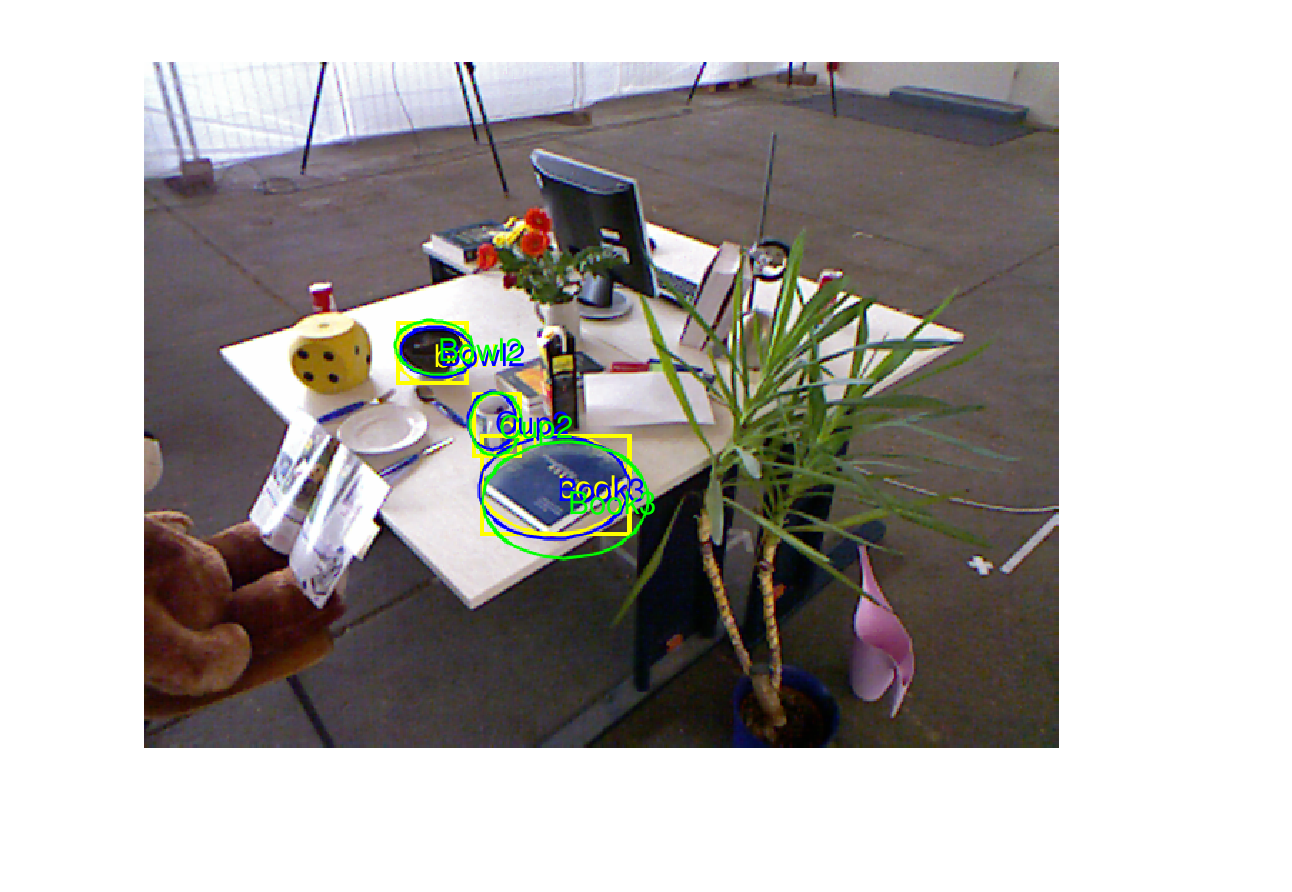}\\
    \includegraphics[width=0.19\linewidth,height=0.14\linewidth,trim=44mm 30mm 35mm 17mm,clip]{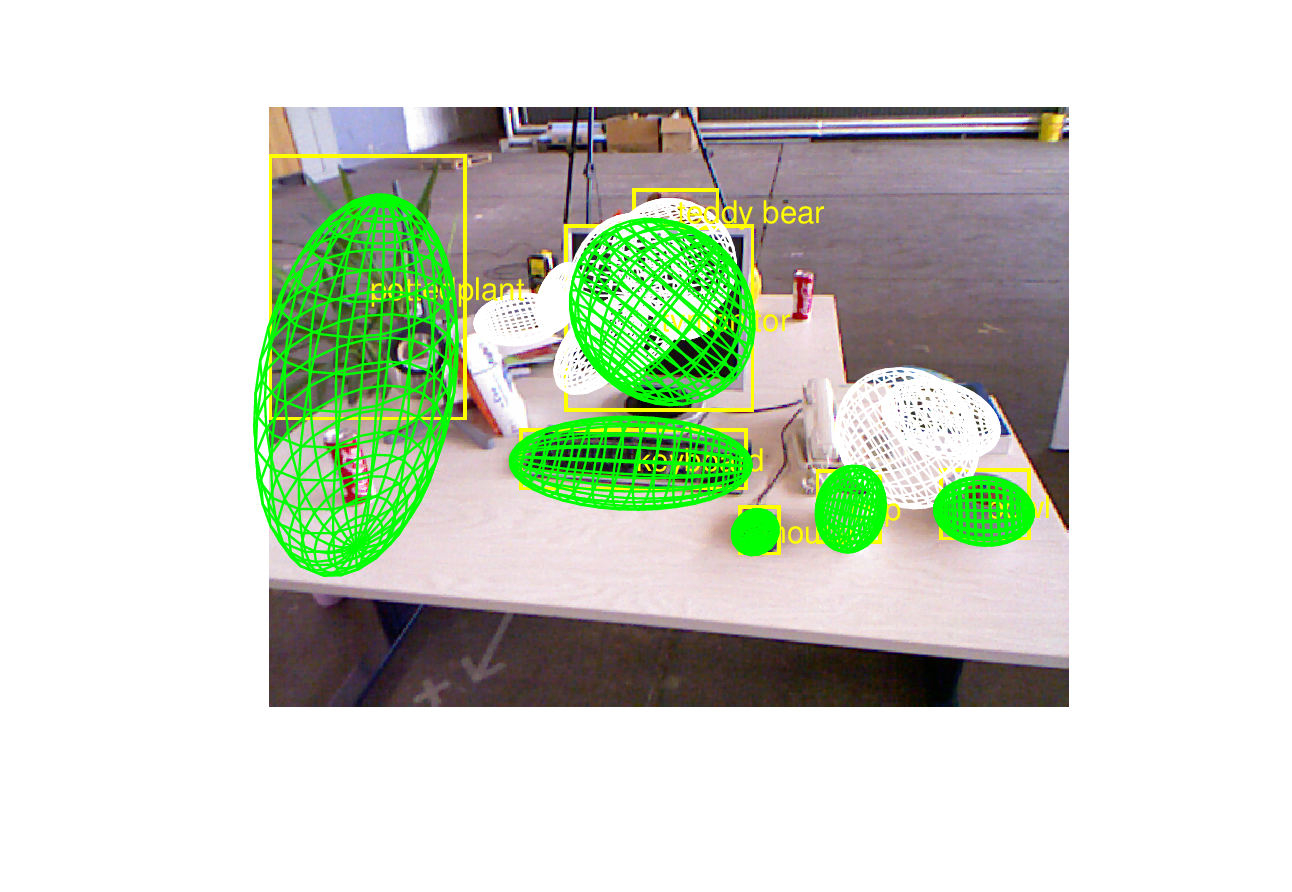}&
    \includegraphics[width=0.19\linewidth,height=0.14\linewidth,trim=23mm 29.5mm 57mm 16mm,clip]{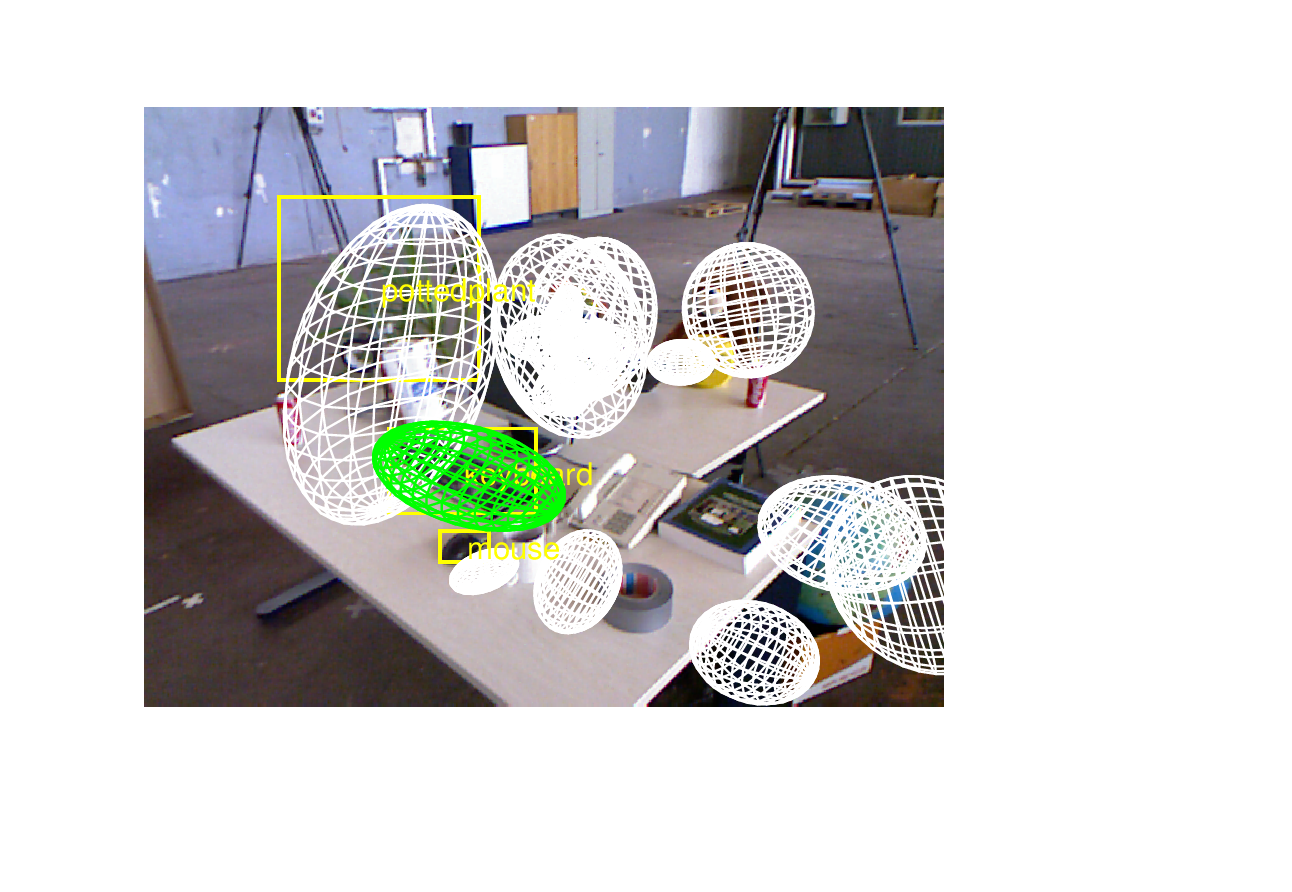}&
    \includegraphics[width=0.19\linewidth,height=0.14\linewidth,trim=40mm 48mm 67mm 18.5mm,clip]{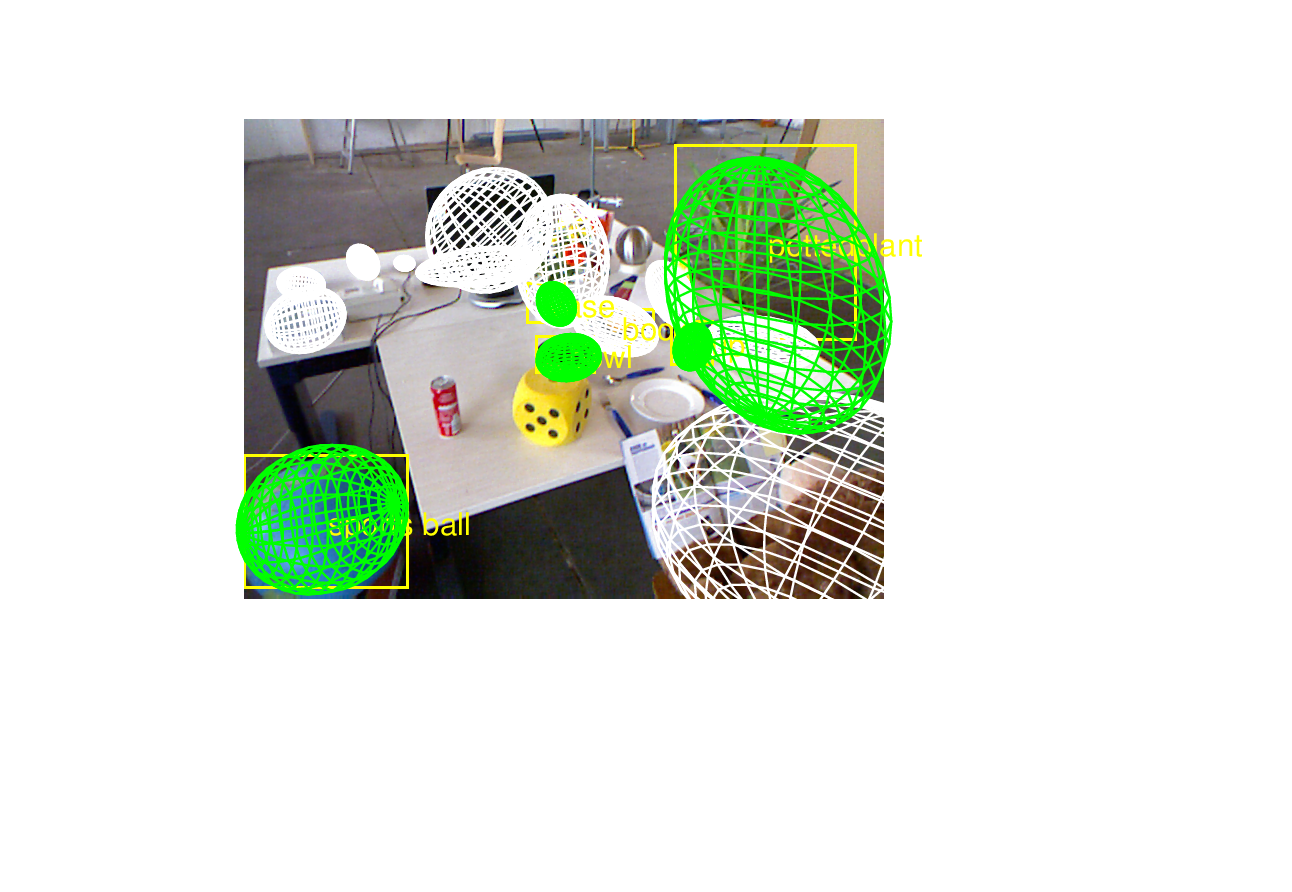}&
    \includegraphics[width=0.19\linewidth,height=0.14\linewidth,trim=43mm 41mm 36mm 6mm,clip]{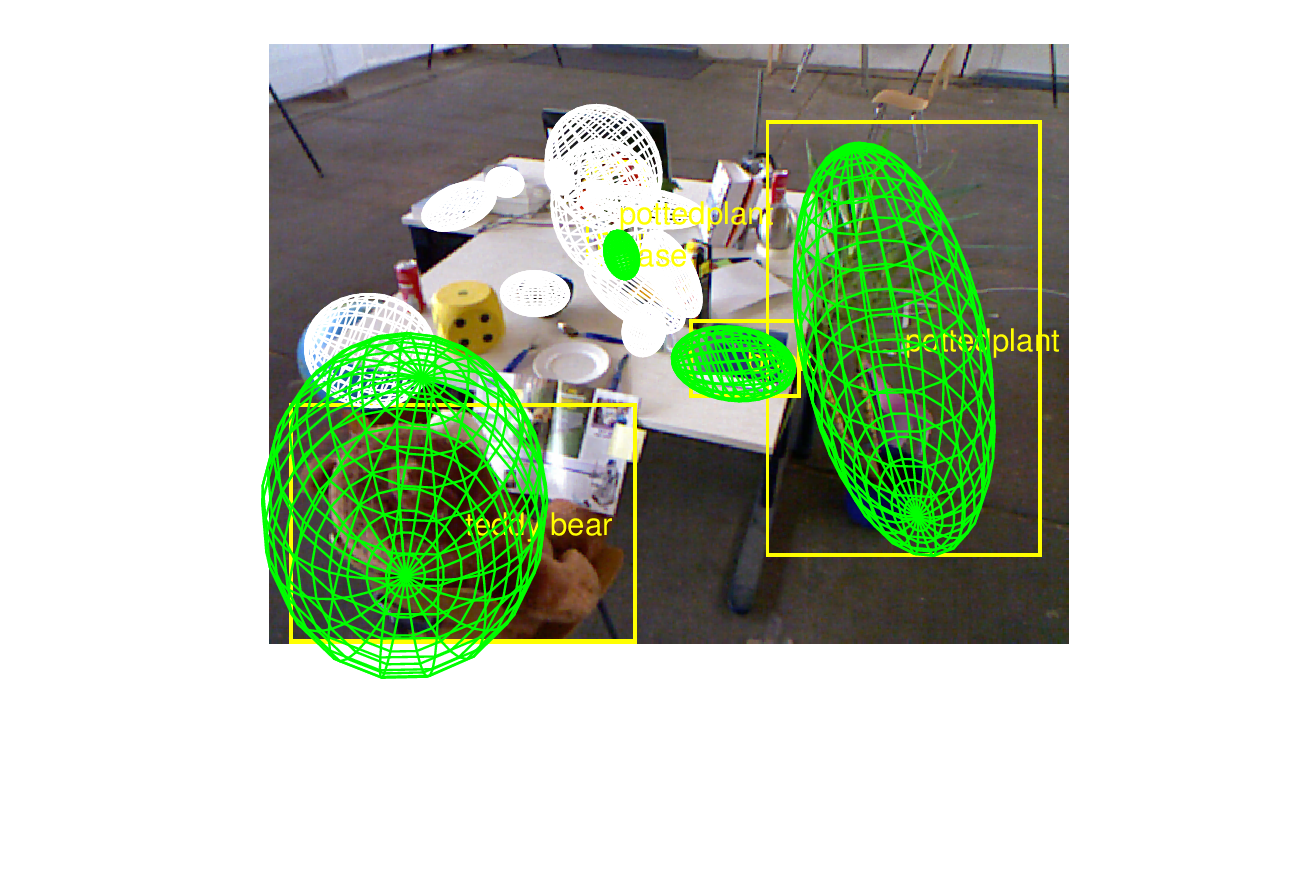}&
    \includegraphics[width=0.19\linewidth,height=0.14\linewidth,trim=60mm 45mm 34mm 12mm,clip]{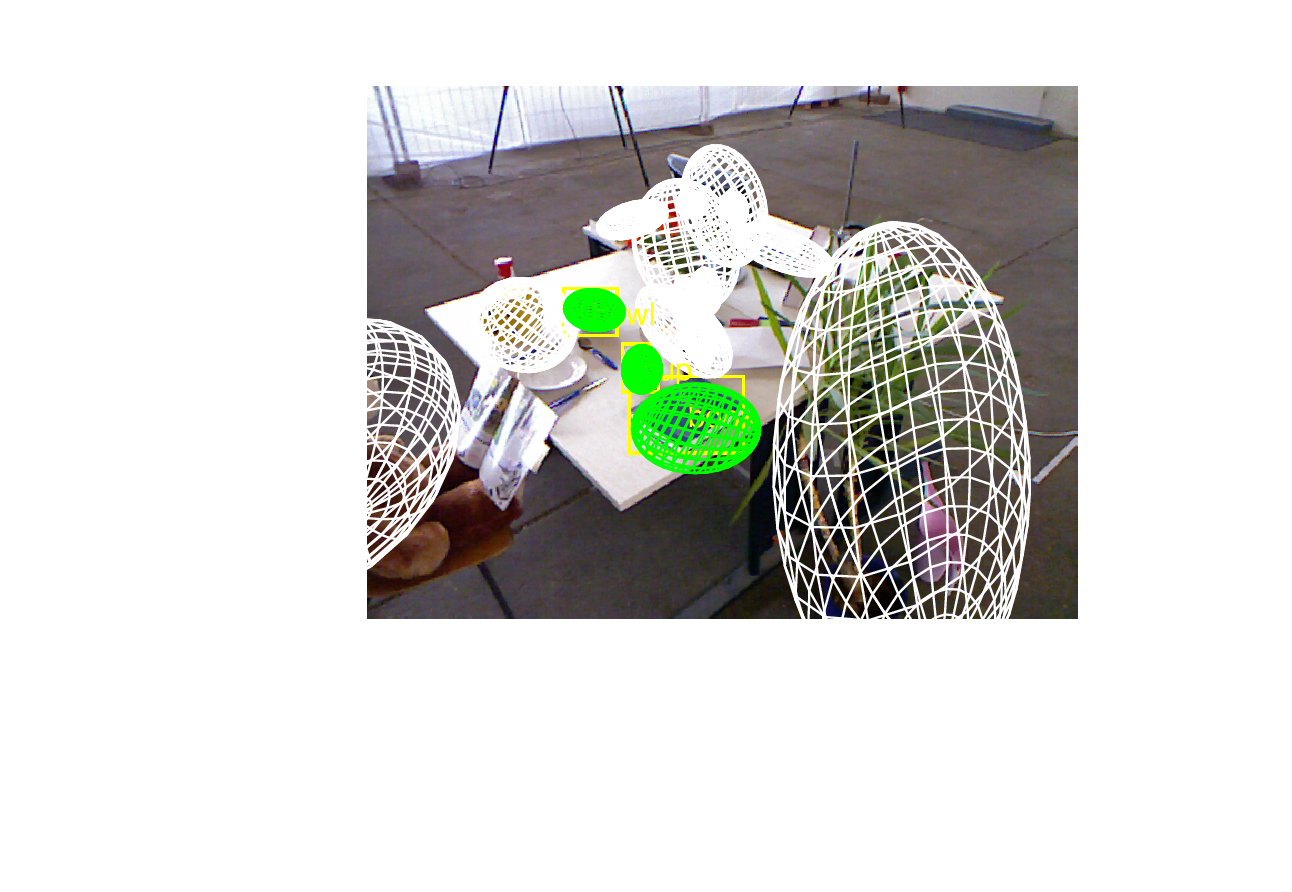}
    \end{tabular}
    \caption{Camera relocalization based on the decoupling between orientation and position, here applied  to images from the RGB-D TUM dataset \citep{ISMAR}. The first row shows the detected boxes, the inscribed ellipsoids (in yellow) and the outlines of the reprojected ellipsoids (in green), with the automatically generated labels. The blue ellipses correspond to the reprojected ellipsoids when the QuadricSLAM residual error is minimized \citep{QSLAM}. The second row shows the reprojected ellipsoids (in green the inliers, in white the outliers and undetected ellipsoids) when using the method in \citep{ISMAR}.}
    \label{fig:ismar}
\end{figure*}

A re-localization algorithm based on Result \ref{res:A2Delta} and leveraging this strategy was presented in \citep{ISMAR}. The system operates in real time from YOLO detections \citep{YOLOv3} and IMU data or vanishing points -- both methods were assessed. Figure \ref{fig:ismar} shows a few qualitative results obtained with images from the standard RGB-D TUM dataset \citep{SturmEEBC12}. Quantitative results as well as a detailed analysis of the advantages and limitations of this algorithm can be found in \citep{ISMAR}. Other applications of Result \ref{res:A2Delta} are presented in \citep{IROS,IROS2,RAL}.

\subsection{Orientation from Position}
\label{ssec:ori}

In this case $\D_{cam}$ or $\D_{ell}$ is known. $A_{cam}$ and $B_{ell}'$ are unknown, but since their eigenvalues are known, $det(A)$, $tr(A^{-1})$, $det(B')$ and $tr(B'^{-1})$ are known. $\s$ can thus be computed from Result \ref{res:Link-sD} and $\mu$ from Result \ref{res:Delta2mu}. Result \ref{res:Delta2A} then describes how to retrieve $A_{cam}$ or $B_{ell}'$ and how to derive the relative camera-ellipsoid orientation in closed-form, up to the cone or ellipsoid symmetries.

\noindent
\ifboxes
    \framebox{\parbox{\linewidth}{
\fi
\begin{res} \label{res:Delta2A}
Assuming that the relative camera-ellipsoid position is known, their relative orientation is given by eigenvectors of
\begin{equation}\label{eq:Delta2B}
    B_{ell}'=\frac{1}{\s}\left(A_{ell}\D_{ell}\D_{ell}^\top A_{ell}+\mu A_{ell}\right)
\end{equation}
in the ellipsoid reference frame, or of
\begin{equation}\label{eq:Delta2A}
    A_{cam}=\frac{\s}{\mu}\left(B_{cam}'-\s B_{cam}'\D_{cam}\D_{cam}^\top B_{cam}'\right)
\end{equation}
in the camera reference frame.
\end{res}
\ifboxes
    }}
\fi

\begin{proof}
Equation \eqref{eq:Delta2B} is \eqref{eq:main_pb} expressed in the ellipsoid frame.

Injecting \eqref{eq:gevp} into \eqref{eq:main_pb} leads to
\begin{equation*}
    \s^2B'\D\D^\top B' +\mu A=\s B'\mbox{,}
\end{equation*}
whence \eqref{eq:Delta2A} by isolating $A$.
\end{proof}

Once for instance $A_{cam}$ is retrieved, one can compute its eigenvalue decomposition:
\begin{equation*}
    A_{cam}=\Rcell\begin{pmatrix}
    1/a^2 & 0 & 0 \\
    0 & 1/b^2 & 0 \\
    0 & 0 & 1/c^2
    \end{pmatrix}\Rcell^\top\mbox{.}
\end{equation*}

For a triaxial ellipsoid ($a\ne b\ne c$), $\Rcell$ has 4 solutions.
\ifhighlightrev
    \rev{
\fi
For a spheroid (\textit{e.g.}, $a=b\ne c$), any rotation preserving the symmetry axis (\textit{e.g.}, $z$-axis) is solution. For a sphere ($a=b=c$), any rotation is solution.
\ifhighlightrev
    }
\fi
Therefore, the ellipsoid orientation can be analytically derived from its position up to the ellipsoid symmetries

\ifhighlightrev
   \rev{
\fi
The main steps allowing to retrieve the camera orientation are summarized in algorithms \ref{alg:pos2ori-ccf} and \ref{alg:pos2ori-ecf} .
\ifhighlightrev
   }
\fi

\begin{algorithm}
\caption{Orientation from position (Camera frame)}
\label{alg:pos2ori-ccf}
\begin{algorithmic}
\State $\D_{cam} \gets \E_{cam}-\CC_{cam}$
\State $\s \gets$ Equation \eqref{eq:Delta2sigma}
\State $\mu \gets$ Equation \eqref{eq:Delta2mu}
\State $A_{cam} \gets$ Equation \eqref{eq:Delta2A}
\State $\Rcell \gets$ eigenvectors of $A_{cam}$
\end{algorithmic}
\end{algorithm}

\begin{algorithm}
\caption{Orientation from position (Ellipsoid frame)}
\label{alg:pos2ori-ecf}
\begin{algorithmic}
\State $\D_{ell} \gets \E_{ell}-\CC_{ell}$
\State $\s \gets$ Equation \eqref{eq:Delta2sigma}
\State $\mu \gets$ Equation \eqref{eq:Delta2mu}
\State $B_{ell}' \gets$ Equation \eqref{eq:Delta2B}
\State $\Rellcone \gets$ eigenvectors of $B_{ell}'$
\State $\Rcell \gets \Rccone\Rellcone^\top$
\end{algorithmic}
\end{algorithm}

\section{Complete Set of P$1$E Solutions}
\label{sec:closedform}

In this section, we introduce the core contribution of the paper, that is closed-form solutions to the general P$1$E problem. Analytical 1DoF solutions are provided based on the fact that the ellipsoid is triaxial (all $A$ eigenvalues are distinct) or not, and the cone is circular (two $B'$ eigenvalues are equal) or not.

In Section \ref{ssec:cooc}, we first present the different types of ellipsoids and cones along with their possible co-occurences.  An overview of the solutions is given in Section \ref{sec:overview}. In Section \ref{ssec:ellipsoid}, we consider the case of a triaxial ellipsoid and present for the first time a \textit{Necessary and Sufficient Condition} (NSC) on $\s$ to be solution of Equation \eqref{eq:main_pb}. Then we derive the analytical expressions of the other variables as functions of $\s$. In Section \ref{ssec:spheroid}, we address the case of the spheroid (ellipsoid with an axis of revolution). That part enables to retrieve, from another formalism, the results presented in \cite{WokesP10}. In Section \ref{sec:sphere}, we finally present the solutions for the sphere.

In what follows, the problem is solved either in the ellipsoid or in the cone coordinate frame. In brief, the choice is linked to the ability to define a frame associated to the considered structure without ambiguities. The case of the triaxial ellipsoid can thus be addressed in both frames since the two structures are  unambiguous. The case of the spheroid is different, and depending on the properties of the cone, solutions are derived in one or the other frame.
\ifhighlightrev
    \rev{
\fi
To improve the readability of the paper, each considered case ends with a summary of the corresponding algorithm for pose recovery.
\ifhighlightrev
    }
\fi

\subsection{Preliminaries: Co-occurences of Ellipsoid and Cone Types}
\label{ssec:cooc}

In this paper, we address the full ellipsoid pose estimation problem, \textit{i.e.} we cover every possible types of ellipsoids and thus cones (see Appendix \ref{secAtypes}). However, a specific type of ellipsoid cannot necessarily be tangent to any type of cone, what we refer to as possible or impossible \textit{co-occurence}.

In brief, Table \ref{tab:cooccurence} summarizes the possible and impossible co-occurences between ellipsoid and cone types.
\ifhighlightrev
    \rev{
\fi
The proofs are provided in Appendix \ref{secA_cooc}.
\ifhighlightrev
    }
\fi
\begin{table*}[ht]
    \centering
    \begin{tabular}{cc|ccc}
         \multicolumn{2}{c}{} & \multicolumn{3}{|c}{Ellipsoid type} \\
         \multicolumn{2}{c|}{} & Triaxial & Spheroid & Sphere \\
         \hline
         \multirow{4}{*}{Projection cone type} & \multirow{2}{*}{Non-circular} & \checkmark & \checkmark & $\times$ \\
         & & \textit{(Algo. \ref{alg:overall-triaxial},\ref{alg:overall-noncirc})} & \textit{(Algo. \ref{alg:overall-noncirc})} \\
         & \multirow{2}{*}{Circular} & $\times$ & \checkmark & \checkmark \\
         & & & \textit{(Algo. \ref{alg:overall-spheroid-circ})} & \textit{(Algo. \ref{alg:overall-sphere})}
    \end{tabular}
    \vspace{0.1cm}
    \caption{Possible co-occurences of ellipsoids and projection cones according to their types. \checkmark$ $ indicates that ellipsoid and cone of the corresponding types may occur simultaneously. $\times$ indicates that they cannot.}
    \label{tab:cooccurence}
\end{table*}

\subsection{Overview  of the Solutions}
\label{sec:overview}

\begin{figure}
  \centering
  \includegraphics[trim=180mm 20mm 165mm 20mm, clip, width=0.9\linewidth]{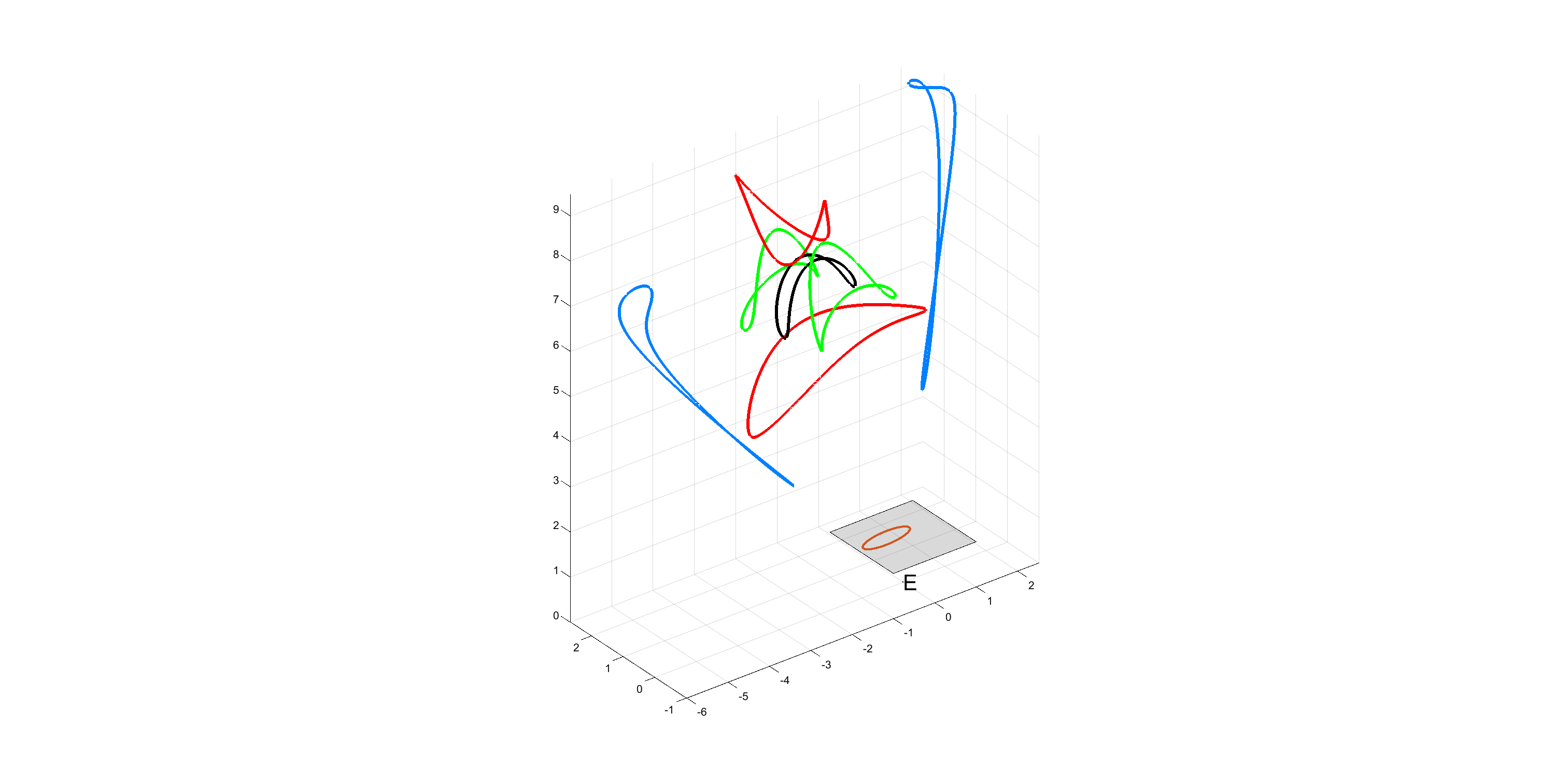}
\caption{Loci of the centers (black) and principal axes endpoints (red, green, blue) of the ellipsoids solutions. The camera center $\E$ appears along with the image plane (grey) and ellipse (orange). A video is available \protect\footnotemark.}
\label{fig:ellipsoids}
\end{figure}

\begin{figure}
  \centering
  \includegraphics[trim=165mm 20mm 150mm 20mm, clip, width=0.9\linewidth]{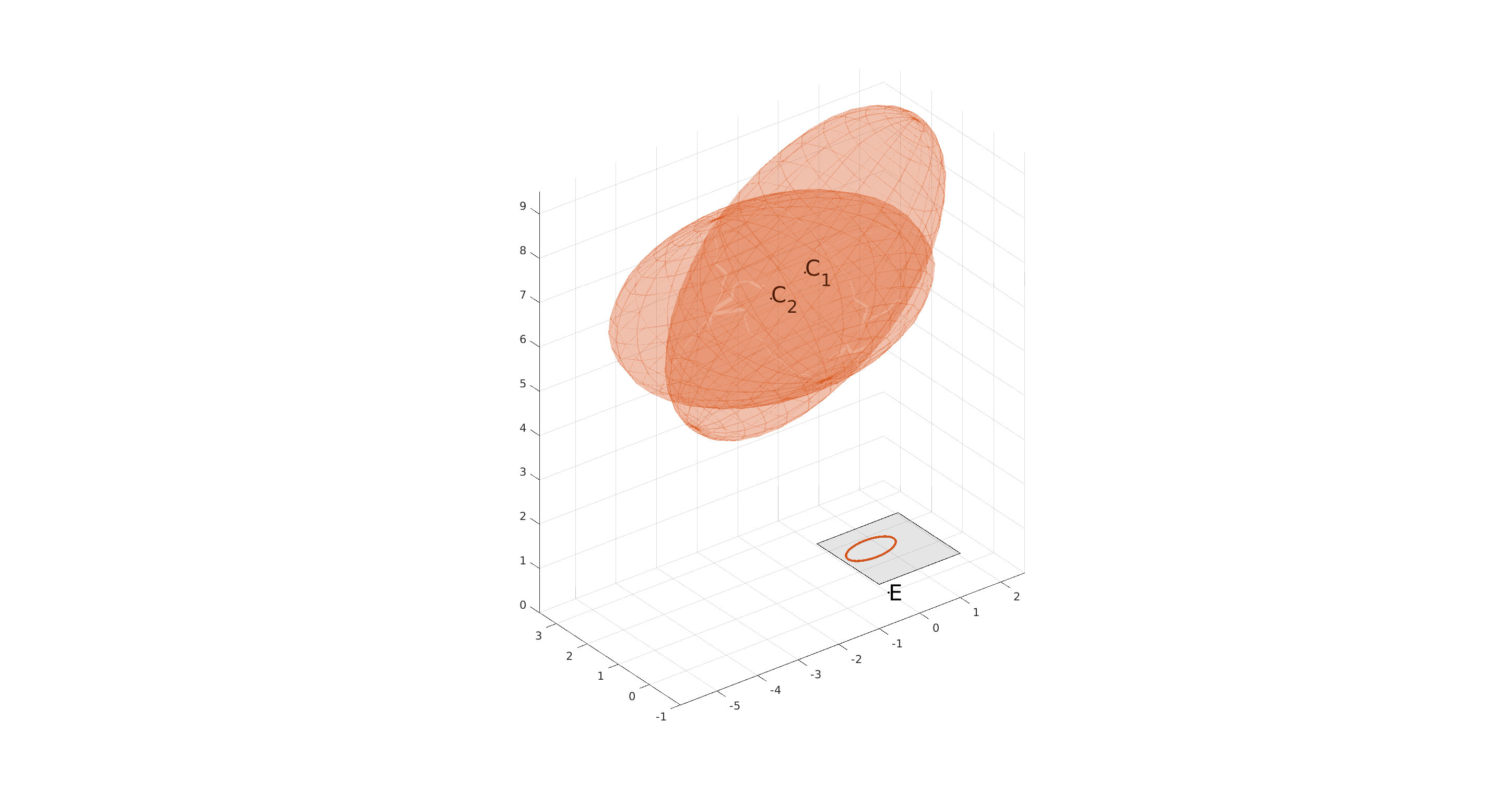}
\caption{The two spheroids solutions with a non-circular backprojection cone.}
\label{fig:spheroids}
\end{figure}

In the rest of this section, we determine the solutions of the Cone Alignment Equation \eqref{eq:main_pb} and derive the camera-ellipsoid relative poses. To this end, we distinguish between the three different types of ellipsoids (triaxial, spheroid, sphere). We demonstrate, in particular, that
\begin{itemize}
    \item there is an infinite number of triaxial fixed-size ellipsoids that are tangent to a given backprojection cone (Fig. \ref{fig:ellipsoids}),
    \item as already demonstrated in \cite{WokesP10}, there are only two fixed-size spheroids solutions (see Fig. \ref{fig:spheroids}).
\end{itemize} 

\begin{figure}
  \centering
  \includegraphics[trim=165mm 20mm 150mm 20mm, clip, width=0.9\linewidth]{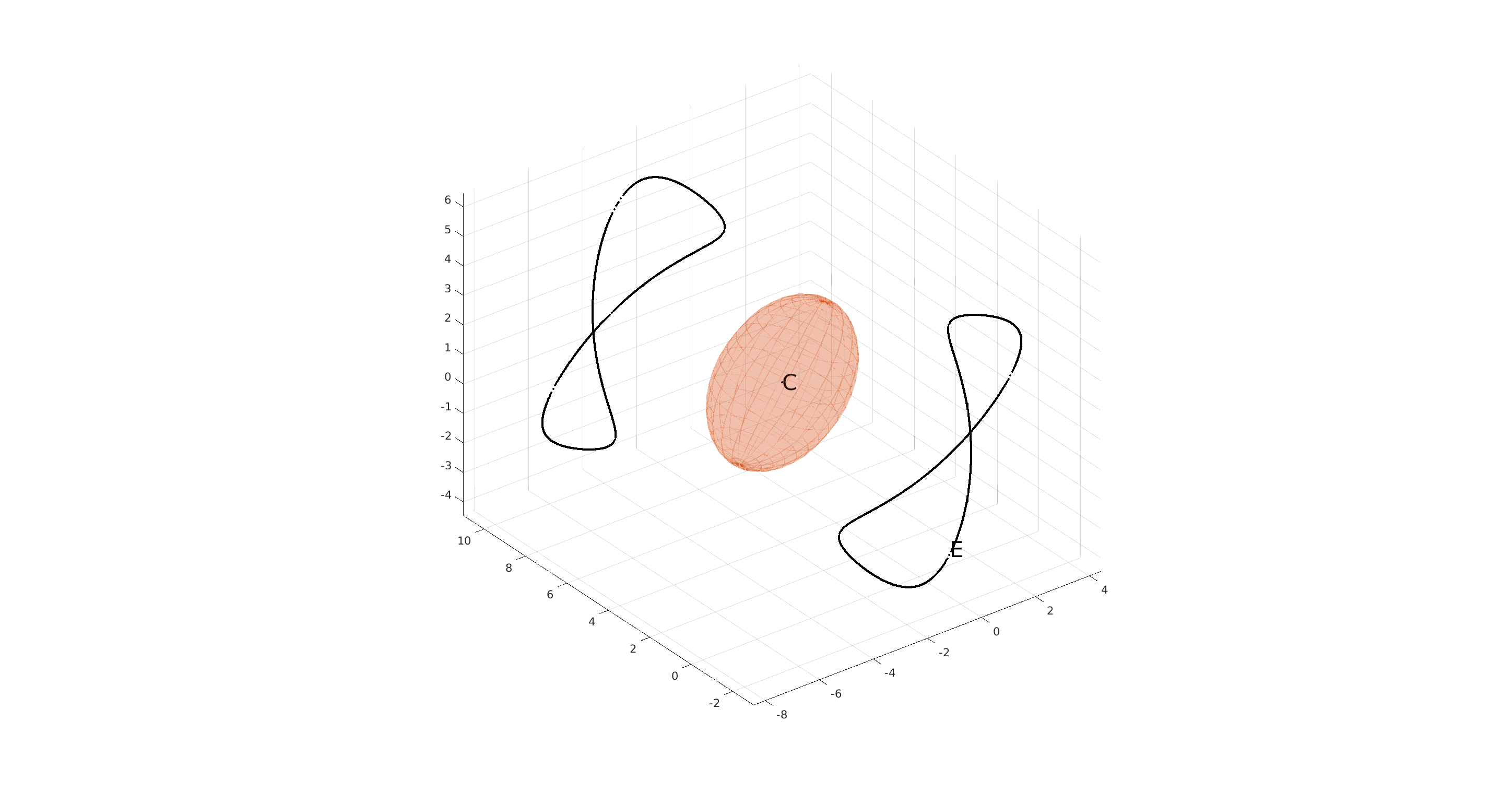}
  \caption{Triaxial ellipsoid: locus of cone vertices {\it i.e.} camera centers.}
  \label{fig:ellipsoids_Rcam}
\end{figure}

In the first case, the infinite number of ellipsoids tangent to the cone (or, conversely, the infinite number of cones tangent to the ellipsoid) explains the infinite number of camera solutions (see Fig. \ref{fig:ellipsoids_Rcam}), and provides a parameterization of them. In the second case, the infinite number of change of basis matrices between the spheroid and the camera explains the infinite number of camera solutions (see Fig. \ref{fig:spheroids_Rcam}). The mathematical developments leading to these results are presented below.

\footnotetext[1]{\href{https://members.loria.fr/moberger/Documents/P1E.mp4}{https://members.loria.fr/moberger/Documents/P1E.mp4}}

\begin{figure}
  \centering
  \includegraphics[trim=165mm 20mm 150mm 20mm, clip, width=0.9\linewidth]{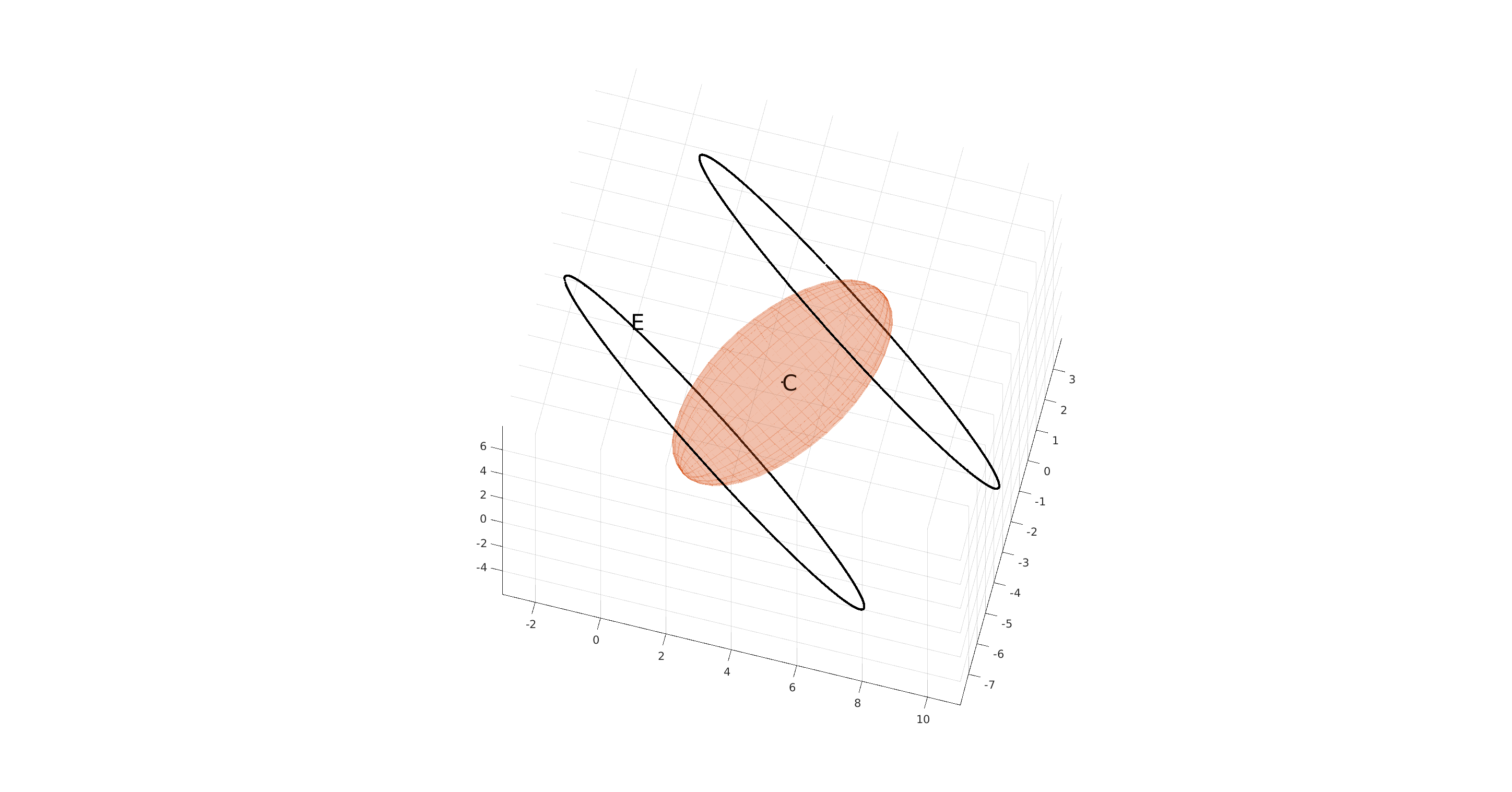}
  \caption{Spheroid: locus of cone vertices {\it i.e.} camera centers.}
  \label{fig:spheroids_Rcam}
\end{figure}

\subsection{The Triaxial Ellipsoid}
\label{ssec:ellipsoid}

\subsubsection{Solving for $\s$}
In practice, not all $\s$ values give rise to a solution of the problem. When the ellipsoid has three distinct radii (triaxial), Theorem \ref{th:VDMA} provides a characterization of the scalars $\s$ solutions of \eqref{eq:main_pb}.

\noindent
\ifboxes
    \framebox{\parbox{\linewidth}{
\fi
\begin{theo} \label{th:VDMA}
Let's denote $(\lambda_{A,1},\lambda_{A,2},\lambda_{A,3})$ the three distinct eigenvalues of $A$. Then $\s=d m^2$ is solution of Equation \eqref{eq:main_pb} if and only if the three entries of vector $M_A^{-1}V(m)$ are all non-negative:
\begin{equation}
    M_A = \begin{pmatrix}
    1 & 1 & 1\\
    \lambda_{A,1} & \lambda_{A,2} & \lambda_{A,3}\\
    \lambda_{A,1}^2 & \lambda_{A,2}^2 & \lambda_{A,3}^2
    \end{pmatrix}\mbox{,}
\end{equation}
\begin{equation}
    V(m) = \begin{pmatrix}
    tr(A^{-1})-\frac{tr(B'^{-1})}{d}m \\ 1-m^3 \\ tr(B')d m^2-tr(A)m^3
    \end{pmatrix}\mbox{,}
\end{equation}
with
\begin{align}
    d &= \sqrt[3]{\frac{det(A)}{det(B')}}\mbox{,}\\
    m &= \sqrt[3]{\mu}\mbox{.} \label{eq:m_def}
\end{align}

It must be noted that $m$ is the only unknown parameter of vector V(m) since all the other ones derive from $A$ and $B'$ eigenvalues.
\end{theo}
\ifboxes
    }}
\fi

\begin{proof}
\framebox{$\Longrightarrow$} Let's assume that Equation \eqref{eq:main_pb} is satisfied:
\begin{equation}\tag{\ref{eq:main_pb}}
    A\D\D^\top A+\mu A=\s B'\mbox{.}
\end{equation}
Therefore, equivalent Equation \eqref{eq:eq_pb} is also satisfied:
\begin{equation}\tag{\ref{eq:eq_pb}}
    \D\D^\top=A^{-1}-\frac{\mu}{\s}B'^{-1}\mbox{.}
\end{equation}

Since the trace of a product of matrices does not depend on the order of the matrices, we have
\begin{align*}
    tr(A\D\D^\top A)&=tr((\D^\top A)(A\D))\\
    &=tr(\D^\top A^2\D)\\
    &=\D^\top A^2\D \mbox{\hspace{0.5cm} \textit{(scalar)},}
\end{align*}
and, similarly
\begin{equation*}
    tr(\D\D^\top)=\D^\top\D\mbox{.}
\end{equation*}

Given, in addition, $\mu=1-\D^\top A\D$, applying $tr()$ to Equations \eqref{eq:eq_pb} and \eqref{eq:main_pb} leads to the following system:
\begin{equation*}
\left\{
    \begin{array}{l}
        \D^\top \D = tr(A^{-1})-\frac{\mu}{\s}tr(B'^{-1})\\
        \D^\top A\D = 1-\mu\\
        \D^\top A^2\D = \s tr(B')-\mu tr(A)\mbox{.}
    \end{array}
\right.
\end{equation*}

Although the two scalar unknowns $\mu$ and $\s$ appear in the right hand side, they can be expressed as functions of a third unknown. Indeed, denoting
\begin{equation*}
    m=\sqrt[3]{\mu}\mbox{\hspace{0.5cm} \textit{(unknown)},}
\end{equation*}
and
\begin{equation*}
    d=\sqrt[3]{\frac{det(A)}{det(B')}}\mbox{\hspace{0.5cm} \textit{(known)},}
\end{equation*}
Equation \eqref{eq:Delta2mu} can be rewritten
\begin{equation*}
    m^3=-\sqrt{\frac{\s^3}{d^3}}\mbox{,}
\end{equation*}
whence, by raising it to the power of $2/3$,
\begin{equation}\label{eq:m2s1}
    \s=dm^2\mbox{.}
\end{equation}
Furthermore, given
\begin{equation*}
    \mu=m^3\mbox{\hspace{0.5cm} \textit{(definition)},}
\end{equation*}
we have
\begin{equation*}
    \frac{\mu}{\s}=\frac{m}{d}\mbox{.}
\end{equation*}

Therefore, $\Delta$ is solution of the following system with unknown $m$:
\begin{equation}\label{eq:sys}
\left\{
    \begin{array}{l}
        \D^\top \D = tr(A^{-1})-\frac{tr(B'^{-1})}{d}m\\
        \D^\top A\D = 1-m^3\\
        \D^\top A^2\D = tr(B')d m^2-tr(A)m^3\mbox{.}
    \end{array}
\right.
\end{equation}

The above equations are independent from the considered coordinate frame. Considering the ellipsoid frame and denoting $(\D_{ell,x},\D_{ell,y},\D_{ell,z})^\top$ the corresponding expression of $\D$, the above system can be rewritten:
\begin{equation*}
    \begin{footnotesize}
    \begin{pmatrix}
    1 & 1 & 1\\
    \lambda_{A,1} & \lambda_{A,2} & \lambda_{A,3}\\
    \lambda_{A,1}^2 & \lambda_{A,2}^2 & \lambda_{A,3}^2
    \end{pmatrix}
    \begin{pmatrix}
    \D_{ell,x}^2 \\ \D_{ell,y}^2 \\ \D_{ell,z}^2
    \end{pmatrix}=
    \begin{pmatrix}
    tr(A^{-1})-\frac{tr(B'^{-1})}{d}m \\
    1-m^3 \\
    tr(B')d m^2-tr(A)m^3
    \end{pmatrix}\mbox{,}
    \end{footnotesize}
\end{equation*}
{\it i.e.}
\begin{equation*}
    M_A
    \begin{pmatrix}
    \D_{ell,x}^2 \\ \D_{ell,y}^2 \\ \D_{ell,z}^2
    \end{pmatrix}=
    V(m)\mbox{.}
\end{equation*}

Since $A$ eigenvalues are all different (triaxial ellipsoid), Vandermonde matrix $M_A$ is not singular (cf Appendix \ref{secA_Vandermonde}). Therefore, the system can be inverted:
\begin{equation}\label{eq:m2Dell}
    \begin{pmatrix}
    \D_{ell,x}^2 \\ \D_{ell,y}^2 \\ \D_{ell,z}^2
    \end{pmatrix}=M_A^{-1}V(m)\mbox{.}
\end{equation}
Left hand side elements are all non-negative, whence the result.

\medskip
\framebox{$\Longleftarrow$} Let's now assume that the three entries of $M_A^{-1}V(m)$ are all non-negative. 

Let $\D_{ell}=(\D_{ell,x},\D_{ell,y},\D_{ell,z})^\top$ be a vector such that
\begin{equation*}
    \begin{pmatrix}
    \D_{ell,x}^2 \\ \D_{ell,y}^2 \\ \D_{ell,z}^2
    \end{pmatrix}=
    M_A^{-1}V(m)\mbox{.}
\end{equation*}
Such a definition is possible due to the positivity of the three entries.

One can therefore demonstrate, with the help of a formal calculus software (the corresponding Maple code is provided in Appendix \ref{sec:VDMA} as reference), that, irrespective of the sign assumptions made for $\D_{ell}$ entries, the matrix
\begin{equation*}
\frac{\s}{\mu}(A_{ell}^{-1}-\D_{ell}\D_{ell}^\top) = \frac{d}{m}(A_{ell}^{-1}-\D_{ell}\D_{ell}^\top)
\end{equation*}
has the same eigenvalues with same multiplicities as $B'^{-1}$. Given that both matrices are diagonalizable (since symmetric),
this amounts to say that they are similar, and thus that Equation \eqref{eq:eq_pb} is satisfied.
\end{proof}

\subsubsection{Solving for Camera Poses}

\begin{figure}
  \centering
  \includegraphics[trim=165mm 20mm 150mm 20mm, clip, width=0.9\linewidth]{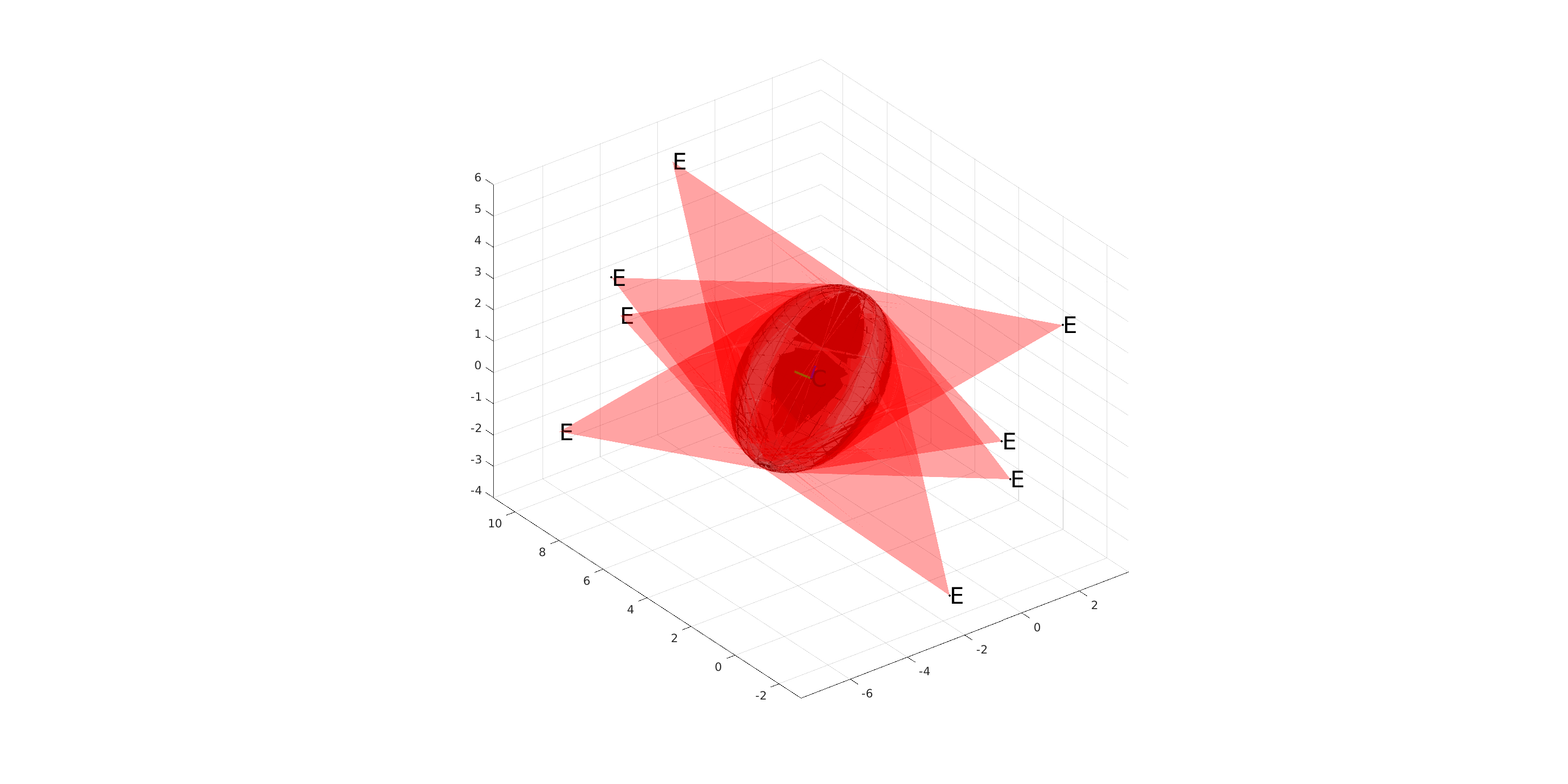}
\caption{Illustrating the eight backprojection cones tangent to the triaxial ellipsoid for a given $\s$ value.}
\label{fig:8cones}
\end{figure}

\noindent
\ifboxes
    \framebox{\parbox{\linewidth}{
\fi
\begin{theo} \label{th:ellipsoid}
Considering a triaxial ellipsoid, each $\s$ solution of \eqref{eq:main_pb} gives rise to \textbf{\textit{eight}} backprojection cones $(\E,B')$ tangent to the ellipsoid. These cones are symmetric with respect to the three principal planes of the ellipsoid (see Fig. \ref{fig:8cones}).

In addition, each backprojection cone defines \textbf{\textit{two}} camera solutions (see Fig. \ref{fig:2cam}).
\end{theo}
\ifboxes
    }}
\fi

\begin{proof}
Theorem \ref{th:VDMA} provides a NSC on $\s$ to be solution of \eqref{eq:main_pb}. Moreover, its proof exhibits that vectors $\D$ solutions are expressed in the ellipsoid frame in the form
\begin{equation}\label{eq:Dfroms_ell}
    \D_{ell}=\begin{pmatrix}
    \pm\sqrt{\D_{ell,x}^2}\\
    \pm\sqrt{\D_{ell,y}^2}\\
    \pm\sqrt{\D_{ell,z}^2}
    \end{pmatrix}
\end{equation}
where
\begin{equation*}
    \begin{pmatrix}
    \D_{ell,x}^2 \\ \D_{ell,y}^2 \\ \D_{ell,z}^2
    \end{pmatrix}=
    M_A^{-1}V(m)
\end{equation*}
There are thus eight vectors $\D_{ell}$ solutions for a given $m$ (thus $\s$), and they are symmetric with respect to the three principal planes of the ellipsoid.

The cone vertices (\textit{i.e.} camera positions) can then be derived:
\begin{equation*}
    \E_{ell}=\CC_{ell}+\D_{ell}=\D_{ell}\mbox{.}
\end{equation*}

\begin{figure}
  \centering
  \includegraphics[trim=165mm 20mm 150mm 20mm, clip, width=0.9\linewidth]{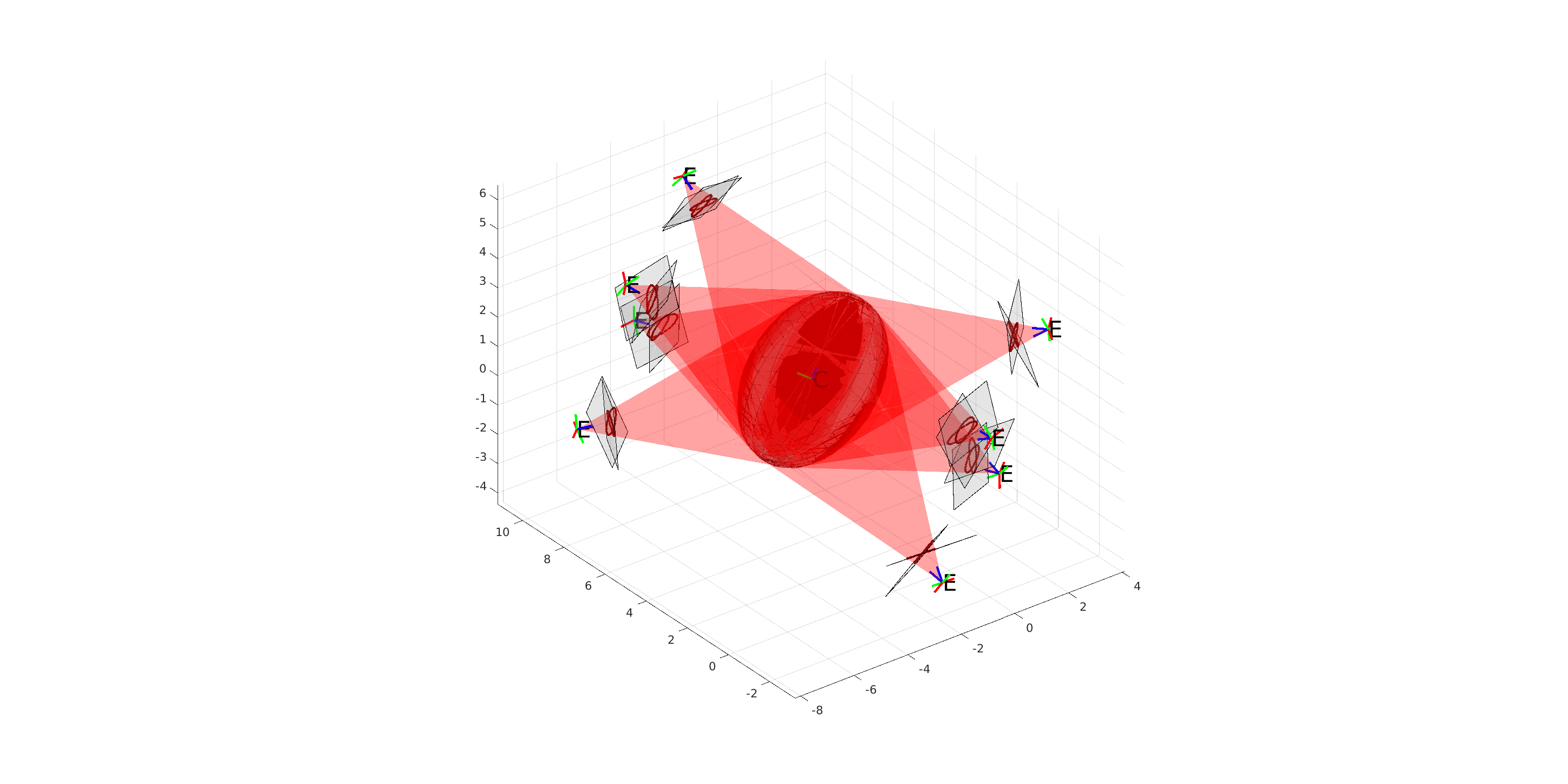}
\caption{Illustrating the two cameras compatible with each backprojection cone tangent to the triaxial ellipsoid.}
\label{fig:2cam}
\end{figure}

Let us now solve for camera orientations.
Equation \eqref{eq:main_pb} provides the expression of $B_{ell}'$:
\begin{equation}\label{eq:Bfroms}
    B'_{ell}=\frac{1}{dm^2}A_{ell}\D_{ell}\D_{ell}^\top A_{ell}+\frac{m}{d}A_{ell}
\end{equation}
Orientations $\Rcell^\top$ of the cameras then verify:
\begin{equation*}
    B_{ell}'=\Rcell^\top B_{cam}'\Rcell\mbox{.}
\end{equation*}

Since the cone is non-circular (see Section \ref{ssec:cooc}), $B_{ell}'$ and $B_{cam}'$ eigenvectors are defined with minimum ambiguity. By arbitrarily fixing the directions of $B_{ell}'$ eigenvectors for instance, it then remains four ways of choosing the directions of $B_{cam}'$ eigenvectors so that the change of basis matrix $\Rcell$ is a rotation matrix. Yet, over the four resulting orientations, only two leads to an ellipsoid located in front of the camera.

\end{proof}

To summarize, to each $\s$ corresponds sixteen camera poses covering eight different positions (see Fig. \ref{fig:2cam}).


\begin{algorithm}
\caption{P$1$E solutions  for the triaxial ellipsoid (expressed in the ellipsoid frame)}
\label{alg:overall-triaxial}
\begin{algorithmic}
\State $m$ $\gets$ solutions of polynomial inequality $M_A^{-1}V(m)\geq0$ (Theorem \ref{th:VDMA})
\State $\D_{ell} \gets$ Equation \eqref{eq:Dfroms_ell}
\State $B_{ell}' \gets$ Equation \eqref{eq:Bfroms}
\State $\E_{ell} \gets \D_{ell}+\CC_{ell}$
\State $\Rellcone \gets$ eigenvectors of $B_{ell}'$
\State $\Rcell \gets \Rccone\Rellcone^\top$
\end{algorithmic}
\end{algorithm}

\subsection{The Spheroid}
\label{ssec:spheroid}

When the ellipsoid has a revolution axis (\textit{i.e.} spheroid), we use a different approach since Vandermonde matrix $M_A$ is now singular and thus cannot be inverted. We then determine the set of spheroids tangent to the backprojection cone, and distinguish between the two possible types of cone. It is worth noting that this problem has already been addressed in \cite{WokesP10} using a different parameterization. The authors especially show that in the general case (non-circular elliptic cone), there are only two tangent spheroids, and we retrieve this result below.

\subsubsection{The Non-circular Elliptic Cone}

Let us first consider a non-circular elliptic cone. Expressing the Cone Alignment Equation in the cone coordinate frame, and given that the three $B'$ eigenvalues are different, $\s$ solutions can be characterized in a similar way to Theorem \ref{th:VDMA}.

\noindent
\ifboxes
    \framebox{\parbox{\linewidth}{
\fi
\begin{res} \label{th:VDMB}
Let's denote $(\lambda_{B',1},\lambda_{B',2},\lambda_{B',3})$ the three distinct eigenvalues of $B'$. Then $\s=d m^2$ is solution of Equation \eqref{eq:main_pb} if and only if the three entries of vector $M_{B'}^{-1}V'(m)$ are all non-negative:
\begin{equation}
    M_{B'}=\begin{pmatrix}
    1 & 1 & 1\\
    \lambda_{B',1} & \lambda_{B',2} & \lambda_{B',3}\\
    \lambda_{B',1}^2 & \lambda_{B',2}^2 & \lambda_{B',3}^2
    \end{pmatrix}\mbox{,}
\end{equation}
\begin{equation}
    V'(m) = 
    \begin{pmatrix}
    1 & 0 & 0 \\
    0 & \frac{1}{dm^2} & 0 \\
    0 & 0 & \frac{1}{d^2 m^4}
    \end{pmatrix}
    V(m)\mbox{.}
\end{equation}
\end{res}
\ifboxes
    }}
\fi

\begin{proof}
The proof is based on the exact same arguments as the proof of Theorem \ref{th:VDMA}. In particular, $M_{B'}^{-1}V'(m)$ is related to the expression $\D_{cone}$ of $\D$ in the cone frame:
\begin{equation}\label{eq:m2Dcone}
    \begin{pmatrix}
    \D_{cone,x}^2 \\ \D_{cone,y}^2 \\ \D_{cone,z}^2
    \end{pmatrix}=M_{B'}^{-1}V'(m)
\end{equation}
\end{proof}

It is interesting noting that the above result is also valid in the case of a triaxial ellipsoid since the cone is then non-circular (Section \ref{ssec:cooc}). It can therefore be used to reconstruct the ellipsoids in the camera coordinate frame (see Fig. \ref{fig:ellipsoids}).

Unlike the triaxial ellipsoid for which there is an infinite number of $\s$ solutions, each one giving rise to a fixed number (16) of camera poses, we demonstrate in Theorem \ref{th:spheroid} that there is only one $\s$ solution for the spheroid, which gives rise to an infinite number of camera poses.

Let's consider $\lambda_{A,single}$ (multiplicity 1) and $\lambda_{A,double}$ (multiplicity 2) the eigenvalues of $A$. Let's also consider $(\lambda_{B',1},\lambda_{B',2},\lambda_{B',3})$ the eigenvalues of $B'$, where $\lambda_{B',1}$ and $\lambda_{B',2}$ have the same sign (opposed to the sign of $\lambda_{B',3}$). Finally, let's assume, even if it requires exchanging them, that $\lvert\lambda_{B',1}\rvert>\lvert\lambda_{B',2}\rvert$. 

\noindent
\ifboxes
    \framebox{\parbox{\linewidth}{
\fi
\begin{theo} \label{th:spheroid}
Considering a spheroid along with a non-circular backprojection cone, there is only one $\s$ value solution of Equation \eqref{eq:main_pb}:
\begin{equation}\label{eq:sspheroid}
\s=
\left\{
    \begin{array}{l}
        \frac{\lambda_{A,single}\lambda_{B',1}}{\lambda_{B',2}\lambda_{B',3}} \mbox{ \hspace{0.5cm} if $\lambda_{A,single}<\lambda_{A,double}$ }\\
        \frac{\lambda_{A,single}\lambda_{B',2}}{\lambda_{B',1}\lambda_{B',3}} \mbox{ \hspace{0.5cm} if $\lambda_{A,single}>\lambda_{A,double}$ }
    \end{array}
\right.
\end{equation}
That $\s$ value gives rise to \textbf{\textit{two}} spheroids tangent to the cone, that are symmetric with respect to one of the cone principal planes (see Fig. \ref{fig:spheroids}).
\end{theo}
\ifboxes
    }}
\fi

\begin{proof}
According to Theorem \ref{th:VDMB}, $\s=dm^2$ is solution of \eqref{eq:main_pb} if and only if the three entries of the following vector are all non-negative: 

\begin{equation*}
    \begin{pmatrix}
    \D_{cone,x}^2(m) \\ \D_{cone,y}^2(m) \\ \D_{cone,z}^2(m)
    \end{pmatrix}=
    M_{B'}^{-1}V'(m)\mbox{.}
\end{equation*}

\paragraph{}Yet, developing the right hand term leads to the following system:

\begin{equation*}
\left\{
    \begin{array}{l}
        \D_{cone,x}^2(m) = \frac{1}{m^2}P_1(m)\\\\
        \D_{cone,y}^2(m) = \frac{1}{m^2}P_2(m)\\\\
        \D_{cone,z}^2(m) = \frac{1}{m^2}P_3(m)
    \end{array}
\right.
\end{equation*}
where
\begin{equation*}
\left\{
    \begin{array}{l}
        P_1(x) = k_1\left(x-\frac{\lambda_{B',1}}{\lambda_{A,single}}d\right)\left(x-\frac{\lambda_{B',1}}{\lambda_{A,double}}d\right)^2\\\\
        P_2(x) = k_2\left(x-\frac{\lambda_{B',2}}{\lambda_{A,single}}d\right)\left(x-\frac{\lambda_{B',2}}{\lambda_{A,double}}d\right)^2\\\\
        P_3(x) = k_3\left(x-\frac{\lambda_{B',3}}{\lambda_{A,single}}d\right)\left(x-\frac{\lambda_{B',3}}{\lambda_{A,double}}d\right)^2
    \end{array}
\right.
\end{equation*}
and where
\begin{equation*}
\left\{
    \begin{array}{l}
        k_1 = \frac{-\lambda_{B',2}\lambda_{B',3}}{\lambda_{B',1}(\lambda_{B',1}-\lambda_{B',2})(\lambda_{B',1}-\lambda_{B',3})d}\\\\
        k_2 = \frac{-\lambda_{B',1}\lambda_{B',3}}{\lambda_{B',2}(\lambda_{B',2}-\lambda_{B',1})(\lambda_{B',2}-\lambda_{B',3})d}\\\\
        k_3 = \frac{-\lambda_{B',1}\lambda_{B',2}}{\lambda_{B',3}(\lambda_{B',3}-\lambda_{B',1})(\lambda_{B',3}-\lambda_{B',2})d}
    \end{array}
\right.
\end{equation*}

The locus of scalars $m$ solutions is the subset of $\RR$ on which $P_1(x)$, $P_2(x)$ and $P_3(x)$ are all non-negative. 

To study the variations of these three polynomials, four cases--described in Table \ref{tab:config} and depending on the relative order of $A$ and $B'$ eigenvalues--need to be considered. Only case \#1 is addressed below, since other cases can be solved using a similar reasoning.

\begin{table}[h]
\centering
\begin{tabular}{|l|c|c|}
  \hline
  & $\lambda_{B',2}<0$ & $\lambda_{B',2}>0$ \\
  \hline
  $\lambda_{A,single}<\lambda_{A,double}$ & \#1 & \#2 \\
  \hline
  $\lambda_{A,single}>\lambda_{A,double}$ & \#3 & \#4 \\
  \hline
\end{tabular}
\smallskip
\caption{Configurations of the problem depending on $A$ and $B'$ eigenvalues.}
\label{tab:config}
\end{table}

Let's denote $S_i$ the root of $P_i(x)$ with multiplicity 1, and $D_i$ the root with multiplicity 2, such that
\begin{equation*}
    S_i=\frac{\lambda_{B',i}}{\lambda_{A,single}}d \mbox{\hspace{0.5cm} and \hspace{0.5cm}} D_i=\frac{\lambda_{B',i}}{\lambda_{A,double}}d.
\end{equation*}

In  configuration \#1, studying the variations of every $P_i$ leads to only one solution to obtain simultaneous non-negative values, which is the root of $P_1$ with multiplicity 2: $x=D_1$ (the proof is given in Appendix \ref{sec:demospheroid}).

 The unique $\s$ solution is then given by
\begin{align*}
    \s&=dD_1^2\\
    &=\frac{\lambda_{A,single}\lambda_{B',1}}{\lambda_{B',2}\lambda_{B',3}}\mbox{,}
\end{align*}
after replacing $d$ by its expression as a function of $A$ and $B'$ eigenvalues.

\medskip
Furthermore, since $D_1$ is a root of $P_1(x)$, vectors $\D_{cone}$ expressed in the cone frame verify:
\begin{equation*}
\left\{
    \begin{array}{l}
        \D_{cone,x}^2 = 0\\
        \D_{cone,y}^2 = \frac{1}{D_1^2}P_2(D_1)\\
        \D_{cone,z}^2 = \frac{1}{D_1^2}P_3(D_1)\mbox{.}
    \end{array}
\right.
\end{equation*}
Then
\begin{equation*}
    \D_{cone}=\begin{pmatrix}
    0 \\
    \pm\sqrt{\frac{1}{D_1^2}P_2(D_1)}\\
    \pm\sqrt{\frac{1}{D_1^2}P_3(D_1)}
    \end{pmatrix}
\end{equation*}

Since the sign of the third entry ($\D_{cone,z}$) is fixed under the chirality constraint (ellipsoid located in front of the camera), there remains two possible expressions for $\D_{cone}$. The two resulting vectors are symmetric with respect to the cone principal plane whose normal is the eigenvector corresponding to eigenvalue $\lambda_{B',2}$.

Equation \eqref{eq:Delta2A} then provides the expressions of $A$ in the cone coordinate frame:
\begin{equation*}
    A_{cone}=\frac{\s}{\mu}B'_{cone}-\frac{\s^2}{\mu}B'_{cone}\D_{cone}\D_{cone}^\top B'_{cone}\mbox{.}
\end{equation*}

One can therefore derive the expressions of $\D$ and $A$ in the camera frame using $\Rccone$ (known):
\begin{align*}
    \D_{cam} &= \Rccone\D_{cone}\mbox{,}\\
    A_{cam} &= \Rccone A_{cone}\Rccone^\top\mbox{,}
\end{align*}
then the spheroid centers are:
\begin{equation*}
    \CC_{cam}=\E_{cam}-\D_{cam}=-\D_{cam}
\end{equation*}
\end{proof}

Now that the spheroid solutions have been determined in the camera coordinate frame, one can deduce the poses of camera solutions.

\noindent
\ifboxes
    \framebox{\parbox{\linewidth}{
\fi
\begin{res} \label{res:spheroid_cameras}
Considering a spheroid along with a non-circular backprojection cone, the axial symmetry of the spheroid leads to an infinite number of camera solutions. The solutions belong to two planes orthogonal to the revolution axis of the spheroid and located at the same distance from its center (see Fig \ref{fig:spheroids_Rcam}).
\end{res}
\ifboxes
    }}
\fi

\begin{proof}
Since there is only one possible value for $\s$, vectors $\D$ all have the same norm (cf. Result \ref{res:Link-sD}), {\it i.e.} the cameras are located at a fixed distance from the spheroid center.

Furthermore, orientations $\Rcell^\top$ of these cameras verify:
\begin{equation*}
    A_{cam}=\Rcell A_{ell}\Rcell^\top
\end{equation*}

Since the spheroid has a revolution axis, arbitrarily fixing $A_{ell}$ eigenvectors for instance leaves two choices for one of $A_{cam}$ eigenvectors (the one corresponding to the revolution axis) and an infinite number of choices for the other two.
\end{proof}



\begin{algorithm}
\caption{P$1$E solutions  for the non-circular cone (expressed in camera frame)}
\label{alg:overall-noncirc}
\begin{algorithmic}
\State $\s \gets$ Equation \eqref{eq:sspheroid}
\State $\mu \gets$ Equation \eqref{eq:Delta2mu}
\State $m \gets$ Definition \eqref{eq:m_def}
\State $\D_{cone} \gets$ Equation \eqref{eq:m2Dcone}
\State $\D_{cam} \gets \Rccone\D_{cone}\Rccone^\top$
\State $A_{cam} \gets$ Equation \eqref{eq:Delta2A}
\State $\CC_{cam} \gets \E_{cam}-\D_{cam}$
\State $\Rcell \gets$ eigenvectors of $A_{cam}$
\end{algorithmic}
\end{algorithm}

\subsubsection{The Circular Cone}

Considering the case of a circular cone (elliptic cone with a revolution axis), we are going to demonstrate that there is only one spheroid tangent to it (Result \ref{res:sspheroid2}).

In this case, $B'$ has two distinct eigenvalues. Let's denote them $\lambda_{B',single}$ (multiplicity 1) and $\lambda_{B',double}$ (multiplicity 2).

\begin{figure}
  \centering
  \includegraphics[trim=10cm 1.7cm 10cm 1.1cm, clip, width=0.9\linewidth]{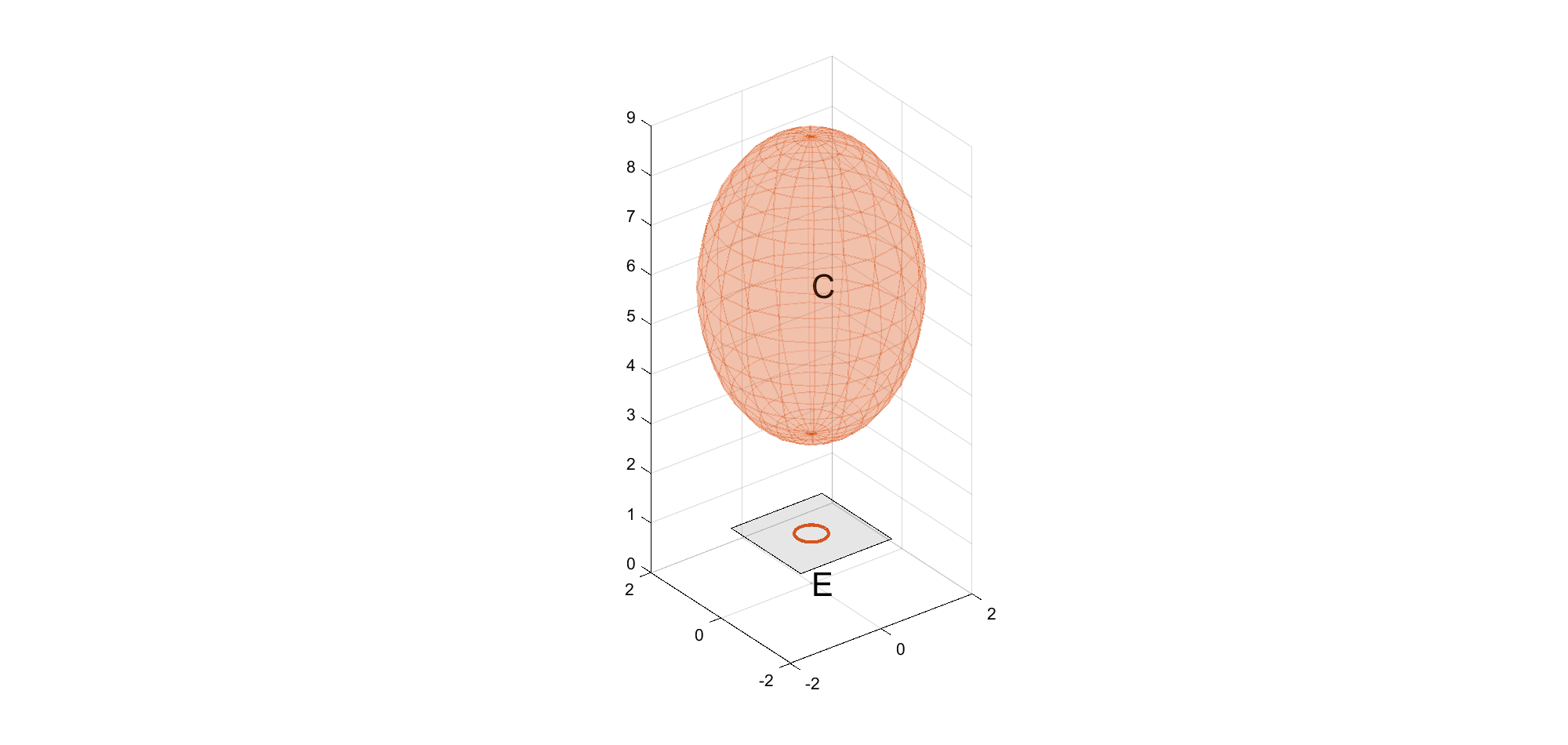}
\caption{The spheroid solution with a circular backprojection cone.}
\label{fig:spheroid2}
\end{figure}

\noindent
\ifboxes
    \framebox{\parbox{\linewidth}{
\fi
\begin{res} \label{res:sspheroid2}
Considering a spheroid along with a circular backprojection cone, there is only one $\s$ value solution of Equation \eqref{eq:main_pb}:
\begin{equation}\label{eq:sspheroid2}
\s=\frac{\lambda_{A,single}}{\lambda_{B',single}}\mbox{.}
\end{equation}
That $\s$ value gives rise to \textbf{\textit{one}} spheroid tangent to the cone, and both revolution axes coincide (See Fig. \ref{fig:spheroid2}). The distance between the cone vertex and the spheroid center is given by:
\begin{equation}\label{eq:Dspheroid2}
\|\D\|=\sqrt{\frac{1}{\lambda_{A,single}}\left(1-\frac{\lambda_{A,single}\lambda_{B',double}}{\lambda_{A,double}\lambda_{B',single}}\right)}\mbox{.}
\end{equation}
\end{res}
\ifboxes
    }}
\fi

Demonstration of Result \ref{res:sspheroid2} is provided in appendix \ref{dem:resspheroid2}.

$A_{cone}$ can then be obtained from \eqref{eq:Delta2A}, then $A_{cam}$ and $\CC_{cam}$ using $\Rccone$ (known).

Now the spheroid solution has been determined in the camera frame, we can infer the corresponding camera poses.

\noindent
\ifboxes
    \framebox{\parbox{\linewidth}{
\fi
\begin{res} \label{res:spheroid_cameras2}
When considering a spheroid along with a circular backprojection cone, there are two solutions for camera position, that are located on the revolution axis of the spheroid and at the same distance from its center, and an infinite number of solutions for camera orientation.
\end{res}
\ifboxes
    }}
\fi

\begin{proof}
Equation \eqref{eq:D_spheroid2} gives the two possible solutions for $\D$. For reasons of symmetry, both of them are actual solutions.

Furthermore, orientations $\Rcell^\top$ of these cameras verify:
\begin{equation*}
    A_{cam}=\Rcell A_{ell}\Rcell^\top
\end{equation*}

Just as when considering a non-circular elliptic cone (Result \ref{res:spheroid_cameras}), we conclude on the infinite number of camera orientations.
\end{proof}

\ifhighlightrev
    \rev{
\fi
A summary of the method is provided in Algorithm  \ref{alg:overall-spheroid-circ}.
\ifhighlightrev
    }
\fi

\begin{algorithm}
\caption{P$1$E solutions for the spheroid with circular cone (expressed in the ellipsoid frame)}
\label{alg:overall-spheroid-circ}
\begin{algorithmic}
\State $\s \gets$ Equation \eqref{eq:sspheroid2}
\State $\|\D\| \gets$ Equation \eqref{eq:Dspheroid2}
\State $D_{ell} \gets $ Equation \eqref{eq:D_spheroid2}
\State $B_{ell}' \gets$ Equation \eqref{eq:Delta2B}
\State $\E_{ell} \gets \D_{ell}+\CC_{ell}$
\State $\Rcell \gets$ left eigenvectors of $B_{ell}'$
\end{algorithmic}
\end{algorithm}

\subsection{The Sphere}
\label{sec:sphere}
When the ellipsoid is a sphere, the matrix $A$ has the same expression in every basis:
\begin{equation}
    A=\lambda_{A,triple}I\mbox{,}
\end{equation}
where $R=\frac{1}{\sqrt{\lambda_{A,triple}}}$ is the sphere radius.

Given its observation as a circle in the image, there is obviously an infinite number of camera solutions, that are located at the same distance from the sphere centre.

\ifhighlightrev
    \rev{
\fi
Using the formalism of our study, we can specify these properties with the following result:
\ifhighlightrev
    }
\fi

\noindent
\ifboxes
    \framebox{\parbox{\linewidth}{
\fi
\begin{res} \label{res:ssphere}
When considering a sphere, there is only one $\s$ solution of Equation \eqref{eq:main_pb}:
\begin{equation}\label{eq:ssphere}
\s=\frac{\lambda_{A,triple}}{\lambda_{B',single}}\mbox{.}
\end{equation}
That $\s$ value defines a unique sphere tangent to the cone, whose center belongs the revolution axis of the cone. The distance from the cone vertex to the sphere center is given by:
\begin{equation}\label{eq:Dsphere}
\|\D\|=\sqrt{\frac{1}{\lambda_{A,triple}}\left(1-\frac{\lambda_{B',double}}{\lambda_{B',single}}\right)}\mbox{.}
\end{equation}
The locus of camera positions is  then a sphere with radius $\|\D\|$ around the ellipsoid center.
\end{res}
\ifboxes
    }}
\fi

Proof of Result \ref{res:ssphere} is provided in Appendix \ref{sec:sphere_annex}.

\ifhighlightrev
    \rev{
\fi
A summary of the method is provided in Algorithm \ref{alg:overall-sphere}.
\ifhighlightrev
    }
\fi

\begin{algorithm}
\caption{P$1$E solutions  for the sphere (expressed in the camera frame)}
\label{alg:overall-sphere}
\begin{algorithmic}
\State $\s \gets$ Equation \eqref{eq:ssphere}
\State $\|\D\| \gets$ Equation \eqref{eq:Dsphere}
\State $D_{cone} \gets \left(0,0,\|\D\|\right)^\top$ (assuming that the $z$-axis is the axis of revolution of the cone)
\State $\D_{cam} \gets \Rccone\D_{cone}\Rccone^\top$
\State $A_{cam} \gets$ Equation \eqref{eq:Delta2A}
\State $\CC_{cam} \gets \E_{cam}-\D_{cam}$
\State $\Rcell \gets$ eigenvectors of $A_{cam}$ (\textit{i.e.} any rotation matrix)
\end{algorithmic}
\end{algorithm}

\subsection{Examples of Retrieved Poses}
\label{ssec:PnE}

\begin{figure}
  \centering
  \includegraphics[trim=80mm 15mm 80mm 5mm, clip, width=0.9\linewidth]{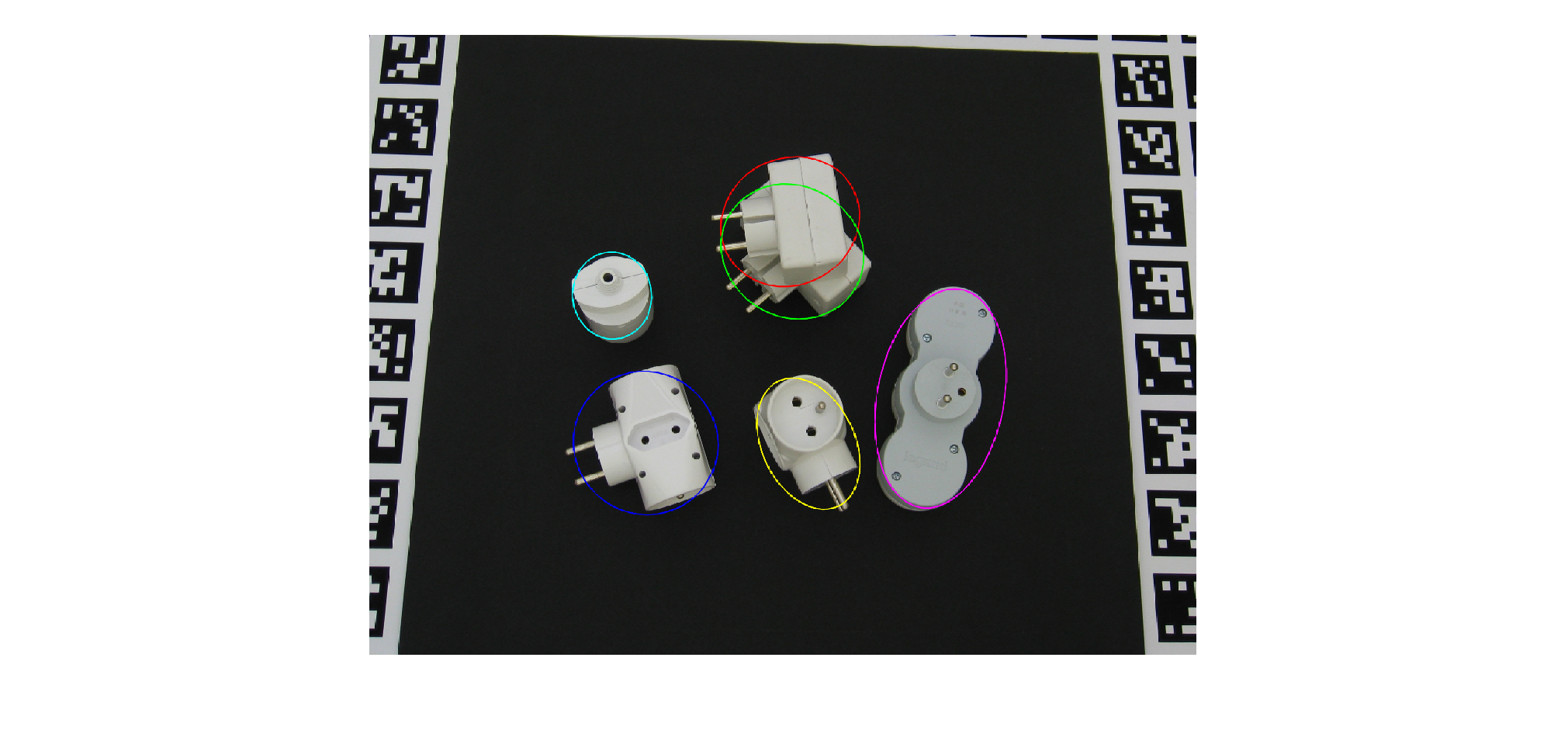}\\
  \includegraphics[trim=15cm 2.4cm 14cm 3.2cm, clip, width=0.95\linewidth]{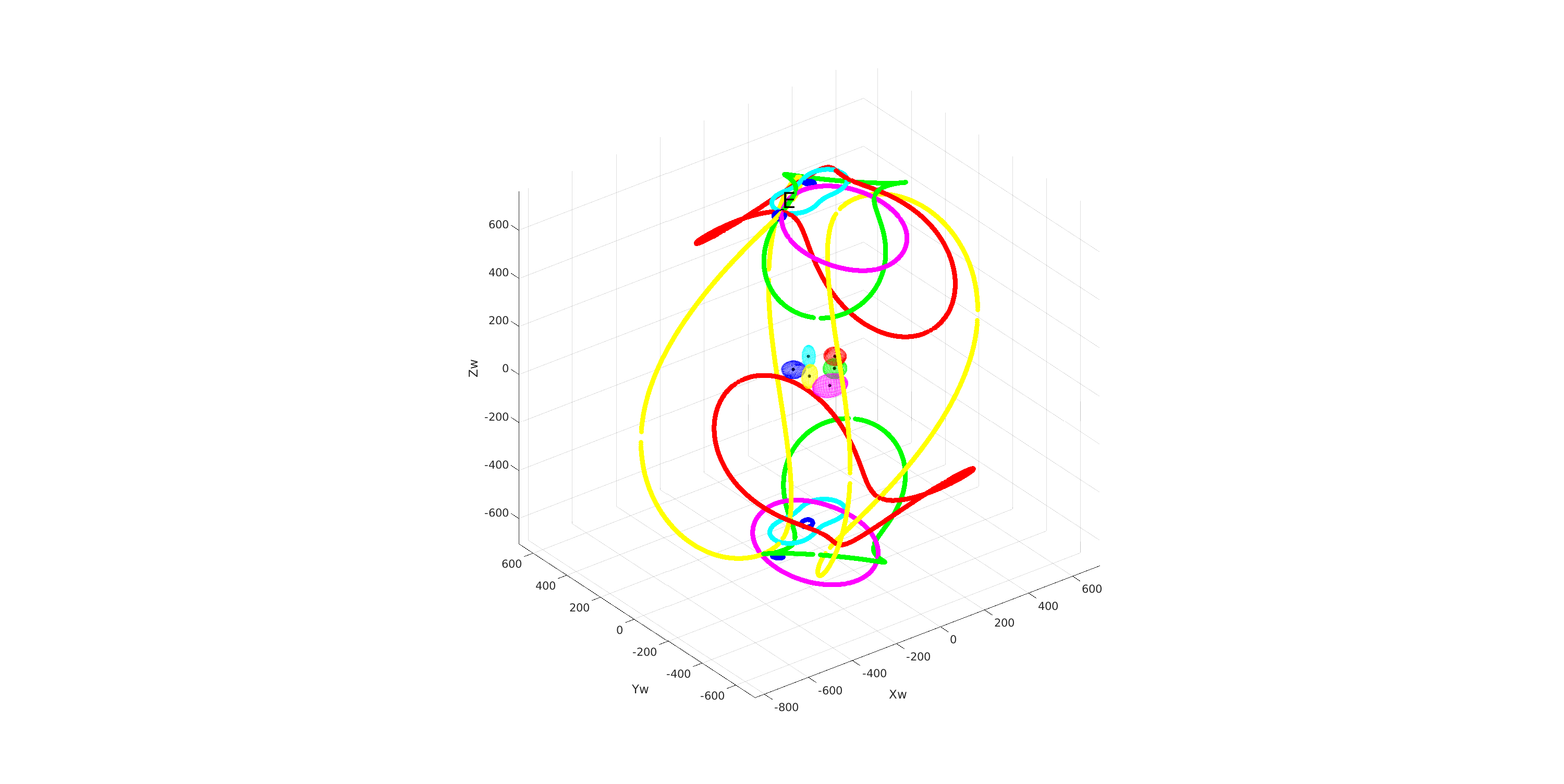}
\caption{Examples of camera loci (i.e. solutions for position) corresponding to six different ellipse-ellipsoid pairs. Corresponding 2D ellipse (top), 3D ellipsoid (bottom) and 3D camera locus (bottom) use the same color. The groundtruth camera position $\E$ is at the unique intersection of the six loci. The image is extracted from the T-LESS dataset \citep{SturmEEBC12}.}
\label{fig:traj}
\end{figure}

We provide in Fig. \ref{fig:traj} a few examples of retrieved camera \rev{loci} from several ellipse-ellipsoid correspondences on a real scene from the T-LESS dataset \citep{SturmEEBC12}. Ellipsoidal models of objects (all triaxial) were reconstructed using \cite{7919240}. Ellipsoids have then been reprojected into one image of the sequence using the groundtruth projection matrix. Finally, the loci of camera solutions were retrieved using our method (Section \ref{ssec:ellipsoid}, Algo. \ref{alg:overall-triaxial}). In the figure, ellipse and ellipsoid colors coincide with those of the loci. Naturally, all the loci intersect at the ground-truth camera position $\E$. 
\ifhighlightrev
    \rev{
\fi
This example illustrates one possible use of our results. Practical aspects are discussed in more details in Section \ref{sec:discussion}.
\ifhighlightrev
    }
\fi

\ifhighlightrev
    \rev{
\fi
\section{P$1$E in Practice}
\label{sec:discussion}
\ifhighlightrev
    }
\fi

\ifhighlightrev
    \rev{
\fi
This section aims at providing a discussion on how the theoretical results presented in this paper can be leveraged in practice (Section \ref{ssec:7-1}). We distinguish the general P$1$E case from the decoupling orientation and position case which do not  have the same level of maturity for applications. We also explain why the triaxial  ellipsoid case is in practice a significant improvement over the spheroid case in Section \ref{sec:triaxialSpheroid}. Finally we give  insights on the sensitivity of our methods to noise in Section \ref{ssec:7-2} and propose  research tracks to improve their robustness to noise in section \ref{sec:robust}. 
\ifhighlightrev
    }
\fi

\ifhighlightrev
    \rev{
\fi
\subsection{Practical interests}
\label{ssec:7-1}
\ifhighlightrev
    }
\fi

\ifhighlightrev
    \rev{
\fi
The orientation-position decoupling (Section \ref{sec:decoupl}), and in particular the position-from-orientation derivation (Section \ref{ssec:pos}), has already demonstrated great practical interest. More precisely, the constraint on the generalized eigenvalues of $\{A,B'\}$ (Result \ref{res:ABeig}) has been used to refine first estimates of the orientation, prior to computing the position~\citep{IROS}. Considering camera orientations obtained either by IMU measurements or by vanishing point detection, camera positions have been computed from multiple ellipse-ellipsoid correspondences in a RANSAC algorithm with minimal sampling size of 1 correspondence using Result \ref{res:A2Delta}~\citep{ISMAR}. A 2-step method to compute the camera orientation from 2 ellipse-ellipsoid correspondences then derive the position using Result \ref{res:A2Delta} has been designed and integrated in a RANSAC algorithm~\citep{IROS2,RAL}. All of these methods have been evaluated on standard camera pose estimation benchmarks. 
The opposite case (orientation from position, Section \ref{ssec:ori}) has been investigated in~\citep{Rathinam_2022_IAC}, where a 2-step method is used to compute the position then derive the orientation (Result \ref{res:Delta2A}) of a spacecraft modeled by an ellipsoid. Applications are however less obvious because it is generally more practical to retrieve an orientation using a sensor carried by the subject than the subject's position in a world frame which, apart from vision-based methods,  requires the use of external tracking sensors (\textit{e.g.} motion capture cameras). Today's cell phones have both GPS and orientation sensors, but there are situations where only GPS is available -- e.g. in a vehicle, for AR displayed on the windshield. Another consideration is that, just as some environments are GPS-denied, others can be hostile to IMU sensors whose magnetometers are sensitive to the presence of metallic objects.
\ifhighlightrev
  }
\fi

\ifhighlightrev
    \rev{
\fi
The results and algorithms presented in Section \ref{sec:closedform} have not yet been leveraged in practice. A direct application, leveraging at least 2 ellipse-ellipsoid correspondences and suggested in Section \ref{ssec:PnE}, would be to compute the pose solutions from each correspondence then find the intersection. Moreover, the P$1$E constraints can be used in combination with other primitives such as points or lines to compute a unique solution.\\
\ifhighlightrev
   }
\fi


\ifhighlightrev
    \rev{
\fi
\subsection{ Triaxial ellipsoid versus spheroid}
\label{sec:triaxialSpheroid}
\ifhighlightrev
    }
\fi
\ifhighlightrev
    \rev{
\fi
One of the main contribution  of this paper is the solution of P$1$E for the generic triaxial ellipsoid. This noticeably extends the results of \cite{WokesP10} regarding spheroids and has several practical impacts. Indeed, in the context of localization from objects, ellipsoidal models are  most of the time built  in a Least Squares sense from a set of ellipses detected in a small number of images \citep{7919240}. General triaxial ellipsoids are thus generally obtained, whereas spheroids are not likely to occur. In addition, our experiments have shown that the triaxial case is effective even when two eigenvalues are close, thus making the triaxial case useful for almost any case in practice. For instance, considering a camera with position [-1,-1,-4], orientation represented by Euler Angles [-45°,10°,10°] and distance to image plane equal to 1 on one side, and considering an ellipsoid with radii a=4, b=2 and c=1.999999 on the other side, the method still produces numerically stable results\footnote{This experiment can be reproduced using our code (file main.m, experiment Xp=4)}.
\ifhighlightrev
    }
\fi

\ifhighlightrev
    \rev{
\fi
In order that the reader has a better idea of the smooth transition between the triaxial and spheroid cases, a video has been created\footnote{\url{https://members.loria.fr/moberger/Documents/P1E.mp4}}. The loci of solutions for a triaxial ellipsoid with size  axis [a=4,b=2,c=1] is first presented. Then, the third eigenvalue $c$ is progressively moved from $1$ to $2$  to obtain the spheroid case. These loci are computed with the triaxial algorithm (\ref{alg:overall-triaxial}) in the ellipsoid coordinate frame until 1.99 and with  the non-circular elliptic cone (Algo. \ref{alg:overall-noncirc}) with c=2. In the camera frame, Algo. \ref{alg:overall-triaxial} is used for all $c$ values.  Two visualizations are provided: the locus  of ellipsoid solutions is shown in the camera frame, either by representing only the center (in black) and radii endpoints (red,green,blue) or by representing the full ellipsoid.
\ifhighlightrev
    }
\fi

\ifhighlightrev
    \rev{
\fi
\subsection{Sensitivity to noise}
\label{ssec:7-2}
\ifhighlightrev
    }
\fi

\ifhighlightrev
    \rev{
\fi
The work conducted in our previous publications provides insights on the sensibility of the P$1$E framework to noise. 
For instance, while only ellipses inscribed in bounding boxes were considered in \cite{IROS,ISMAR,IROS2,RAL}, a sufficient level of pose accuracy was achieved due to the fact that camera orientations were determined using other methods. Subsequent work has nonetheless demonstrated that detecting more accurate ellipses can further improve the pose accuracy~\citep{ZinsSB20}.
\ifhighlightrev
    }
\fi

\ifhighlightrev
    \rev{
\fi
When the position is known (e.g., regressed by a ConvNet, or computed from the scalar unknowns as in methods of Section \ref{sec:closedform}), preliminary results and analysis presented in \cite{Rathinam_2022_IAC} suggest that the orientation-from-position derivation (Section \ref{ssec:ori}) is highly sensitive to the noise on the ellipse detection, due to the mathematical details of the derivation. More precisely, one can observe that the norm of the relative position $\|\D\|$ is raised to the power of 4 to obtain $\s$ (Equation \eqref{eq:Delta2sigma}), which is itself raised to the power of 1.5 to compute $\mu$ (Equation \eqref{eq:Delta2mu}). This drastically heightens the error resulting from the computation of $\D$. Moreover, the matrix $\D\D^\top$ is made of the pairwise products between $\D$ elements, here again multiplying the errors. 
\ifhighlightrev
    }
\fi

\ifhighlightrev
    \rev{
\fi
\subsection{Improving robustness to noise}\label{sec:robust}
\ifhighlightrev
    }
\fi
\ifhighlightrev
    \rev{
\fi
In practice, we observe that the recovered camera positions  are sensitive to the  noise on the detected ellipse. Theoretically, one eigenvalue of \abb with multiplicity 2 should be found. Due to the noisy ellipse detection, this is in practice not the case. We thus find among the eigenvalues the two which are the closest and identify the remaining one as $\s$. Though this process is effective when noise is small, we think that a method which explicitly take into account the constraints of an eigenvalue with multiplicity two should be considered. We have made progress in this direction in \cite{IROS}  where we exploit the fact that the characteristic polynomial should have a double root which implies that  the discriminant of this polynomial should be zero.   However, such a solution does not really solve the problem  which is due to an imperfect correspondence between the ellipse and the ellipsoid. A more promising research direction  we want to investigate  in the future is to identify an  ellipse close to the detected one which provides the smallest difference, ideally zero, between the two closest eigenvalues of  \abb. A synthetic formulation of the idea  could be for instance
$\widehat{C}=argmin_{\|C-\widehat{C}\|<\epsilon} \|\lambda_1 \left(AB'_{\widehat{C}}\right)-\lambda_2 \left(AB'_{\widehat{C}}\right)\|$, where $C$ is the detected ellipse, $B'_{\widehat{C}}$ is the cone build from ellipse $\widehat{C}$, $\lambda_1\left(AB'_{\widehat{C}}\right)$ and $\lambda_2\left(AB'_{\widehat{C}}\right)$ are the two closest generalized eigenvalues of $\{A,B'_{\widehat{C}}\}$, and $\epsilon$ denotes a small threshold between $C$ and $\widehat{C}$ parameters. This should allow to recover better ellipse-ellipsoid correspondence and  is likely to improve the stability of the pose solutions.
Another solution based on machine learning could be to use 3D aware  tracking \citep{ZinsSB20} to detect the 2D ellipse. Though less generic than the previous solution, this may help to reach stability in the solutions.
\ifhighlightrev
    }
\fi

\section{Conclusion}
\label{sec:conclusion}

We propose in this paper a complete characterization of the P$1$E problem and noticeably extend previous works that only partially addressed this problem. Besides its theoretical interests, this paper also proposes a constructive solution for the camera loci and discusses the potential practical applications. The closed-form solution provided for the position-from-orientation case has proven its practical interest. The orientation-from-position solution, even if not yet fully exploited, may also represent a convenient manner to simplify the pose estimation problem.

Future investigations concern numerical aspects and especially extending the sensitivity analysis of the methods to noise. Another important concern will be the joint use of the method with a minimal number of other image features, such as points or other ellipse-ellipsoid correspondences, to ensure the computation of a unique solution.

\vspace{0.3cm}

\noindent\textbf{Acknowledgements } \textit{The work presented in this paper was carried out at Université de Lorraine, CNRS, Inria, LORIA. The writing effort was partly funded by the Luxembourg National Research Fund (FNR) under the project reference BRIDGES2020/IS/14755859/MEET-A/Aouada.}

\backmatter

\section*{Declarations}

The authors declare that they have no conflict of interest.

\begin{appendices}

\section{Equivalent Problem Formulations}\label{secA1}
To prove that Equations \eqref{eq:main_pb} and \eqref{eq:eq_pb} are  equivalent, we demonstrate below that $\eqref{eq:main_pb}$ implies $\eqref{eq:eq_pb}$ then $\eqref{eq:eq_pb}$ implies $\eqref{eq:main_pb}$.

\noindent
\framebox{$\eqref{eq:main_pb} \implies \eqref{eq:eq_pb}$} Multiply \eqref{eq:main_pb} on the right by $\D$ to obtain $A\D=\s B'\D$ (see Appendix B).
Whence
\begin{equation*}
A\D\D^\top A+\mu A = \s B' \implies \s B'-\s B'\D\D^\top A = \mu A
\end{equation*}
Left-multiplying by $B'^{-1}$
\begin{equation*}
\s I-\s\D\D^\top A = \mu B'^{-1}A
\end{equation*}
Then right-multiplying by $A^{-1}$
\begin{equation*}
\s A^{-1}-\s\D\D^\top = \mu B'^{-1}
\end{equation*}
Finally ($\s\neq0$)
\begin{equation*}
A^{-1}-\D\D^\top = \frac{\mu}{\s} B'^{-1}
\end{equation*}

\noindent
\framebox{$\eqref{eq:eq_pb} \implies \eqref{eq:main_pb}$}
Multiply \eqref{eq:eq_pb} on the right by $A$ then on the left by $B'$ to obtain
\begin{equation*}
    B'-B'\D\D^\top A = \frac{\mu}{\s} A
\end{equation*}
Then right-multiply by $\D$ to obtain $\mu B'\D=\frac{\mu}{\s} A\D$, whence ($\mu\neq0$) $A\D=\s B'\D$.

Injecting that result into he previous equation leads to
\begin{equation*}
    \s B'-A\D\D^\top A = \mu A
\end{equation*}

\section{Proof of Result \ref{res:ABeig}}
\label{appendix:ABeig}
Let us first prove that  the couple \abb has exactly two distinct generalized values, that are non zero and of opposite signs.
\begin{proof}
The generalized eigenvalues of \abb are non-zero because $B'^{-1}A$ is not singular.

We can then observe that 
\begin{equation*}
    Q(x)=\mu x^2-(\mu+1)\s x+\s^2
\end{equation*}
is an annihilator polynomial of $B'^{-1}A$ (see proof in Appendix \ref{secA_Q}):
\begin{equation}\label{eq:anni_poly}
    \mu (B'^{-1}A)^2-(\mu+1)\s B'^{-1}A+\s^2 I=\mbox{\textit{0}.}
\end{equation}

In linear algebra, the minimal polynomial $\pi(.)$ is defined as the monic annihilator polynomial having the lowest possible degree. It can be shown \citep{lang2002} that (i) $\pi(.)$ divides any annihilator polynomial and (ii) the roots of $\pi(.)$ are identical to the roots of the characteristic polynomial. 
Since $Q$ is an annihilator polynomial of degree 2, we can thus infer that $B'^{-1}A$, and thus \ab, has at most two distinct eigenvalues.

We are now going to prove by contradiction that \abb has exactly two distinct eigenvalues. Let's thus assume that the couple has only one eigenvalue with multiplicity 3 denoted $\s_0$.

Since $A$ is positive definite and $B'$ is symmetric, the couple \abb has the following properties \citep{GoluVanl96} (Corollary 8.7.2, p. 462):
\begin{enumerate}
    \item their generalized eigenvalues are real,
    \item their reducing subspaces are of the same dimension as the multiplicity of the associated eigenvalues,
    \item their generalized eigenvectors form a basis of $\RR^3$, and those with distinct eigenvalues are $A$-orthogonal.
\end{enumerate}

According to property 2. above, we have 
\begin{equation*}
    dim(Ker(A-\s_0B'))=3\mbox{,}
\end{equation*}
\textit{i.e.}
\begin{equation*}
    A=\s_0B'\mbox{,}
\end{equation*}
which is impossible because $A$ represents an ellipsoid whereas $B'$ represents a cone. So $\boldsymbol{\{A,B'\}}$ has exactly two distinct generalized eigenvalues.

Let's then denote $\s_1$ (multiplicity 1) and $\s_2$ (multiplicity 2) these two eigenvalues. Observing that $\frac{1}{\s_1}$ and $\frac{1}{\s_2}$ are the generalized eigenvalues of $\{B',A\}$, we can write, according to \cite{minimax} (Theorem 3)
\begin{equation*}
\begin{split}
    \forall \X&\in\RR^{3}\backslash{\bf \{0\}},\\ &min\left(\frac{1}{\s_1},\frac{1}{\s_2}\right)\le \frac{\X^\top B'\X}{\X^\top A\X}\le max\left(\frac{1}{\s_1},\frac{1}{\s_2}\right)
\end{split}
\end{equation*}
If $\s_1$ and $\s_2$ were of the same sign, then $\forall \X\in\RR^3\backslash{\bf \{0\}}$, $\X^\top B'\X$ would be of that sign (since $\X^\top A\X>0$). Yet, it is impossible since $B'$ is neither positive nor negative definite (cone). We thus conclude that the two distinct eigenvalues are of opposite signs.
\end{proof}



Let us now prove that $\s$ is the  generalized  eigenvalue of \abb with multiplicity 1 
\begin{proof}
Let's consider $(\s_1,\de_1)$ and $(\s_2,\de_2)$ the generalized eigenvalues and eigenvectors of \ab, such that $\|\de_i\|=1$.

We are going to prove \eqref{eq:A2sigma} by contradiction.

Let's suppose that there is $k\in\RR^*$ such that $(A,\s_2,k\de_2)$ is solution of Equation \eqref{eq:main_pb}.

By injecting these values into \eqref{eq:main_pb}, we therefore have
\begin{equation*}
    B'-\frac{1}{\s_2}A=MA\mbox{,}
\end{equation*}
where 
\begin{equation*}
    M=\frac{k^2}{\s_2}(A\de_2\de_2^\top-\de_2^\top A\de_2I)\mbox{.}
\end{equation*}
According to property 2 of the proof of Result \ref{res:ABeig}, 
\begin{equation*}
    dim(Ker(B'-\frac{1}{\s_2}A))=2\mbox{,}
\end{equation*}
whence, since A is invertible, 
\begin{equation*}
    dim(Ker(MA))=dim(Ker(M))=2\mbox{.}
\end{equation*}
However, defining
\begin{equation*}
    \de_2^{\perp}=\{\X\in \RR^3 / \X\perp \de_2\}
\end{equation*}
the subspace of dimension 2 orthogonal to $\de_2$, we observe that, $\forall \X \in \de_2^{\perp}$,
\begin{align*}
M\X &= \frac{k^2}{\s_2}A\de_2\de_2^\top \X-\frac{k^2}{\s_2}\de_2^\top A\de_2\X \\
&= \frac{k^2}{\s_2}A\de_2(\de_2\cdot\X)-\frac{k^2}{\s_2}\de_2^\top A\de_2\X \\
&= -\frac{k^2}{\s_2}\de_2^\top A\de_2 \X\mbox{.}
\end{align*}
Since $A$ is positive definite, $\de_2^\top A\de_2>0$, whence it comes
\begin{equation*}
    \forall \X \in \de_2^{\perp}\backslash{\bf \{0\}}\mbox{, }M\X\ne\mathbf{0}\mbox{.}
\end{equation*}

It means that
\begin{equation*}
    \de_2^{\perp}\cap Ker(M)=\{\mathbf{0}\}\mbox{,}
\end{equation*}
whence the direct sum 
\begin{equation*}
    \de_2^{\perp}\bigoplus Ker(M)
\end{equation*}
is a subspace of $\RR^3$ with dimension
\begin{equation*}
    dim(\de_2^{\perp}) + dim(Ker(M)) = 2+2=4\mbox{.}
\end{equation*}
We end up with a contradiction since $4>dim(\RR^3)=3$.

As a result, triplets $(A,\s_2,k\de_2)$ cannot be solutions of \eqref{eq:main_pb}, thus solutions are necessarily in the form $(A,\s_1,k\de_1)$, where $k\in\RR^*$.
\end{proof}

\section{Proving that $Q(B'^{-1}A)=0$}\label{secA_Q}

Replacing \eqref{eq:gevp} into \eqref{eq:main_pb}, we obtain:
\begin{equation*}
    \s^2B'\D\D^\top B'-(\s\D^\top B'\D-1)A=\s B'
\end{equation*}

We can then deduce the following expression for A:
\begin{equation*}
    A=\frac{\s}{1-\s\D^\top B'\D}(B'-\s B'\D\D^\top B')
\end{equation*}

Whence, denoting $I$ the identity matrix and defining $f=\frac{\s}{1-\s\D^\top B'\D}$, then left-multiplying by $B'^{-1}$, we obtain
\begin{align*}
    B'^{-1}A &= f(I-\s\D\D^\top B')
\end{align*}

Squaring that expression leads to
\begin{align*}
    (B'&^{-1}A)^2 = f^2(I-\s\D\D^\top B')^2\\
    &= f^2(I-2\s\D\D^\top B'+\s^2\D(\D^\top B'\D)\D^\top B')\\
    &= f^2(I-2\s\D\D^\top B'+\s^2(\D^\top B'\D)\D\D^\top B')\\
    &= f^2(I-\s(2-\s\D^\top B'\D)\D\D^\top B')
\end{align*}

Defining $\mu=1-\s\D^\top B'\D=1-\D^\top A\D$:
\begin{align*}
    (B'^{-1}A)^2 &= f^2(I-\s(\mu+1)\D\D^\top B')\\
    &= f^2((\mu+1)(I-\s\D\D^\top B')-\mu I)\\
    &= f(\mu+1)B'^{-1}A-f^2\mu I\\
    &= \frac{\s}{\mu}(\mu+1)B'^{-1}A-\frac{\s^2}{\mu}I
\end{align*}

Finally, we have
\begin{equation*}
    \mu(B'^{-1}A)^2 = \s(\mu+1)B'^{-1}A-\s^2I
\end{equation*}

Whence, denoting $Q(x)=\mu x^2-(\mu+1)\s x+\s^2$,
\begin{equation*}
    Q(B'^{-1}A)=0
\end{equation*}

\section{Characterizations of $\mu$}
\label{appendix:mu}
{\bf Proof of result \ref{res:mu}}
\begin{proof}
The trace of $B'^{-1}A$ is given by its eigenvalues:
\begin{equation*}
    tr(B'^{-1}A)=\s_1+2\s_2\mbox{.}
\end{equation*}
Whence, by squaring the matrix,
\begin{equation*}
    tr((B'^{-1}A)^2)=\s_1^2+2\s_2^2\mbox{.}
\end{equation*}
Therefore, since $tr(I)=3$ and $\s=\s_1$, applying the operator $tr()$ to Equation \eqref{eq:anni_poly} leads to
\begin{equation*}
    \mu(\s_1^2+2\s_2^2)-\s_1(\mu+1)(\s_1+2\s_2)+3\s_1^2 =0\mbox{,}
\end{equation*}
which is equivalent to
\begin{equation*}
    \mu\s_2(\s_2-\s_1)=\s_1(\s_2-\s_1)\mbox{,}
\end{equation*}
\textit{i.e.}
\begin{equation*}
    \mu=\frac{\s_1}{\s_2}\mbox{.}
\end{equation*}
\end{proof}

{\bf Proof of result \ref{res:Delta2mu}}
\begin{proof}
Determinant of $B'^{-1}A$ is given by its eigenvalues:
\begin{equation*}
    det(B'^{-1}A)=\s_1\s_2^2\mbox{,}
\end{equation*}
\textit{i.e.}
\begin{equation}\label{eq:detABs1s2}
    \frac{det(A)}{det(B')}=\s_1\s_2^2\mbox{.}
\end{equation}
We obtain \eqref{eq:Delta2mu} by injecting \eqref{eq:A2sigma} \eqref{eq:A2mu} into \eqref{eq:detABs1s2} and using $\mu<0$.
\end{proof}

\section{Proof of co-occurences}\label{secA_cooc}

Obviously, only circular cones can be tangent to a sphere. Furthermore, we are going to prove that only a non-circular elliptic cone can be tangent to a triaxial ellipsoid.

Let's prove it by contradiction, and assume that the projection cone has a revolution axis (circular cone).

Let's also assume that the ellipsoid center $\CC$ does not belong to that axis. Since the ellipsoid is tangent to the cone, any new ellipsoid obtained by rotating the original one around the cone revolution axis shall still be tangent to the cone, thus be solution of \eqref{eq:main_pb}. Yet, in this case, the locus of ellipsoid centers would be a circle located in a plane orthogonal to that axis and whose center would belong to it. Every center would thus be at a fixed distance to the cone vertex $\E$, whence there would be an infinite number of $\D$ solutions for the same $\s$, given \eqref{eq:Delta2sigma}. However, this contradicts Equation \eqref{eq:m2Dell}. Therefore, the ellipsoid center must belong to the revolution axis of the cone.

If the ellipsoid center belonged to the cone revolution axis, then $\D$ would be parallel to that axis, \textit{i.e.} would be an eigenvector of $B'$, whence an eigenvector of $A$ given \eqref{eq:eq_pb}. However, in such a case, the symmetries of the cone-ellipsoid pair would impose that the tangent ellipse (intersection between the ellipsoid and the polar plane derived from $\E$ \citep{Wylie}) belongs to a plane orthogonal to the cone revolution axis, that is also a principal axis of the ellipsoid. Therefore, that tangent ellipse should be both a circle (orthogonal section of a circular cone) and a non-circular ellipse (section of an ellipsoid by a plane parallel to one of its principal planes), which is impossible. Therefore, the cone cannot have a revolution axis.

\section{Vandermonde Matrix}\label{secA_Vandermonde}

A Vandermonde matrix is a matrix with the terms of a geometric progression in each row or column:
\begin{equation*}
    V=\begin{pmatrix}
    1 & 1 & 1 & \cdots & 1 \\
    x_1 & x_2 & x_3 & \cdots & x_n \\
    x_1^2 & x_2^2 & x_3^2 & \cdots & x_n^2 \\
    \vdots & \vdots & \vdots & \ddots & \vdots \\
    x_1^{m-1} & x_2^{m-1} & x_3^{m-1} & \cdots & x_n^{m-1}
    \end{pmatrix}\mbox{.}
\end{equation*}

The determinant of a square Vandermonde matrix (when $m=n$) is given by
\begin{equation*}
    det(V)=\prod_{1\le i<j\le n}(x_j-x_i)\mbox{.}
\end{equation*}
Therefore, $V$ is not singular (\textit{i.e.} $det(V)\neq0$) if and only if all $x_i$ are distinct.

\section{Demonstration of result   \ref{res:sspheroid2}}\label{dem:resspheroid2}

\begin{proof}
Given that $A$ has two distinct eigenvalues, its minimal polynomial is
\begin{equation*}
    \pi_A(x)=(x-\lambda_{A,single})(x-\lambda_{A,double})\mbox{.}
\end{equation*}

$\pi_A$ being an annihilator polynomial of $A$, evaluating $\pi_A(A)$ gives
\begin{equation*}
    A^2=\alpha A+\beta I\mbox{,}
\end{equation*}
where
\begin{align*}
    \alpha&=\lambda_{A,single}+\lambda_{A,double}\mbox{,}\\
    \beta&=-\lambda_{A,single}\lambda_{A,double}\mbox{.}
\end{align*}

Left-multiplying by $\D^\top$ and right-multiplying by $\D$ gives
\begin{equation}\label{eq:pi_spheroid}
    \D^\top A^2\D=\alpha\D^\top A\D+\beta\D^\top\D\mbox{.}
\end{equation}

Injecting the expressions of $\D^\top A^2\D$, $\D^\top A\D$ and $\D^\top\D$ as functions of $m$ from \eqref{eq:sys} into \eqref{eq:pi_spheroid} and observing that  
\begin{equation*}
    tr(A)=\lambda_{A,single}+2\lambda_{A,double}
\end{equation*}
and that
\begin{equation*}
    tr(A^{-1})=\frac{1}{\lambda_{A,single}}+\frac{2}{\lambda_{A,double}}\mbox{,}
\end{equation*}
we obtain, after simplification
\begin{equation*}
    m^3=\frac{tr(B')d}{\lambda_{A,double}}m^2+\frac{\lambda_{A,single}tr(B'^{-1})}{d}m-\frac{\lambda_{A,single}}{\lambda_{A,double}}\mbox{.}
\end{equation*}

Let's call $P_{spheroid}(x)$ the above polynomial whose $m$ is root:
\begin{align*}
    &P_{spheroid}(x)=\\
    &\hspace{3mm}x^3-\frac{tr(B')d}{\lambda_{A,double}}x^2-\frac{\lambda_{A,single}tr(B'^{-1})}{d}x+\frac{\lambda_{A,single}}{\lambda_{A,double}}\mbox{.}
\end{align*}

Developing $tr(B')$ and $tr(B'^{-1})$, one can observe that it can be rewritten
\begin{equation*}
    P_{spheroid}(x) = \left(x-\frac{\lambda_{B',single}}{\lambda_{A,double}}d\right)\left(x-\frac{\lambda_{B',double}}{\lambda_{A,double}}d\right)^2\mbox{.}
\end{equation*}

At this stage, one can note that the sign of $d$ is the sign of $\lambda_{B',simple}$:
\begin{equation*}
    d=\sqrt[3]{\frac{det(A)}{det(B')}}=\sqrt[3]{\frac{\lambda_{A,single}\lambda_{A,double}^2}{\lambda_{B',single}\lambda_{B',double}^2}}\mbox{.}
\end{equation*}
Thus the signs of the roots of $P_{spheroid}(x)$ are:
\begin{equation*}
    \frac{\lambda_{B',single}}{\lambda_{A,double}}d>0 \mbox{\hspace{0.5cm} and \hspace{0.5cm}} \frac{\lambda_{B',double}}{\lambda_{A,double}}d<0\mbox{.}
\end{equation*}

Since $m<0$, the only possible value for it is the second one. Therefore, $\s$ is given by:
\begin{align*}
    \s&=d\left(\frac{\lambda_{B',double}}{\lambda_{A,double}}d\right)^2\\
    &=\frac{\lambda_{A,single}}{\lambda_{B',single}}\mbox{.}
\end{align*}

Let's now focus on $\D$, and consider the first two equations of System \ref{eq:sys}: 
\begin{equation*}
\left\{
    \begin{array}{l}
        \D^\top \D = tr(A^{-1})-\frac{tr(B'^{-1})}{d}m\\
        \D^\top A\D = 1-m^3
    \end{array}
\right.
\end{equation*}
Considering the ellipsoid frame, its left hand side can be rewritten in a matrix form:
\begin{equation*}
    \begin{pmatrix}
    1 & 1 \\
    \lambda_{A,single} & \lambda_{A,double}
    \end{pmatrix}
    \begin{pmatrix}
    \D_{ell,x}^2 \\ \D_{ell,y}^2+\D_{ell,z}^2
    \end{pmatrix}
\end{equation*}
That Vandermonde matrix is not singular given that $ \lambda_{A,single}\neq\lambda_{A,double}$, thus the system can be inverted.

After developing the right hand side, we finally obtain
\begin{equation*}
    \D_{ell,x}^2 = \frac{1}{\lambda_{A,single}}\left(1-\frac{\lambda_{A,single}\lambda_{B',double}}{\lambda_{A,double}\lambda_{B',single}}\right)\mbox{,}
\end{equation*}
and
\begin{equation*}
    \D_{ell,y}^2+\D_{ell,z}^2 = 0\mbox{.}
\end{equation*}

Therefore,
\begin{equation} \label{eq:D_spheroid2}
    \D_{ell}=
    \begin{pmatrix}
    \pm\sqrt{\frac{1}{\lambda_{A,single}}\left(1-\frac{\lambda_{A,single}\lambda_{B',double}}{\lambda_{A,double}\lambda_{B',simple}}\right)} \\
    0 \\
    0
    \end{pmatrix}\mbox{.}
\end{equation}

$\D$ is thus an eigenvector of $A$ corresponding to eigenvalue $\lambda_{A,single}$, {\it i.e.} coincides with the revolution axis of the spheroid.

Equation \eqref{eq:gevp} then requires that $\D$ is also an eigenvector of $B'$. It must be the one corresponding to the revolution axis of the cone since, if not, the ellipsoid center would be located outside of the cone. In that respect, both axes of revolutions (cone and spheroid) coincide, and $\|\D\|$ is given by:
\begin{equation*}
    \|\D\|=\sqrt{\frac{1}{\lambda_{A,single}}\left(1-\frac{\lambda_{A,simple}\lambda_{B',double}}{\lambda_{A,double}\lambda_{B',single}}\right)}\mbox{.}
\end{equation*}
\end{proof}
\section{Ellipsoid and Cone types}\label{secAtypes}

See Tables \ref{tab:ellipsoid} and \ref{tab:cone}.

\begin{table*}
    \centering
    \begin{tabular}{|c|c|c|c|}
        \hline
        & Triaxial ellipsoid & Spheroid & Sphere \\
        \hline
        \raisebox{2.1cm}{Illustration}
        &
        \includegraphics[trim =130mm 10mm 120mm 0mm, clip, width=0.25\linewidth]{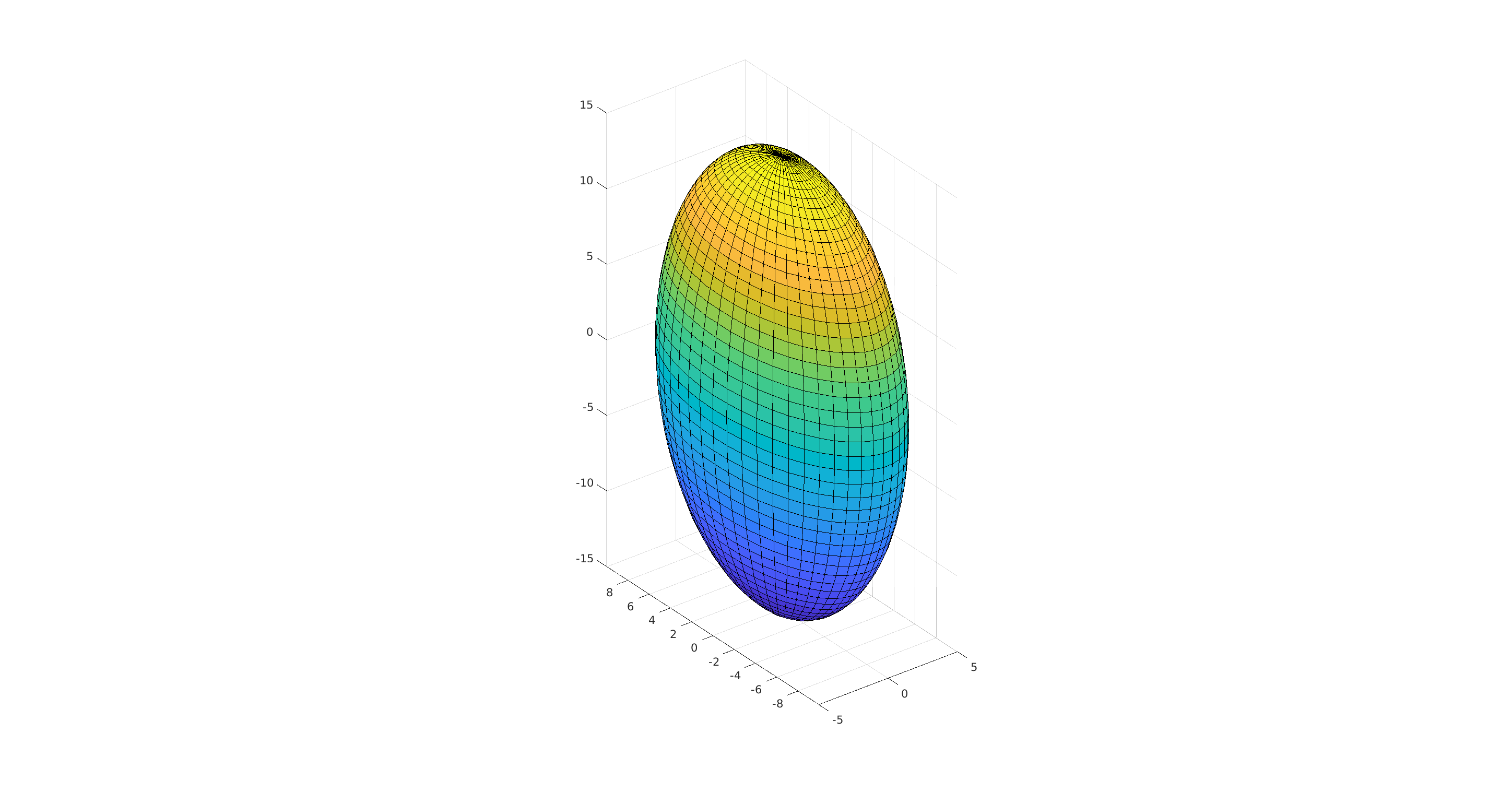}
        &
        \includegraphics[trim =130mm 10mm 120mm 0mm, clip, width=0.25\linewidth]{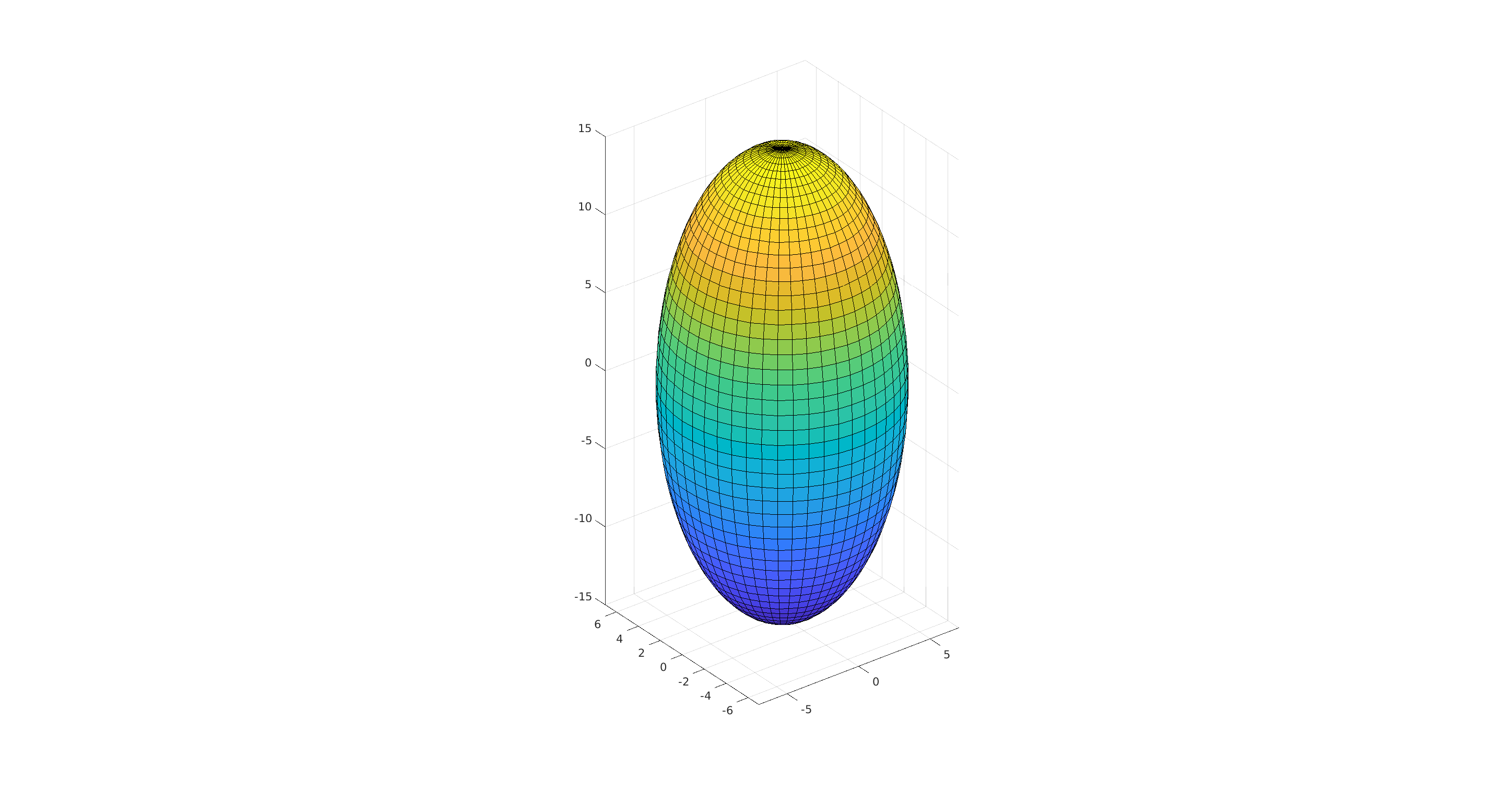}
        &
        \includegraphics[trim =130mm 10mm 120mm 0mm, clip, width=0.25\linewidth]{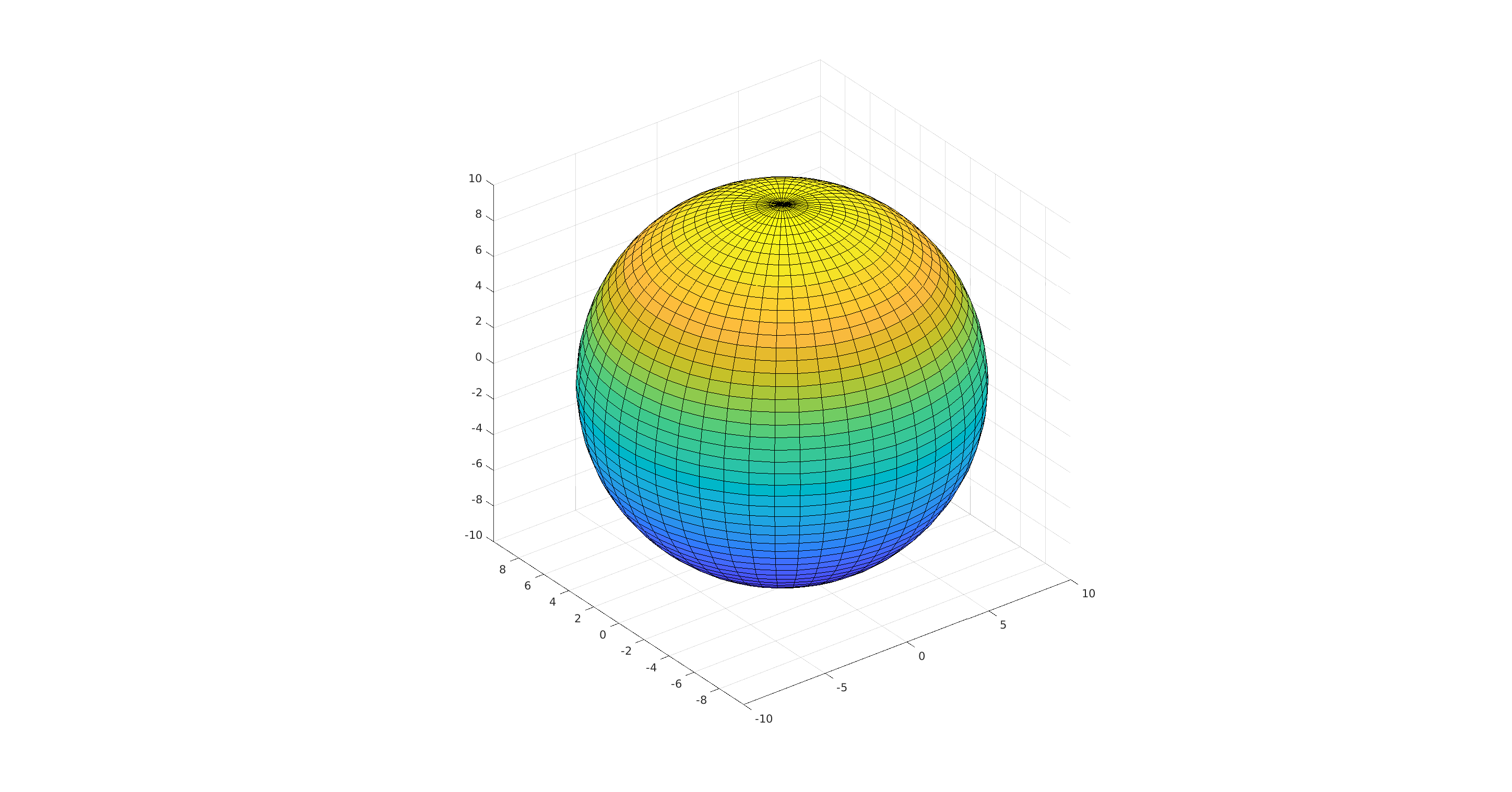}\\
        \hline
        Lengths & \multirow{3}{*}{$a,b,c$} & \multirow{3}{*}{$a,b$} & \multirow{3}{*}{$r$} \\
        of principal & & & \\
        axes & & & \\
        \hline
        Eigenvalues & \multirow{2}{*}{$\frac{1}{a^2},\frac{1}{b^2},\frac{1}{c^2}$} & \multirow{2}{*}{$\frac{1}{a^2},\frac{1}{b^2},\frac{1}{b^2}$} & \multirow{2}{*}{$\frac{1}{r^2},\frac{1}{r^2},\frac{1}{r^2}$} \\
        of $A$ & & & \\
        \hline
        Signs of & \multirow{2}{*}{$+,+,+$} & \multirow{2}{*}{$+,+,+$} & \multirow{2}{*}{$+,+,+$} \\
        eigenvalues & & & \\
        \hline
        Characteristic & \multirow{3}{*}{$P_A(x)=(x-a)(x-b)(x-c)$} & \multirow{3}{*}{$P_A(x)=(x-a)(x-b)^2$} & \multirow{3}{*}{$P_A(x)=(x-r)^3$} \\
        polynomial & & & \\
        of $A$ & & & \\
        \hline
        Minimal & \multirow{3}{*}{$\pi_A(x)=(x-a)(x-b)(x-c)$} & \multirow{3}{*}{$\pi_A(x)=(x-a)(x-b)$} & \multirow{3}{*}{$\pi_A(x)=(x-r)$} \\
        polynomial & & & \\
        of $A$ & & & \\
        \hline
    \end{tabular}
    \smallskip
    \caption{The different types of ellipsoids.}
    \label{tab:ellipsoid}
\end{table*}

\begin{table*}
    \centering
    \begin{tabular}{|c|c|c|}
        \hline
        & Non-circular elliptic cone & Circular cone \\
        \hline
        \raisebox{2.1cm}{Illustration}
        &
        \includegraphics[trim =130mm 10mm 120mm 0mm, clip, width=0.25\linewidth]{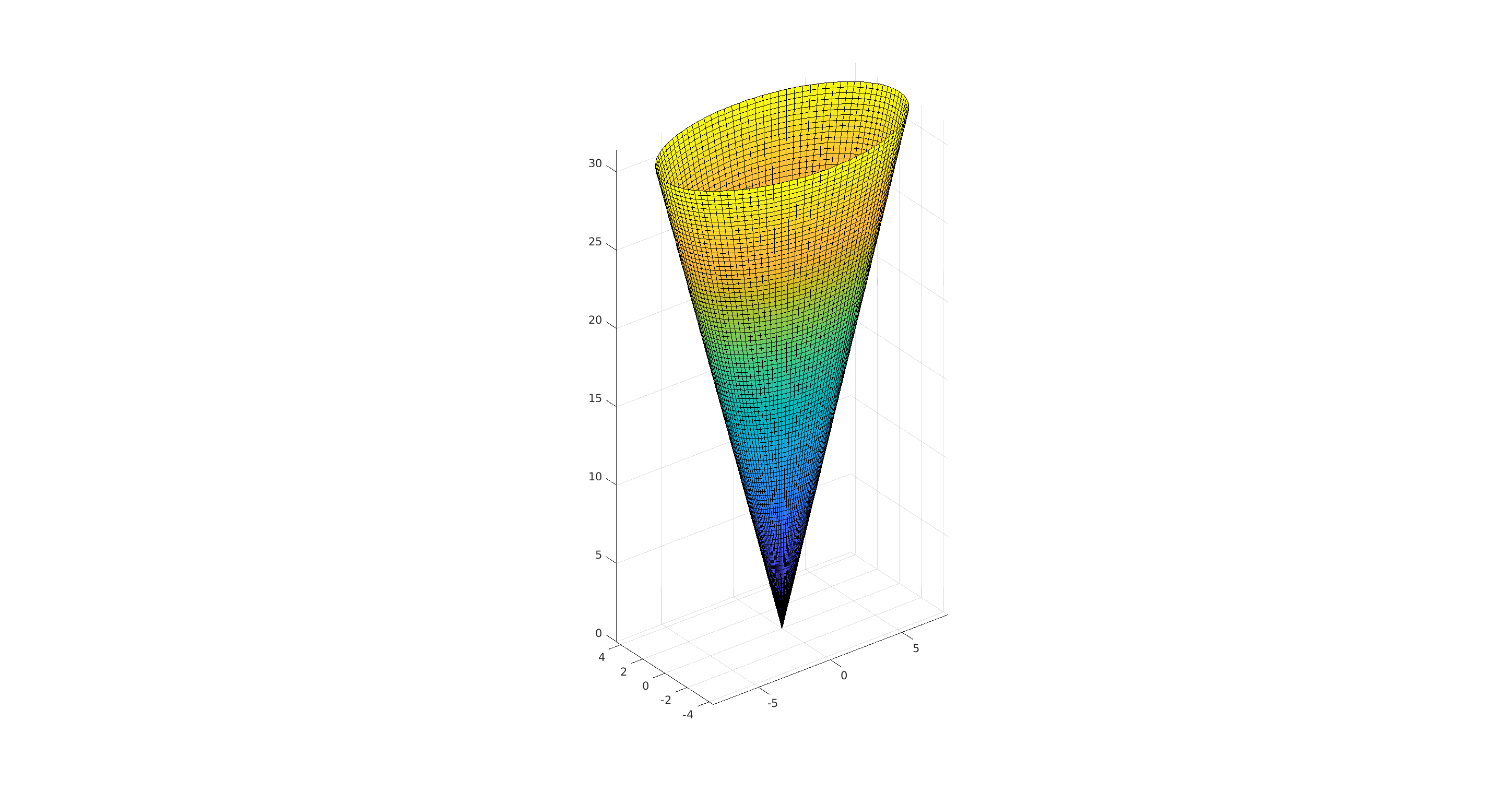}
        &
        \includegraphics[trim =120mm 10mm 120mm 10mm, clip, width=0.25\linewidth]{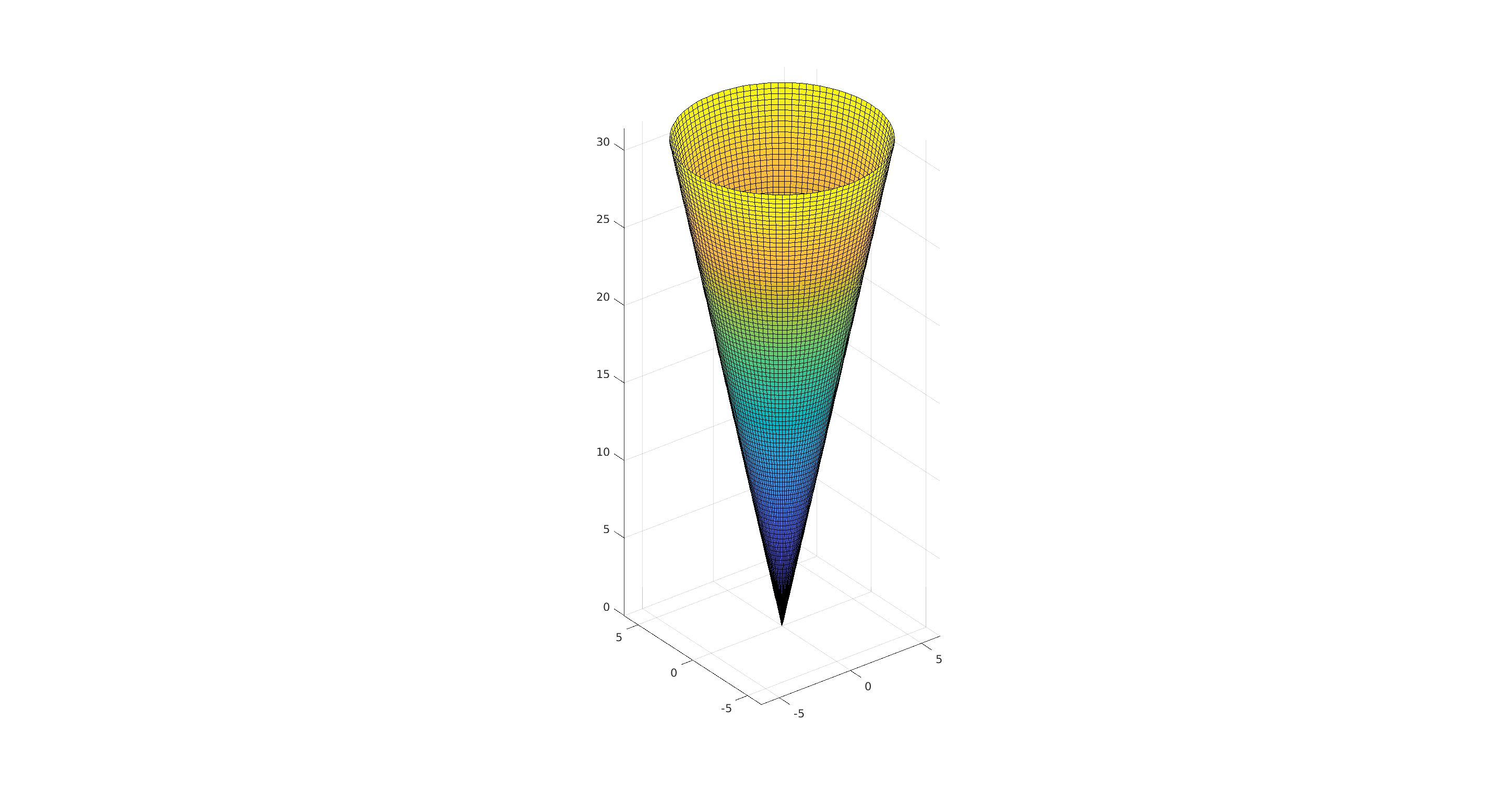}
        \\
        \hline
        & & \\
        Eigenvalues of $B$ & $\lambda_1,\lambda_2,\lambda_3$ & $\lambda_1,\lambda_2,\lambda_2$ \\
        & & \\
        \hline
         & $-,+,+$ & $-,+,+$ \\
        Signs of eigenvalues & or & or \\
         & $+,-,-$ & $+,-,-$ \\
        \hline
        Characteristic & \multirow{2}{*}{$P_B(x)=(x-\lambda_1)(x-\lambda_2)(x-\lambda_3)$} & \multirow{2}{*}{$P_B(x)=(x-\lambda_1)(x-\lambda_2)^2$} \\
        polynomial of $B$ & & \\
        \hline
        Minimal & \multirow{2}{*}{$\pi_B(x)=(x-\lambda_1)(x-\lambda_2)(x-\lambda_3)$} & \multirow{2}{*}{$\pi_B(x)=(x-\lambda_1)(x-\lambda_2)$} \\
        polynomial of $B$ & & \\
        \hline
    \end{tabular}
    \smallskip
    \caption{The different types of elliptic cones.}
    \label{tab:cone}
\end{table*}

\section{Theorem \ref{th:VDMA}: Maple code}
\label{sec:VDMA}
\begin{footnotesize}
\begin{lstlisting}
> with(linalg);
> A:=matrix([[lA1,0,0],[0,lA2,0],[0,0,lA3]]);
> B:=matrix([[lB1,0,0],[0,lB2,0],[0,0,lB3]]);
> M_A:=transpose(vandermonde([lA1,lA2,lA3]));
> d:=(det(A)/det(B))^(1/3);
> V:=transpose(matrix([[trace(inverse(A))-
    trace(inverse(B))*m/d,1-m^3,
    trace(B)*d*m^2-trace(A)*m^3]]));
> Delta2:=multiply(inverse(M_A),V);
> Delta:=transpose(matrix([[sqrt(Delta2[1,1]),
    sqrt(Delta2[2,1]),sqrt(Delta2[3,1])]]));
> inv_B:=evalm(d/m*(inverse(A)-multiply(Delta,
    transpose(Delta))));
> eigenvalues(inv_B);
\end{lstlisting}
\end{footnotesize}

\section{Theorem \ref{th:VDMB}: Maple code}
\label{sec:VDMB}
\begin{footnotesize}
\begin{lstlisting}
> with(linalg);
> A:=matrix([[lA1,0,0],[0,lA2,0],[0,0,lA2]]);
> B:=matrix([[lB1,0,0],[0,lB2,0],[0,0,lB3]]);
> M_B:=transpose(vandermonde([lB1,lB2,lB3]));
> d:=(det(A)/det(B))^(1/3);
> V_:=transpose(matrix([[trace(inverse(A))-
    trace(inverse(B))*m/d,1-m^3,
    trace(B)*d*m^2-trace(A)*m^3]]));
> V:=multiply(inverse(matrix([[1,0,0],[0,d*m^2
    ,0],[0,0,d^2*m^4]])),V_);
> Delta2:=multiply(inverse(M_B),V);
> Delta:=transpose(matrix([[sqrt(Delta2[1,1]),
    sqrt(Delta2[2,1]),sqrt(Delta2[3,1])]]));
> inv_A:=evalm(multiply(Delta,transpose(Delta))
    +evalm(m/d*inverse(B)));
> eigenvalues(inv_A);
\end{lstlisting}
\end{footnotesize}

\section{Solving the Polynomial Equation in Theorem \ref{th:spheroid}}
\label{sec:demospheroid}

In case \#1, the signs of $B'$ eigenvalues ensure that 
\begin{equation*}
    d=\sqrt[3]{\frac{\lambda_{A,single}\lambda_{A,double}^2}{\lambda_{B',1}\lambda_{B',2}\lambda_{B',3}}}>0\mbox{,}
\end{equation*}
then that
\begin{equation*}
    k_1<0 \mbox{ , } k_2>0 \mbox{ and } k_3<0\mbox{.}
\end{equation*}

Let's denote $S_i$ the root of $P_i(x)$ with multiplicity 1, and $D_i$ the root with multiplicity 2, such that
\begin{equation*}
    S_i=\frac{\lambda_{B',i}}{\lambda_{A,single}}d \mbox{\hspace{0.5cm} and \hspace{0.5cm}} D_i=\frac{\lambda_{B',i}}{\lambda_{A,double}}d.
\end{equation*}

Considering the signs of $\lambda_{B',i}$ and the fact that $\lambda_{A,simple}<\lambda_{A,double}$, it comes
\begin{align*}
    S_1&<D_1<0\mbox{,}\\
    S_2&<D_2<0\mbox{,}\\
    0&<D_3<S_3\mbox{.}
\end{align*}

One can therefore note that the roots of $P_1(x)$ and $P_2(x)$ are negative, while those of $P_3(x)$ are positive. Since, in addition, $k_3<0$, we have
\begin{equation*}
    \forall x\leq 0\mbox{, }P_3(x)>0.
\end{equation*}

Let's now focus on the signs of $P_1(x)$ and $P_2(x)$ to determine the locus of possible $m$ values. Since $\lambda_{B',1}<\lambda_{B',2}$, their roots verify
\begin{align*}
    S_1&<S_2\mbox{,}\\
    D_1&<D_2\mbox{.}
\end{align*}

Therefore, one can distinguish between two configurations regaring the roots order:
\begin{equation*}
    S_1<D_1<S_2<D_2\mbox{,}
\end{equation*}
or
\begin{equation*}
    S_1<S_2<D_1<D_2\mbox{.}
\end{equation*}

For the first case, the variations of the three polynomials are presented in Table \ref{tab:config1_1}.

\begin{table*}[th]
\centering
\begin{tabular}{|l|c c c c c c c c c c c c c|}
  \hline
  $x$ & $-\infty$ & & & $S_1$ & & $D_1$ & & $S_2$ & & $D_2$ & & & 0 \\
  \hline
  $P_1(x)$ & & + & & 0 & - & 0 & & & - & & & & \\
  \hline
  $P_2(x)$ & & & & - & & & & 0 & + & 0 & & + & \\
  \hline
  $P_3(x)$ & & & & & & & + & & & & & & \\
  \hline
\end{tabular}
\smallskip
\caption{Signs of $P_i(x)$ when $S_1<D_1<S_2<D_2$.}
\label{tab:config1_1}
\end{table*}

One can observe that the three polynomials are never all non-negative, thus this configuration is impossible. For the second case, however, there is one value for which all three polynomials are non-negative: $D_1$ (see Table \ref{tab:config1_2}).

\begin{table*}[th]
\centering
\begin{tabular}{|l|c c c c c c c c c c c c c|}
  \hline
  $x$ & $-\infty$ & & & $S_1$ & & $S_2$ & & $D_1$ & & $D_2$ & & & 0 \\
  \hline
  $P_1(x)$ & & + & & 0 & & - & & 0 & & - & & & \\
  \hline
  $P_2(x)$ & & & - & & & 0 & & + & & 0 & & + & \\
  \hline
  $P_3(x)$ & & & & & & & + & & & & & & \\
  \hline
\end{tabular}
\smallskip
\caption{Signs of $P_i(x)$ when $S_1<S_2<D_1<D_2$.}
\label{tab:config1_2}
\end{table*}

\section{Proof of Result \ref{res:ssphere}}\label{sec:sphere_annex}

\begin{proof}
Left-multiplying $A=\lambda_{A,triple}I$ by $\D^\top$ right-multiplying it by $\D$, we obtain
\begin{equation*}
    \D^\top A\D=\lambda_{A,triple}\D^\top\D\mbox{.}
\end{equation*}

Injecting the first two equations of System \eqref{eq:sys}, we then have
\begin{equation*}
    1-m^3=\lambda_{A,triple}\left(tr(A^{-1})-\frac{tr(B'^{-1})}{d}m\right)\mbox{.}
\end{equation*}

Developing $tr(A^{-1})$ and $tr(B'^{-1})$ leads to
\begin{equation*}
    1-m^3=3-\frac{\lambda_{A,triple}}{d}\frac{\lambda_{B',double}+2\lambda_{B',simple}}{\lambda_{B',simple}\lambda_{B',double}}m
\end{equation*}

Furthermore, one can observe that
\begin{equation*}
    d=\sqrt[3]{\frac{det(A)}{det(B')}}=\frac{\lambda_{A,triple}}{\lambda_{B',simple}^{1/3}\lambda_{B',double}^{2/3}}\mbox{.}
\end{equation*} 
Whence, injecting this into the former equation:
\begin{align*}
    0
    &=m^3-\frac{\lambda_{B',double}+2\lambda_{B',simple}}{\lambda_{B',simple}^{2/3}\lambda_{B',double}^{1/3}}m+2\\
    &=m^3-\left(\frac{\lambda_{B',double}^{2/3}}{\lambda_{B',simple}^{2/3}}+2\frac{\lambda_{B',simple}^{1/3}}{\lambda_{B',double}^{1/3}}\right)m+2
\end{align*}

Denoting
\begin{equation*}
    R=\sqrt[3]{\frac{\lambda_{B',double}}{\lambda_{B',simple}}}\mbox{,}
\end{equation*}
the last equation means that $m$ is root of the polynomial
\begin{equation*}
    P_{sphere}(x)=x^3-\left(R^2+\frac{2}{R}\right)x+2\mbox{.}
\end{equation*}
Yet, $R$ is an obvious root of this polynomial:
\begin{equation*}
    P_{sphere}(x)=\left(x-R\right)\left(x^2+Rx-\frac{2}{R}\right)
\end{equation*}

Even if obtaining a formal expression of the two other roots is not straighforward, Vieta's formulas provide the following constraints:
\begin{equation*}
\left\{
    \begin{array}{l}
    x_1+x_2+R=0\mbox{,}\\
    x_1x_2R=-2\mbox{.}
    \end{array}
\right.
\end{equation*}
If roots $x_1$ are $x_2$ complex, then $R$ is the only possible value for $m$. If they are real, then, since $R<0$ ($B'$ eigenvalues are of opposite signs), the second formula requires that $x_1$ and $x_2$ are of the same sign, and the first formula requires that they are positive. Finally,
\begin{equation*}
    m=R\mbox{.}
\end{equation*}

Corresponding $\s$ value is:
\begin{align*}
    \s&=dR^2\\
    &=\frac{\lambda_{A,triple}}{\lambda_{B',single}^{1/3}\lambda_{B',double}^{2/3}}\frac{\lambda_{B',double}^{2/3}}{\lambda_{B',single}^{2/3}}\\
    &=\frac{\lambda_{A,triple}}{\lambda_{B',single}}\mbox{.}
\end{align*}

Applying $tr()$ to Equation \eqref{eq:eq_pb} gives the value of $\|\D\|^2$:
\begin{equation*}
    \|\D\|^2=\frac{3}{\lambda_{A,triple}}-\frac{\mu}{\s}\left(\frac{1}{\lambda_{B',single}}+\frac{2}{\lambda_{B',double}}\right)\mbox{.}
\end{equation*}

Given
\begin{equation*}
    \s=\frac{\lambda_{A,triple}}{\lambda_{B',simple}}\mbox{,}
\end{equation*}
and
\begin{equation*}
    \mu=m^3=\frac{\lambda_{B',double}}{\lambda_{B',simple}}\mbox{,}
\end{equation*}
the right hand side can be rewritten
\begin{equation*}
    \frac{3}{\lambda_{A,triple}}-\frac{\lambda_{B',double}}{\lambda_{A,triple}}\left(\frac{1}{\lambda_{B',simple}}+\frac{2}{\lambda_{B',double}}\right)\mbox{,}
\end{equation*}
\textit{i.e.}
\begin{align*}
    \|\D\|^2&=\frac{3}{\lambda_{A,triple}}-\frac{\lambda_{B',double}}{\lambda_{B',simple}}\frac{1}{\lambda_{A,triple}}-\frac{2}{\lambda_{A,triple}}\\
    &=\left(1-\frac{\lambda_{B',double}}{\lambda_{B',simple}}\right)\frac{1}{\lambda_{A,triple}}\mbox{.}
\end{align*}
\end{proof}

\end{appendices}

\bibliography{sn-bibliography}

\end{document}